\documentclass[twoside,11pt]{article}

\usepackage{tkz-graph}
\GraphInit[vstyle = Shade]
\tikzset{
  VertexStyle/.append style = { inner sep=5pt,
                                text = black, font = \Large\bfseries},
  EdgeStyle/.append style = {->, bend left} }

\usepackage{tikz}
\usetikzlibrary{decorations.markings}
\usetikzlibrary{shapes.geometric}

\pgfdeclarelayer{edgelayer}
\pgfdeclarelayer{nodelayer}
\pgfsetlayers{edgelayer,nodelayer,main}

\tikzstyle{none}=[inner sep=0pt]
\definecolor{hexcolor0xff0000}{rgb}{1.000,0.000,0.000}
\definecolor{hexcolor0x000000}{rgb}{0.000,0.000,0.000}
\definecolor{hexcolor0x00ff00}{rgb}{0.000,1.000,0.000}
\definecolor{hexcolor0x000000}{rgb}{0.000,0.000,0.000}
\definecolor{hexcolor0x000000}{rgb}{0.000,0.000,0.000}
\definecolor{hexcolor0x000000}{rgb}{0.000,0.000,0.000}

\tikzstyle{rn}=[circle,fill=hexcolor0xff0000,draw=hexcolor0x000000,line width=0.8 pt]
\tikzstyle{gn}=[circle,fill=hexcolor0x00ff00,draw=hexcolor0x000000,line width=0.8 pt]
\tikzstyle{yn}=[circle,fill=Yellow,draw=hexcolor0x000000,line width=0.8 pt]

\tikzstyle{simple}=[-,draw=hexcolor0x000000,line width=2.000]
\tikzstyle{arrow}=[-,draw=hexcolor0x000000,postaction={decorate},decoration={markings,mark=at position .5 with {\arrow{>}}},line width=2.000]
\tikzstyle{tick}=[-,draw=hexcolor0x000000,postaction={decorate},decoration={markings,mark=at position .5 with {\draw (0,-0.1) -- (0,0.1);}},line width=2.000]

\usepackage{adjustbox}
\usepackage{array}
\usepackage{booktabs}
\usepackage{multirow}
\usepackage{xcolor}
\usepackage{colortbl}

\newcolumntype{R}[2]{%
    >{\adjustbox{angle=#1,lap=\width-(#2)}\bgroup}%
    l%
    <{\egroup}%
}

\tikzstyle myBG=[line width=3pt,opacity=1.0]

\usepackage{jmlr2e}
\usepackage{color}

\usepackage{natbib}

\usepackage{amssymb}
\usepackage{amsmath} 
\usepackage{mathrsfs}
\usepackage{mathtools}
\usepackage{bigdelim}
\usepackage{subcaption}
\usepackage{pgfplots}
\usepackage{enumerate}
\usepackage{bm}         %For bolding with greek letters
\usepackage{accents}
\newcommand\norm[1]{\left\lVert#1\right\rVert}

\usepackage{rotating}
\usepackage{verbatim}
\usepackage{mathtools}
\usepackage{hhline}
\newcommand{\Mypm}{\mathbin{\tikz [x=1.4ex,y=1.4ex,line width=.1ex] \draw (0.0,0) -- (1.0,0) (0.5,0.08) -- (0.5,0.92) (0.0,0.5) -- (1.0,0.5);}}%

\newcolumntype{P}[1]{>{\centering\arraybackslash}p{#1}}

\usepackage{algorithm}
\usepackage{algpseudocode}
\newcommand\munderbar[1]{%
  \underaccent{\bar}{#1}}

\newcommand\ci{\perp\!\!\!\perp}

\newcommand{\qed}{\nobreak \ifvmode \relax \else
      \ifdim\lastskip<1.5em \hskip-\lastskip
      \hskip1.5em plus0em minus0.5em \fi \nobreak
      \vrule height0.75em width0.5em depth0.25em\fi}

\ShortHeadings{An HASMM Model for Informatively Censored Temporal Data}{Alaa and van der Schaar}
\firstpageno{1}

\begin{document}

\title{A Hidden Absorbing Semi-Markov Model for Informatively Censored Temporal Data: Learning and Inference}

\author{\name Ahmed M. Alaa$^{\dagger}$ \email ahmedmalaa@ucla.edu \\
        \addr $^{\dagger}$Electrical Engineering Department\\
         University of California, Los Angeles (UCLA)\\
         Los Angeles,  CA 90095-1594, USA\\
				\\
        \name Mihaela van der Schaar$^{\ast,\dagger}$ \email mihaela.vanderschaar@eng.ox.ac.uk\\ 
				\addr $^{\ast}$Department of Engineering Science\\
         University of Oxford\\
         Parks Road, Oxford OX1 3PJ, UK}

\editor{xxxxxxxxxxxxx}

\maketitle
\begin{abstract}
Modeling continuous-time physiological processes that manifest a patient's evolving clinical states is a key step in approaching many problems in healthcare. In this paper, we develop the {\it Hidden Absorbing Semi-Markov Model} (HASMM): a versatile probabilistic model that is capable of capturing the modern electronic health record (EHR) data. Unlike existing models, the HASMM accommodates irregularly sampled, temporally correlated, and informatively censored physiological data, and can describe non-stationary clinical state transitions. Learning the HASMM parameters from the EHR data is achieved via a novel {\it forward-filtering backward-sampling} Monte-Carlo EM algorithm that exploits the knowledge of the end-point clinical outcomes (informative censoring) in the EHR data, and implements the E-step by sequentially sampling the patients' clinical states in the reverse-time direction while conditioning on the future states. Real-time inferences are drawn via a forward-filtering algorithm that operates on a virtually constructed discrete-time {\it embedded Markov chain} that mirrors the patient's continuous-time state trajectory. We demonstrate the prognostic utility of the HASMM in a critical care prognosis setting using a real-world dataset for patients admitted to the Ronald Reagan UCLA Medical Center. In particular, we show that using HASMMs, a patient's clinical deterioration can be predicted 8-9 hours prior to intensive care unit admission, with a 22$\%$ AUC gain compared to the Rothman index, which is the state-of-the-art critical care risk scoring technology.
\end{abstract}

\section{Introduction}
\label{sec1}
\begin{figure*}[t]
        \centering
        \includegraphics[width=5.5in]{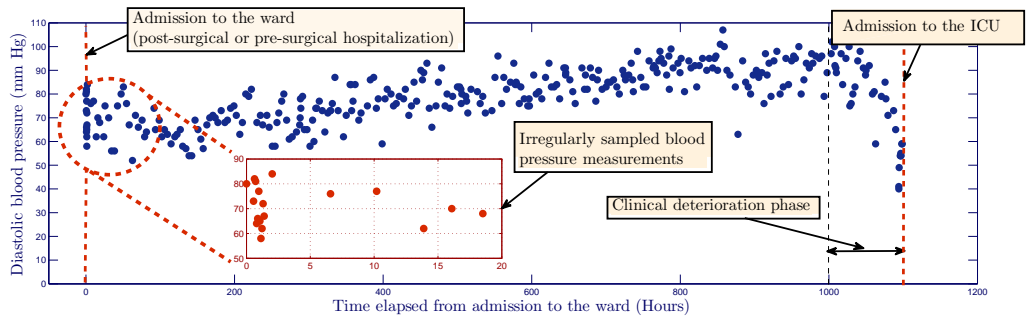}
				\captionsetup{font= small}
        \caption{An episode of the diastolic blood pressure measurements (as recorded in the EHR) for a patient hospitalized in a regular ward for 50 days and then admitted to the ICU after the ward staff realized she is clinically deteriorating. Measurements are censored in accordance with the ICU admission time.}
\label{Fiq1}
\end{figure*}
\begin{figure*}[t]
        \centering
        \includegraphics[width=5.5in]{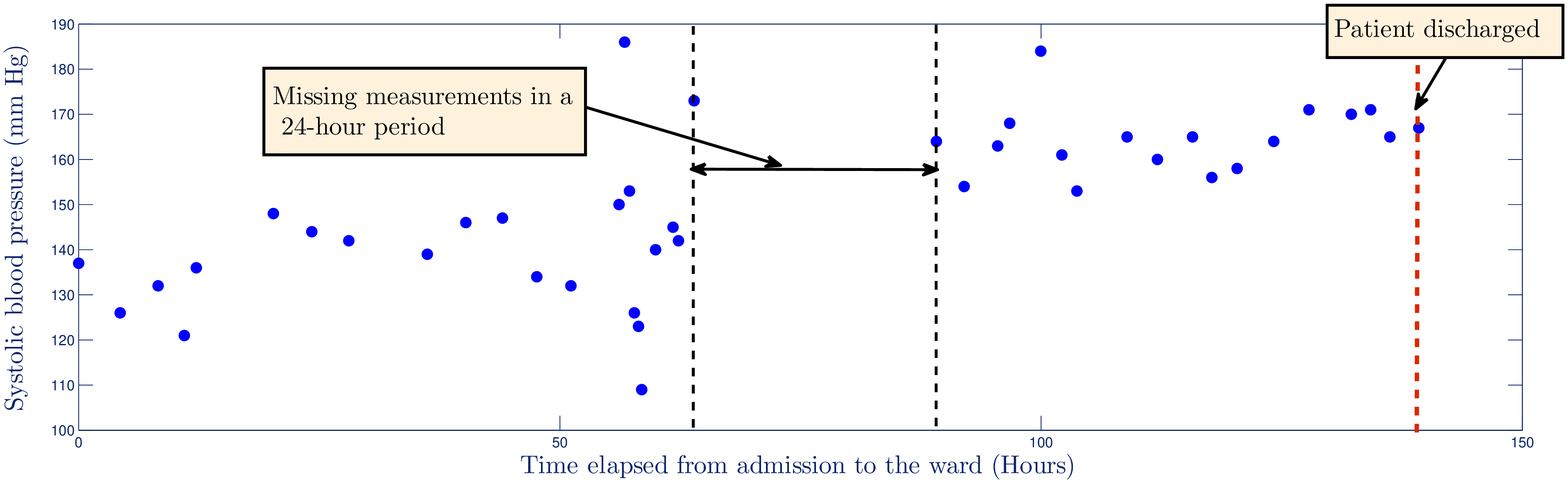}
				\captionsetup{font= small}
        \caption{An episode of the systolic blood pressure measurements for a patient hospitalized in a regular ward for 6 days and then discharged home by the ward staff. Measurements are missing in a 24-hour period during the patient's stay in the ward.}
\label{Fiq2}
\end{figure*}

Modeling the clinical conditions of a patient using evidential physiological data is a ubiquitous problem that arises in many healthcare settings, including disease progression modeling (\cite{schulam2015framework, mould2012models, wang2014unsupervised, jackson2003multistate, sweeting2010multi, liu2015efficient}) and critical care prognosis (\cite{moreno2005saps, matos2006detection, ForecastICU, hoiles2016non, alaa2016personalized}). Accurate physiological modeling in these settings confers an {\it instrumental value} that manifests in the ability to provide early diagnosis, individualized treatments and timely interventions (e.g. early warning systems in critical care hospital wards (\cite{ForecastICU}), early diagnosis for Scleroderma patients (\cite{varga2012scleroderma, ibrahim2016balancing}), early detection of a progressing breast cancer (\cite{bartkova2005dna}), etc). Physiological modeling also confers an {\it epistemic value} that manifests in the knowledge extracted from data about the progression and severity phases of a disease (\cite{stelfox2012intensive})), or the short-term dynamics of the physiological behavior of critically ill patients (\cite{li2013tracking}). \\  

The recent availability of data in the electronic health records (EHR)\footnote{A recent data brief from the Office for National Coordinator (ONC) for healthcare technology shows that the adoption of EHR in US hospitals exhibited a spectacular increase from 9.4$\%$ in 2008, 27.6$\%$ in 2011, to 75.5$\%$ in 2014 (\cite{charles2015electronic}).} creates a promising horizon for establishing rich and complex physiological models (\cite{gunter2005emergence}). Modern EHRs comprise {\it episodic} data records for individual (anonymized) patients; every patient's episode is a temporal sequence of clinical findings (e.g. visual field index for Glaucoma patients (\cite{liu2015efficient}), CD4 cell counts for HIV-infected patients (\cite{guihenneuc2000modeling}), etc), lab test results (e.g. white cell blood count for post-operative patients under immunosuppressive drugs (\cite{cholette2012washing}), etc), or vital signs (e.g. blood pressure and $O_2$ saturation (\cite{ForecastICU})). The time span of these episodes may be as short as few days in short-term hospitalization episodes (e.g. patients with solid tumors, hematological malignancies or neutropenia who are hospitalized in regular wards before or after a surgery (\cite{kause2004comparison, hogan2012preventable, kirkland2013clinical})), or as long as few years in longitudinal episodes (e.g. chronic obstructive pulmonary disease may evolve from a mild Stage I to a very severe Stage IV over a time span of 10 years (\cite{pedersen2011global, wang2014unsupervised})). In this paper, we develop a versatile physiological model that fits a wide spectrum of healthcare settings, providing means for data-driven clinical prognosis. In the next Subsection, we expose our modeling rationale and list the modeling challenges posed by the structure of modern EHR data. We conclude this Section by summarizing our contributions in Subsection \ref{secc12}. 

\subsection{Modeling Rationale and Challenges}
\label{secc11}
\subsubsection{Rationale}
Previous physiological models have branched into two different modeling paths with respect to the way a patient's clinical states are defined. One strand of literature adopts {\it fully observable models}; these models assume that clinical states are quantifiable via {\it observable} clinical markers or disease severity measures (e.g. PFVC in Scleroderma (\cite{schulam2015framework}), GFR in kidney disease (\cite{eddy2006chronic}), etc). Another strand of literature adopts {\it latent variable models}, which assume that clinical states are latent and manifest only through proximal, noisy physiological measurements. Table \ref{table1} lists some notable previous works that fall under each modeling category.  
 
\begin{table}[H]
\caption{Modeling methodologies in previous works.} 
\label{table1} 
\begin{center}
    \begin{tabular}{ | c || P{10cm} |}
    \hline
    {\bf Methodology} & {\bf Previous Works} \\ \hline \hline
		&  \\
    {\bf Fully observable models} & $\bullet$ HIV (\cite{dessie2014multi, foucher2005semi}) $\bullet$ Chronic kidney diseases (\cite{eddy2006chronic}) $\bullet$ Scleroderma (\cite{schulam2015framework}) $\bullet$ ICU (\cite{ghassemi2015multivariate}).\\ 
		&  \\ \hline
		&  \\ 
    {\bf Latent variable models}  & $\bullet$ Alzheimer (\cite{chen2011non}) $\bullet$ HIV (\cite{guihenneuc2000modeling}) $\bullet$ Glaucoma (\cite{liu2015efficient}) $\bullet$ Comorbidities (\cite{wang2014unsupervised}).\\ 
		&  \\
    \hline
    \end{tabular}
\end{center}
\end{table}
Our modeling choice is to go with a latent variable model. The rationale behind this choice is explicated as follows.
\begin{itemize}
\item In a wide range of problems, a concrete clinical marker that can be directly used as a surrogate for the patient's true clinical condition is available. This is especially true in critical care settings where no solid definition or measure of a ``clinical state" exists (\cite{li2013tracking}). Previous works that adopted a clinical risk score as a surrogate for the clinical state in critical care settings have found that other physiological features, when augmented with the clinical risk score, still hold a significant predictive power with respect to end-point clinical outcome (\cite{ghassemi2015multivariate}). This implies that a clinical risk score or a severity of illness measure (such as APACHE II, SAPS and SOFA (\cite{knaus1991apache, subbe2001validation})) is not a sufficient measure of a patient's true clinical condition, and hence cannot be reliably modeled as an observable clinical state. 
\item The same line of argument extends to disease progression models: (\cite{jackson2003multistate}) has shown that significant modeling gain can be attained by treating clinical markers and diagnostic assessments as noisy, potentially erroneous manifest variables for the patient's true clinical state rather than defining a clinical state in terms of those markers. 
\item For various chronic disease, such as HIV, Scleroderma, and kidney disease, progression stages are well defined in terms of observable clinical markers (CD4 cell count, PFVC and GFR). However, a latent variable model can help validate and assess the current domain knowledge-based clinical practice guidelines by learning alternative, data-driven guidelines. Other diseases, such as COPD, have their progression stages manifesting only through symptoms (e.g. chronic bronchitis, emphysema and chronic airway obstruction (\cite{wang2014unsupervised})), which may or may not accurately reflect the disease's true state, and hence a latent variable model is necessary.     
\item Conclusive clinical markers that reveal a patient's true state may be available only occasionally in a patient's longitudinal episodes. For instance, in a breast cancer progression setting, most of the data points associated with a patient's longitudinal episode would be imaging test results (e.g. BI-RADS scores of a mammogram or an MRI (\cite{gail2010comparing, taghipour2013parameter})), which are noisy markers for the existence of a tumor, whereas a conclusive biopsy result that truly reveals whether the patient is in a preclinical or clinical breast cancer state may not be available because the patient did not undergo a biopsy test. A latent variable model better suits such settings.   
\item A fully observable model does not provide diagnostic utility since it assumes that an already observable clinical marker provides an immediate, domain-knowledge-based diagnosis for the patient. Contrarily, a latent variable model leaves room for diagnoses to be learned from the evidential data by learning the association between physiological evidence and clinical states, which may help inform and improve clinical practice.   
\end{itemize}

\subsubsection{Challenges}
Hidden Markov Models (HMMs) and their variants have been widely deployed as temporal latent variable models for dynamical systems (\cite{smyth1994hidden}; \cite{zhang2001segmentation}; \cite{giampieri2005analysis}; \cite{genon2000stochastic}; \cite{ghahramani1997factorial}). Such models have achieved considerable success in various applications, such as topic modeling (\cite{gruber2007hidden}), speaker diarization (\cite{fox2011sticky}), and speech recognition (\cite{rabiner1989tutorial}). However, the nature of the clinical setting, together with the format of the modern EHR data pose the following set of serious challenges that confound classical HMM models:\\
\\     
{\bf (A) Non-stationarity:} Recently developed disease progression models, such those in (\cite{wang2014unsupervised}) and (\cite{liu2015efficient}), use conventional stationary Markov chain models. In particular, they assume that state transition probabilities are independent of time. However, this assumption is seriously at odds with even casual observational studies which show that the probability of transiting from the current state to another state depends on the time spent in the current state (\cite{lagakos1978semi, huzurbazar2004multistate, gillaizeau2015multistate}). This effect, which violates the memorylessness assumptions adopted by continuous-time Markovian models, was verified in patients who underwent renal transplantation (\cite{foucher2007semi, foucher2008flexible}), patients who are HIV infected (\cite{joly1999penalized, dessie2014multi, foucher2005semi}), and patients with chronic obstructive pulmonary disease (\cite{bakal2014heart, wang2014unsupervised}).\\
\\
{\bf (B) Irregularly spaced observations:} The times at which the clinical findings of a patient (vital signs or lab tests) are observed is controlled either by clinicians (in the case of hospitalized inpatients), or by the patient's visit times (in the case of a chronic disease follow up). The time interval between every two measurements may vary from one patient to another, and may also vary for the same patient within her episode. This is reflected in the structure of the episodes in the EHR records, as shown in Figure \ref{Fiq1} and \ref{Fiq2}. Figure~\ref{Fiq1} depicts an actual diastolic blood pressure episode for a patient hospitalized in a regular ward for 1200 hours (50 days)\footnote{A detailed description for the data involved in this paper is provided in Section \ref{sec4}.}. The patient's stay in the ward was concluded with an admission to the ICU after the ward staff realized she was clinically deteriorating. As we can see, the blood pressure measurements in the first 20 hours were initially taken with a rate of 1 sample per hour, and then later the rate changed to 1 sample every 5 hours\footnote{While Figure \ref{Fiq1} illustrates a short-term episode for a critical care patient, similar effects are experienced in longitudinal episodes for patients with chronic disease (see Figure 4 in (\cite{wang2014unsupervised})).}. Thus, a direct application of a regular, discrete-time HMM (e.g. the models in (\cite{murphy2002hidden, fox2011sticky, fox2011bayesian, rabiner1989tutorial, yu2010hidden, matos2006detection, guihenneuc2000modeling})) will not suffice for jointly describing the latent states and observations, and hence ensuring accurate inferences. \\
\\ 
{\bf (C) Discrete observations of a continuous-time phenomena:} A patient's physiological signals and latent states evolve in continuous time; however, the observed physiological measurements are gathered at discrete time steps. The intervals between observed measurements can vary quite significantly; as we can see in Figure \ref{Fiq2}, the systolic blood pressure for a patient who stayed in a ward for 140 hours exhibits an entire day without measurements~\footnote{This may have resulted due to the patient undergoing a surgery or an intervention, or because the EHR recording system accidentally did not receive the data from the clinicians during that day.}. This means that the patient may encounter multiple hidden state transitions without any associated observed data. These effects make learning and inference problems more complicated since the inference algorithms need to consider potential unobserved trajectories of state evolution between every two timestamps. This challenge has been recently addressed in (\cite{nodelman2012expectation, wang2014unsupervised, liu2015efficient}), but only on the basis of memoryless Markov chain models for the hidden states, for which tractable inferences that rely on the solutions to Chapman-Kolmogorov equations can be executed. However, incorporating non-stationarity in state transitions (i.e. addressing challenge (1) in this list) would make the problem of reasoning about a continuous-time process through discrete observations much more complicated.\\
\\
{\bf (D) Lack of supervision:} The episodes in the EHR may be labeled with the aid of domain knowledge (e.g. the stages and symptoms of some chronic diseases, such as chronic kidney disease (\cite{eddy2006chronic}), are known to clinicians and may be provided in the EHR). However, in many cases, including the case of (post or pre-operative) critical care, we do not have access to any labels for the patients' states. Hence, unsupervised learning approaches need to be used for learning model parameters from EHR episodes. While unsupervised learning of discrete-time HMMs has been extensively studied and is well understood (e.g. the Baum-Welch EM algorithm is predominant in such settings (\cite{zhang2001segmentation, yu2010hidden, rabiner1989tutorial})), the problem of unsupervised learning of continuous-time models for which both the patient's states and state transition times are hidden is far less understood, and indeed far more complicated. \\
\\
{\bf (E) Censored observations:} Episodes in the EHR are usually terminated by an informative intervention or event, such as death, ICU admission, discharge, etc. This is known as {\it informative censoring} (\cite{scharfstein2002estimation, huang2002frailty, link1989model}). Unlike classical HMM settings where training sets comprise fixed length, or arbitrarily-censored, HMM sequence instances, a typical EHR dataset would comprise a set of episodes with different durations, and the duration of each episodes is itself informative of the entire state evolution trajectory. Learning in such settings requires novel algorithms that can efficiently compute the likelihood of observing a set of episodes conditioned on their durations and terminating states.  \\ 

\subsection{Summary of Contributions}
\label{secc12}
In order to address the challenges above, we develop a new model --which we call the {\it Hidden Absorbing Semi-Markov Model} (HASMM)-- as a versatile generative model for a patient's (physiological) episode as recorded in the EHR. The HASMM captures non-stationary transitions for a patient's clinical state via a continuous-time semi-Markov model with explicitly specified state sojourn time distributions. Informative censoring is captured via absorbing states that designate clinical endpoint outcomes (e.g. cardiac arrest, mortality, recovery, etc); entering an absorbing state of an HASMM stimulates censoring events (e.g. clinical deterioration leads to an ICU admission which terminates the physiological observations for a monitored patient in a ward, etc). Observable variables are modeled via a multi-task Gaussian process (\cite{bonilla2007multi}), for which the observation times (i.e. follow up visits, vital sign gathering, lab tests, etc) are modeled as a point process. Using multi-task Gaussian process with state-dependent hyper-parameters, an HASMM accounts for both correlations among different physiological variables, in addition to the temporal correlations among the observation variables that are generated by the same hidden state during its sojourn period. In that sense, an HASMM is a segment model (\cite{ostendorf1996hmm}) and also a {\it state-switching} model (\cite{fox2011bayesian})). \\

To allow for real-time inference of a patient's state, we develop a forward-filtering HASMM inference algorithm that can estimate a patient's latent state using her history of irregularly sampled physiological measurements. The inference algorithm operates by constructing a virtual, discrete-time {\it embedded Markov chain} that fully describes the patient's state transitions at observation times. The embedded Markov chain is constructed in an offline stage by solving a system of {\it Volterra integral equations of the second kind} using the {\it successive approximation} method; the solution to this system of equations, which parallels the Chapman-Kolmogorov equations in ordinary Markov chains, describe the HASMM's semi-Markovian state transitions as observed at arbitrarily selected discrete timestamps. \\       

Offline learning of the HASMM model parameters from patients' episodes in an EHR is a daunting task. Since the HASMM is a continuous-time model, we cannot directly use the classical Baum-Welch EM algorithms for learning its parameters (\cite{rabiner1989tutorial}). Moreover, the semi-Markovianity of an HASMM yields an intractable integral in the E-step of the Expectation-Maximization (EM) formulation. Since the HASMM's state transitions are not captured by the conventional continuous-time Markov chain transition rate matrices, we cannot make use of the {\it Expm} and {\it Unif} methods that were used in (\cite{hobolth2011summary}), and more recently in (\cite{liu2015efficient}) for evaluating the integrals involved in the E-step of learning continuous-time HMMs. To address this challenge, we develop a novel {\it forward-filtering backward-sampling Monte Carlo EM} (FFBS-MCEM) algorithm that approximates the integral involved in the E-step by efficiently sampling the latent clinical trajectories conditioned on observations in the EHR by exploiting the informative censoring of the patients' episodes. The FFBS-MCEM algorithm samples the latent clinical states of every (offline) patient episode in the EHR as follows: it starts from the known clinical endpoints, and sequentially samples the patient's states by traversing in the reverse-time direction while conditioning on the future states, and then uses the sampled state trajectories to evaluate a Monte Carlo approximation for the E-step.  \\    

The rest of the paper is organized as follows. In Section \ref{sec2}, we present the HASMM model. The HASMM inference algorithm is developed in Section \ref{sec3}, and the learning algorithm is developed in Section \ref{sec3.3}. In Section \ref{sec4}, we demonstrate the utility of the HASMM in the problem of critical care prognosis using a real-world dataset for patients admitted to Ronald Reagan UCLA Medical Center. Conclusions are drawn in Section \ref{sec5}.

\section{The Hidden Absorbing Semi-Markov Model (HASMM)}
\label{sec2}
In this section, we introduce the basic abstract structure of the continuous-time HASMM (Subsection \ref{sec2.1}), and then we propose the distributional specifications for the model's variables (Subsection \ref{sec2.2}).
\subsection{Abstract Model}
\label{sec2.1}
We start by describing the HASMM's hidden state evolution process, and then we describe the structure of its observable variables.
\subsubsection{Hidden States}
\label{sec2.1.1}
We consider a filtered probability space $(\Omega, \mathcal{F},\{\mathcal{F}_{t}\}_{t\in\mathbb{R}_{+}}, \mathbb{P})$, over which a continuous-time stochastic process $X(t)$ is defined on $t \in \mathbb{R}_{+}$. The process $X(t)$ corresponds to a temporal trajectory of the patient's hidden clinical states, which take on values from a finite {\it state-space} $\mathcal{X} = \{1,2,.\,.\,., N\}$. Because the process $X(t)$ takes on only finitely many values, it can be decomposed in the form\footnote{By convention, we set $\tau_1 = 0$.}
\begin{equation}
X(t) = \sum_{n} X_{n} \cdot {\bf 1}_{\left\{\tau_{n}\leq t < \tau_{n+1}\right\}},
\label{eq1}
\end{equation}
where $\left(X(t)\right)_{t\in\mathbb{R}_{+}}$ is a c\`adl\`ag path (i.e. right-continuous with left limits), and the interval $[\tau_{n},\tau_{n+1})$ is the time interval accommodating the $n^{th}$ hidden state, which takes on a value $X_{n} \in \mathcal{X}$. Every path $\left(X(t)\right)_{t\in\mathbb{R}_{+}}$ on the stochastic basis $(\Omega, \mathcal{F},\{\mathcal{F}_{t}\}_{t\in\mathbb{R}_{+}}, \mathbb{P})$ is a {\it semi-Markov path} (\cite{janssen1984finite, durrett2010probability}), where the {\it sojourn time} of state $n$, which we denote as $S_{n} = \tau_{n+1}-\tau_{n}$, is drawn from a {\it state-specific} distribution $v_j(S_{n} = s\left|\lambda_{j}\right.) = d\mathbb{P}(S_{n} = s\left|X_n = j\right.),$ with $\lambda_{j}$ being a state-specific {\it duration parameter} associated with state $j \in \mathcal{X}$. Unlike ordinary time-homogeneous semi-Markov transitions, in which the transition probabilities among states are assumed to be constant conditioned on there being a transition from the current state (\cite{gillaizeau2015multistate, murphy2002hidden, johnson2013bayesian, yu2010hidden, dewar2012inference, guedon2007exploring}), our model accounts for {\it duration-dependent} semi-Markov transitions. In other words, the transition probability from one state to another depends on the time elapsed in the current state, i.e.
\begin{align}
\mathbb{P}(X_{n+1} = j | X_{n} = i, S_{n} = s) = g_{ij}(s), 
\label{eq2}
\end{align}
where $g_{ij}:\mathbb{R}_{+}\rightarrow [0,1],\, \forall i, j \in \mathcal{X}$ is a {\it transition function} for which $\frac{\partial g_{ij}(s)}{\partial s}$ is well defined, and $\sum_{j=1}^{N} g_{ij}(s) = 1, \forall s \in \mathbb{R}_{+}, i \in \mathcal{X}$.\\ 

Now consider the bivariate (renewal) process $(X_n, S_n)_{n \in \mathbb{N}_{+}}$, which comprises the sequence of states and sojourn times. The semi-Markovian nature of $X(t)$ implies that  $(X_n, S_n)_{n \in \mathbb{N}_{+}}$ satisfies the following condition on its transition probabilities 
\begin{align}
\mathbb{P}(X_{n+1} = j, S_{n} \leq s | \mathcal{F}_{\tau^{-}_{n}}) &= \mathbb{P}(X_{n+1} = j, S_{n} \leq s | X_{n} = i)\nonumber \\
&= \mathbb{P}(X_{n+1} = j| X_{n} = i, S_{n} \leq s) \cdot \mathbb{P}(S_{n} \leq s | X_{n} = i) \nonumber \\
&= \mathbb{E}_{S_{n}}\left[g_{ij}(S_{n})| S_{n} \leq s\right] \cdot V_i(s|\lambda_i) \nonumber \\
&= \bar{g}_{ij}(s) \cdot V_i(s|\lambda_i),
\label{eq3}
\end{align}
where $V_i(.)$ is the cumulative distribution function of state $i$'s sojourn time, and $\bar{g}_{ij}(s)$ is the probability mass function that reflects the probability that a patient's next state being $j$ given that she was at state $i$ and her sojourn time in $i$ is less than (or equal to) $s$. Based on (\ref{eq3}), we define the {\it semi-Markov transition kernel} as a matrix-valued function ${\bf Q}: \mathbb{R}_{+} \rightarrow [0,1]^{N \times N}$ that describes the dynamics of $X(t)$ in continuous time, with entries ${\bf Q}(s) = (Q_{ij}(s))_{i,j \in \mathcal{X}}$ that are given by 
\begin{align}
Q_{ij}(s)= \bar{g}_{ij}(s) \cdot V_i(s|\lambda_i).
\label{eq4}
\end{align}

Since the observable episode for a patient can begin in an arbitrary clinical state (because we only observe the physiological measurements starting from the time when the patients are hospitalized or start taking clinical tests), then it follows that the initial state $X_1$ is random\footnote{We do not consider left-censored observations in this model.}. The initial state distribution is given by
\[{\bf p}^{o} = [p^{o}_1, p^{o}_2,.\,.\,., p^{o}_N]^{T},\]
where $p^{o}_j = \mathbb{P}(X(0) = j),$ and $\sum_{j=1}^{N}p^{o}_j = 1$. \\

The hidden states reflect different levels of clinical risk or severity (e.g. progression stage indexes of a chronic disease or phases of clinical deterioration (\cite{sweeting2010multi};\cite{chen2011non})). In that sense, state $1$ is regarded as the ``least risky state", and state $N$ is regarded as the ``most risky state". We define and interpret states $1$ and $N$ as follows:  
\begin{itemize}
\item {\bf State $1$} is denoted as the {\it safe state}, and represents the state at which the patient is at minimum (or no) risk (e.g. clinically stable post-operative patient, etc).  
\item {\bf State $N$} is denoted as the {\it catastrophic state}, and represents the state at which the patient is at severe risk or encounters an adverse event (e.g. a very severe stage of a chronic disease (\cite{bakal2014heart}), a cardiac or respiratory arrest (\cite{subbe2001validation}), mortality (\cite{knaus1991apache}), etc). \\
\end{itemize} 

\begin{figure}[t]
        \centering
\begin{tikzpicture}
  \SetGraphUnit{3}
  \Vertex{3}
  \WE(3){2}
  \WE(2){1}
	\EA(3){4}
  \EA(4){5}
  \Edge(2)(1)
	\Edge(3)(1)
	\Edge(4)(1)
	\Edge(2)(5)
	\Edge(3)(5)
	\Edge(4)(5)
	\Edge(2)(3)
	\Edge(3)(2)
	\Edge(3)(4)
	\Edge(4)(3)	
  \Edge(2)(4)
	\Edge(4)(2)
  %\Edge[label = 2](2)(3)
  %\Edge[label = 3](3)(2)
  %\Edge[label = 4](2)(1)
  \Loop[dist = 2cm, dir = NO](1.west)
  \Loop[dist = 2cm, dir = SO](5.east)
  %\tikzset{EdgeStyle/.append style = {bend left = 50}}
  %\Edge[label = 7](1)(3)
  %\Edge[label = 8](3)(1)
\end{tikzpicture}
				\captionsetup{font= small}
        \caption{The Markov chain model for a 5-state HASMM.}
\label{Figgx2}
\end{figure}
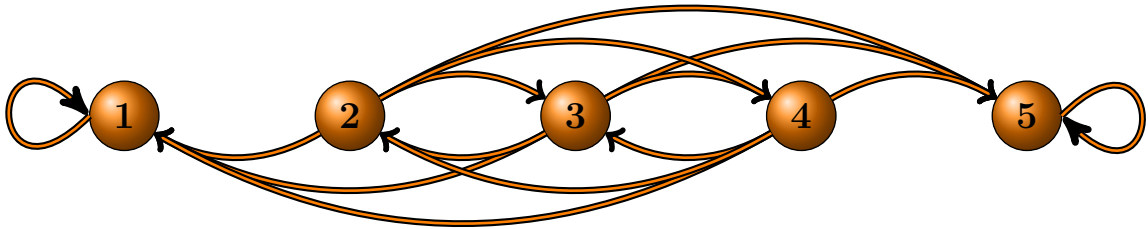

We assume that whenever the system enters either state $1$ or state $N$, it remains there forever\footnote{The model can be easily extended to accommodate an arbitrary number of competing absorbing states.}. Therefore, we model states $\{1,N\}$ as {\it absorbing states}, whereas we model the remaining states in $\mathcal{X}\setminus\{1,N\}$ as {\it transient states} that represent intermediate levels of risk. Following the assumptions in (\cite{ murphy2002hidden, johnson2013bayesian}), we eliminate the self-transitions for all transient states by setting $g_{ii}(s) = 0, Q_{ii}(s) = 0, \forall s \in \mathbb{R}_{+}, i \in \mathcal{X}\setminus\{1,N\},$ whereas we restrict the transitions from states $1$ and $N$ to self-transitions only, i.e. $g_{ii}(s) = 1, i \in \{1,N\}$. Figure \ref{Figgx2} depict the Markov chain for the sequence $\{X_n\}_{n \in \mathbb{N}_{+}}$. \\

%\begin{figure}[t]
%        \centering
%        \includegraphics[width=3.5in]{Figg2.eps}
%				\captionsetup{font= small}
%        \caption{The Markov chain model for a 5-state HASMM.}
%\label{Figgx2}
%\end{figure}

We define $\mathcal{A}_{1}$ as the event that the path $\left(X(t)\right)_{t\in\mathbb{R}_{+}}$ is absorbed in the safe state $1$, i.e. $\mathcal{A}_{1} = \{\lim_{t \rightarrow \infty} X(t) = 1\}$, and $\mathcal{A}_{N}$ as the event that $\left(X(t)\right)_{t\in\mathbb{R}_{+}}$ is absorbed in the catastrophic state $N$, i.e. $\mathcal{A}_{N} = \{\lim_{t \rightarrow \infty} X(t) = N\}$. Since $\left(X(t)\right)_{t\in\mathbb{R}_{+}}$ is an absorbing semi-Markov chain\footnote{We assume that the transition functions $g_{i1}(s)$ and $g_{iN}(s)$ for any transient state $i$ is non-zero for every $s$. Hence, it follows that $\left(X(t)\right)_{t\in\mathbb{R}_{+}}$ is an absorbing semi-Markov chain since it has 2 absorbing states, each of which can be visited starting from any other state (\cite{durrett2010probability}).}, we know that $\mathbb{P}(\mathcal{A}_{1} \vee \mathcal{A}_{N}) = 1,$ and since the events $\mathcal{A}_{1}$ and $\mathcal{A}_{N}$ are mutually exclusive, it follows that $\mathbb{P}(\mathcal{A}_{N}) = 1-\mathbb{P}(\mathcal{A}_{1})$. The quantity $\mathbb{P}(\mathcal{A}_{N})$ describes a patient's prior risk of ending in the catastrophic state, whereas $\mathbb{P}(\mathcal{A}_{N}\left|\mathcal{F}_{t}\right.)$ describes the patient's posterior risk of ending in the catastrophic state having observed its evolution history up to time $t$\footnote{In the clinical applications under consideration, transient states can be ordered by their respective relative risks of encountering event $\mathcal{A}_{N}$ in the subsequent transitions, i.e. in a 5-state chain, it is more likely for the patient to be absorbed in state 5 in the future when it is in state 4 than when it is in state 3. For instance, it is more likely for a patient's chronic obstructive pulmonary disease that is currently assessed to have a severity degree of GOLD1 (mild severity as defined in the GOLD standard \cite{pedersen2011global}) to progress (in the near future) to a severity degree of GOLD2 (moderate) rather than GOLD3 (severe).}. Define $T_{s}$ as an $\mathcal{F}$-stopping time representing the absorption time of the path $\left(X(t)\right)_{t\in\mathbb{R}_{+}}$ in either state $1$ or state $N$, i.e.
\[T_{s} = \inf\{t \in \mathbb{R}_{+}: X(t) \in \{1,N\}\}.\]
% \footnote{Since the sequence $\{X_n\}_n$ is almost surely stopped (i.e. $T_s < \infty$ with probability 1), then the number of transitions exhibited by $\{X_n\}_n$ until the absorption time $T_s$ (i.e. number of jumps in $X(t)$) is almost surely finite.}
Finally, we define $K$ as the (random) number of state realizations in the sequence $\{X_n\}^{K}_{n=1}$ up to the stopping time $T_s$, which has to be concluded by either state $1$ or $N$, e.g. when $\mathcal{X} = 4,$ the sequences $\{1\}, \{4\}, \{2,3,2,3,4\},$ and $\{3,2,1\}$ are valid, random-length realizations of $\{X_n\}^{K}_{n=1}$, and each represents a certain state evolution trajectory for the patient. 

\subsubsection{Observations and Censoring}
The path $\left(X(t)\right)_{t\in\mathbb{R}_{+}}$ is unobservable; what is observable is a corresponding process $\left(Y(t)\right)_{t\in\mathbb{R}_{+}}$ on $(\Omega, \mathcal{F},\{\mathcal{F}_{t}\}_{t\in\mathbb{R}_{+}}, \mathbb{P})$, the values of which are drawn from an {\it observation-space} $\mathcal{Y}$, and whose distributional properties are dependent on the latent states' path $\left(X(t)\right)_{t\in\mathbb{R}_{+}}$. The observable process $\left(Y(t)\right)_{t\in\mathbb{R}_{+}}$ can be put in the form  
\begin{equation}
Y(t) = \sum_{n} Y_{n}(t) \cdot {\bf 1}_{\left\{\tau_{n}\leq t < \tau_{n+1}\right\}},
\label{eq5}
\end{equation}
where $\left(Y(t)\right)_{t\in\mathbb{R}_{+}}$ is a c\`adl\`ag path, comprising a sequence of function-valued variables $\{Y_{n}(t)\}_{n=1}^{K},$ with $Y_{n}:[\tau_{n},\tau_{n+1}) \rightarrow \mathcal{Y}$. Even though the path $\left(Y(t)\right)_{t\in\mathbb{R}_{+}}$ is accessible, only a sequence of irregularly spaced samples of it is observed over time, and is denoted by $\left\{Y(t_m)\right\}_{t_m \in \mathcal{T}}$, where $\mathcal{T} = \{t_1, t_2, .\,.\,., t_M\}$ is the set of observed measurements, and $M$ is the total number of such measurements. We say that the process is censored if $M<\infty$; typical episodes in an EHR are censored: observations stop at some point of time due to a release from care, an ICU admission, mortality, etc. \\

The sampling times in $\mathcal{T}$ represent the times at which a patient with a chronic disease took clinical tests (i.e. time intervals in $\mathcal{T}$ spans years), or the times at which clinicians have gathered vital signs for a monitored critically ill patient in a hospital ward (i.e. time intervals in $\mathcal{T}$ span days or hours). We assume that the sampling times in $\mathcal{T}$ are drawn from a {\it point-process} $\Phi(\zeta) = \sum_{m\in\mathbb{N}_{+}} \delta_{t_{m}},$ which is defined on $(\Omega, \mathcal{F},\{\mathcal{F}_{t}\}_{t\in\mathbb{N}}, \mathbb{P})$, and with $\delta_{t}$ being the Dirac measure. The point process $\Phi(\zeta)$ is parametrized by an intensity parameter $\zeta$, but is assumed to be independent of the latent states path\footnote{This means that the sampling times are uninformative of the latent states, which makes the inference problem more challenging. The HASMM model can be easily extended to incorporate a state-dependent sampling process using a Cox process (\cite{lando1998cox}) or a Hawkes process (\cite{hawkes1974cluster}) to modulate the intensity parameter $\zeta$. A good discussion on conditional intensity models can be found in (\cite{qin2015auxiliary}).}. Define $\mathcal{T}_n$ as the set of $M_n$ samples that are gathered during the interval\footnote{Note that what is observed is a sequence of sampling times $\mathcal{T}$, the elements of which are not labeled by the corresponding state indexes, for that the states are latent, i.e. the sets $\mathcal{T}_n$ are latent.} $[\tau_n,\tau_{n+1})$, i.e. $\mathcal{T}_n = \{t_m: t_m \in \mathcal{T}, t_m \in [\tau_n,\tau_{n+1})\}, M_n = |\mathcal{T}_n|,$ and $\sum_{n}M_n = M$. Since $\mathcal{T}_n$ could possibly be empty ($\mathcal{T}_n = \emptyset$), some states can have no corresponding observations (i.e. an inpatient may exhibit a transition to a deteriorating state during the night, even though her blood pressure were not measured during the night. Recall the illustration in Figure \ref{Fiq2}). \\

The paths $\{Y_{n}(t)\}^{K}_{n=1}$ are assumed to be conditionally independent given the hidden sequence of states $\{X_{n}\}^{K}_{n=1}$, and hence we have that 
\[\{Y(t_m)\}_{t_m \in \mathcal{T}_{n}} \ci \{Y(t_m)\}_{t_m \in \mathcal{T}_{n+1}} \left| \, X_{n}, X_{n+1}\right., \forall n \in \{1,2,.\,.\,., K-1\}.\]
The observed samples generated under every state $X_n$ and sampled at the times in $\mathcal{T}_{n}$ are drawn from $\mathcal{Y}$ according to a distribution $\mathbb{P}(\{Y(t_m)\}_{t_m \in \mathcal{T}_{n}}\left|X_{n} = j, \Theta_j\right.),$ where $\Theta_j$ is an {\it emission parameter} that controls the distributional properties of the observations generated under state $j$.\\ 
   
The number of observation samples is finite: the observed sequence is {\it censored} at some point of time, which we call the censoring time $T_c$, after which no more observation samples are available. Censoring reflects an external intervention/event that terminated the observation sequence, i.e. death, intensive care unit (ICU) admission, etc. Censoring is {\it informative} (\cite{scharfstein2002estimation};\cite{huang2002frailty};\cite{link1989model}), because the censoring time is correlated with the absorption time $T_s$, and $T_s$ strictly precedes $T_c$ (in an almost sure sense). That is, $T_c$ is an $\mathcal{F}$-stopping time that is given by $T_c = T_s + S_K$, i.e. once the patient enters state $1$ or state $N$, the observations stop after the patient's sojourn time in that state (i.e. observations stop after a time $S_K$ from the entrance in the absorbing state). Therefore, the duration distributions $v_1(s|\lambda_1)$ and $v_N(s|\lambda_N)$ of states $1$ and $N$ are used to determine the censoring times conditioned on the chain $\{X_n\}_{n=1}^{K}$ being absorbed at time $T_s$.\\   

Every sample from the HASMM is an episode comprising a random-length sequence of hidden states $\{X_n\}^{K}_{n=1}$, and a random-length sequence of observations $\{Y(t_{m})\}^{M}_{m=1}$ together with the associated observation times. We only observe $\{Y(t_{m})\}^{M}_{m=1}$; the path of latent states $X(t)$, the number of realized states $K$, the association between observations and states (i.e. the sets $\mathcal{T}_n$) are all unobserved, which makes the inference problem very challenging, but captures the realistic EHR data format and the associated inferential hurdles. In the next subsection, we specify the model's generative process and present an algorithm to generate episodic samples from an HASMM.

\subsection{Model Specification and Generative Process} 
\label{sec2.2}
As have been discussed in Subsection \ref{sec2.1}, the hidden and observables variables of an HASMM can be listed as follows:
\begin{itemize}
\item {\bf Hidden variables:} The hidden states sequence $\{X_{n}\}^{K}_{n=1}$ and the states' sojourn times $\{S_{n}\}^{K}_{n=1}$ (or equivalently, the transition times $\{\tau_{n}\}^{K}_{n=1}$). 
\item {\bf Observable variables:} The observed episode $\left\{Y(t_m)\right\}_{m=1}^{M}$ and the associated sampling times $\mathcal{T} = \left\{t_m\right\}_{m=1}^{M}$. 
\end{itemize}
The HASMM model parameters that generate both the hidden and observable variables are encompassed in the parameter set $\Gamma,$ i.e. 
\[\Gamma = \left(\underbrace{N}_{\tiny \mbox{State cardinality}}, \underbrace{{\bf \lambda} = \{\lambda_j\}_{j=1}^{N}}_{\tiny \mbox{State duration}}, \underbrace{{\bf p}^{o}}_{\tiny \mbox{Initial states}}, \underbrace{{\bf Q} = \{Q_{ij}(s)\}_{i,j=1}^{N}}_{\tiny \mbox{Transitions}}, \underbrace{{\bf \Theta} = \{\Theta_j\}_{j=1}^{N}}_{\tiny \mbox{Emission}},\underbrace{\zeta}_{\tiny \mbox{Sampling}}\right).\] 
Since the point process $\Phi(\zeta)$ does not reveal any information about the latent states, and hence plays no role in inference, we will drop it from the parameter set $\Gamma$ in the rest of the paper. In the following, we specify the distributional properties for both the hidden and observable variables.

\subsubsection{Distributional specifications for the hidden variables}
We model the state sojourn time of each state $i \in \mathcal{X}$ via a Gamma distribution. The selection of a Gamma distribution ensures that the generative process encompasses ordinary continuous-time Markov models for the path $(X(t))_{t \in \mathbb{R}_{+}}$, since the exponential distribution\footnote{Note that a semi-Markov chain reduces to a Markov chain if the sojourn times are exponentially distributed.} is a special case of the Gamma distribution (\cite{durrett2010probability}). Thus, if the underlying physiology of the patient is naturally characterized by memoryless state transitions, this will be automatically learned from the data via the parameters of the Gamma distribution. The sojourn time distribution for state $i$ is given by  
\[v_{i}(s|\lambda_i = \{\lambda_{i,s}, \lambda_{i,r}\}) = \frac{1}{\Gamma(\lambda_{i,s})} \cdot \lambda_{i,r}^{\lambda_{i,s}} \cdot s^{\lambda_{i,s}} \cdot e^{-s \cdot \lambda_{i,r}}, s \geq 0,\]
where $\lambda_{i,s}>0$ and $\lambda_{i,r}>0$ are the shape and rate parameters of the Gamma distribution respectively.\\  
\\
\begin{figure}[!tbp]
  \centering
  \begin{minipage}[b]{0.4\textwidth}
\includegraphics[width=2.75in]{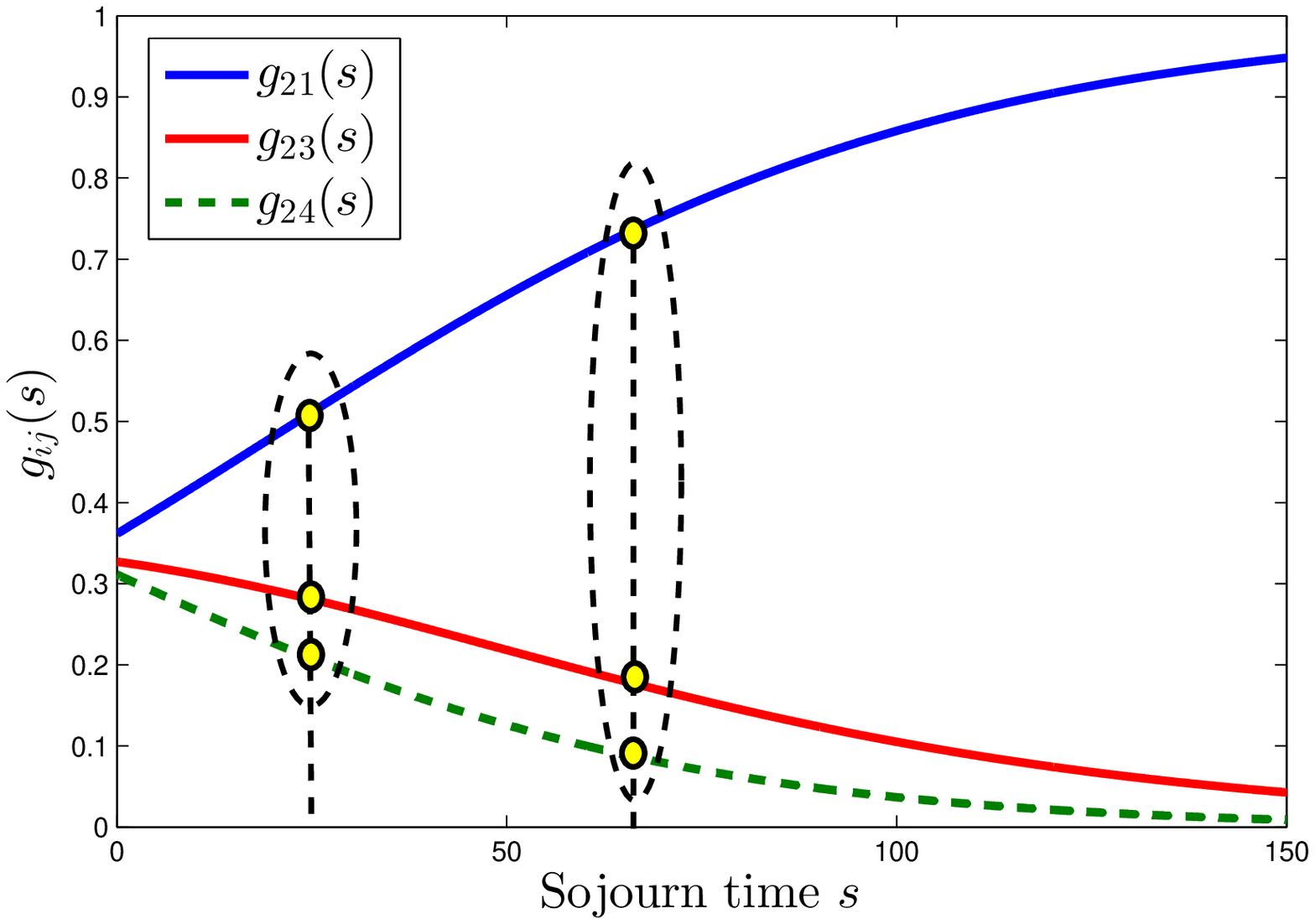}
				\captionsetup{font= small}
        \caption{Exemplary transition functions $(g_{2j})_{j=1}^{4}$ for a 4-state HASMM.}
	\label{FigQ}			
  \end{minipage}
  \hfill
  \begin{minipage}[b]{0.5\textwidth}
\includegraphics[width=3.25in]{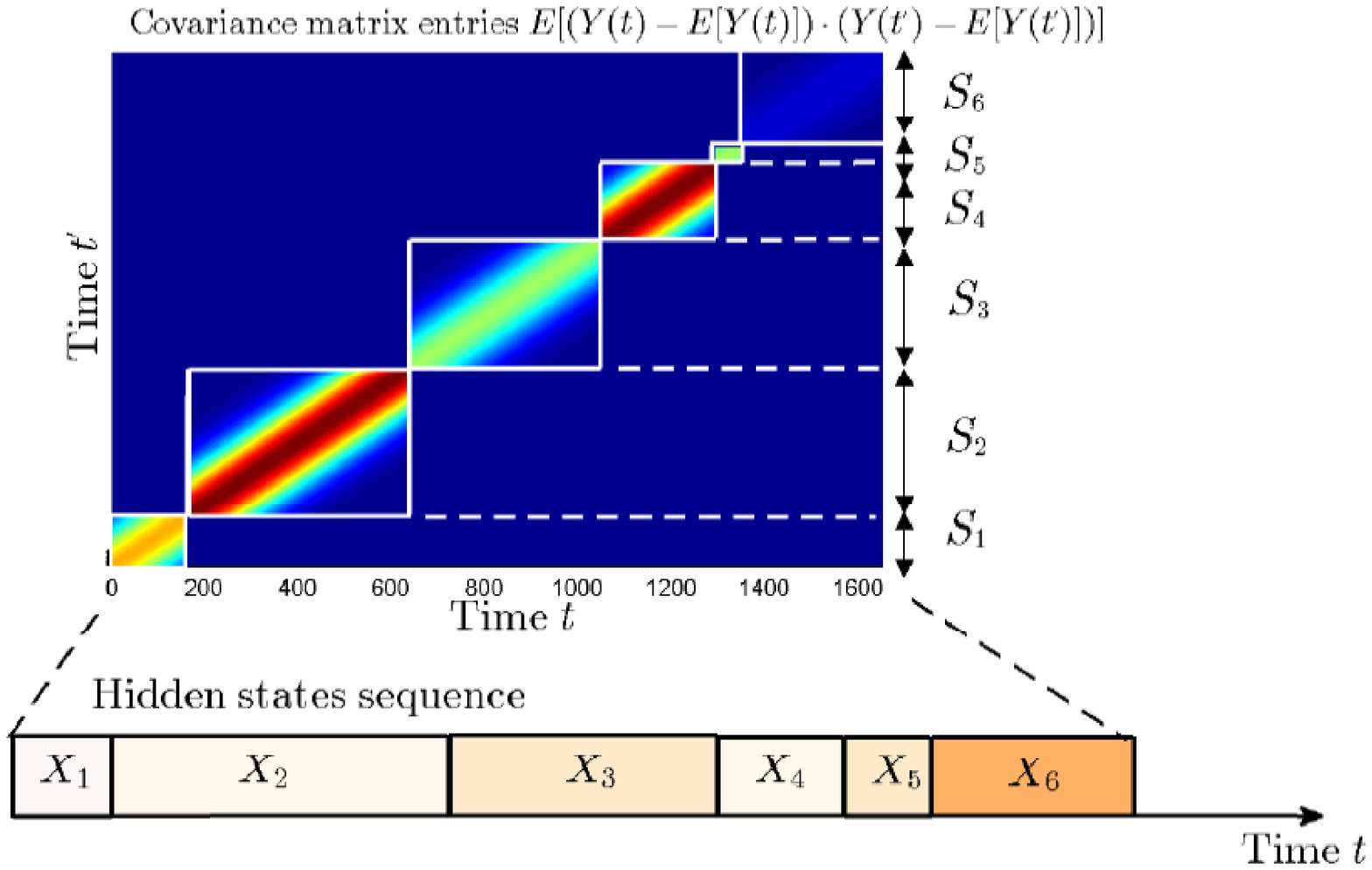}
				\captionsetup{font= small}
        \caption{Depiction for the correlation structure of the observable variables for an underlying state sequence $\{X_n\}^{6}_{n=1}$.}		
	\label{FigQ2}
  \end{minipage}
\end{figure}
Now we specify the structure of the transition kernel ${\bf Q}(s) = (Q_{ij}(s))_{i,j}, i, j \in \mathcal{X}$. Recall from (\ref{eq4}) that the each element in the transition kernel matrix can be written as $\mathbb{E}_{S}\left[g_{ij}(S)| S \leq s\right] \cdot V_i(s|\lambda_i)$. Having specified the distribution $v_i(s|\lambda_i)$ as a Gamma distribution, it remains to specify the function $g_{ij}(s)$ in order to construct the elements of ${\bf Q}(s)$. The transition functions $(g_{ij}(s))_{i,j}$ are given by {\it multinomial logistic} functions as follows
\begin{align} 
g_{ij}(s) &= \frac{e^{(\eta_{ij}+\beta_{ij} \cdot s)}}{\sum_{k=1}^{N} e^{(\eta_{ik}+\beta_{ik} \cdot s)}}, \forall i\neq j, i \notin \{1,N\} \nonumber \\
g_{ii}(s) &= 0, \forall i \in \{2,.\,.\,.\,,N-1\}, \nonumber \\
g_{ii}(s) &= 1, \forall i \in \{1,N\}, 
\label{eq6}
\end{align}
where $\eta_{ij}, \beta_{ij} \in \mathbb{R}_{+}$. The parameters $(\eta_{ij})^{N}_{j=1}$ determine the baseline values for the transition probability mass from state $i$ to state $j$, i.e. $g_{ij}(0)$, whereas the parameters $\beta_{ij}$ controls the dependence of the transition probability mass on the sojourn time\footnote{Similar effects for the sojourn time on the transition probabilities has been demonstrated in the progression of breast cancer from healthy to preclinical states in (\cite{taghipour2013parameter}), where age (the main risk factor for breast cancer) was shown to affect the probability of progressing across the states of healthy to preclinical, clinical and death. These effects may be also prevailing in other diseases, or in critical care settings where the length of time during which a patient stays clinically stable may imply that the patient is more likely to transit to a more healthy state in the future. Through the HASMM model, we can recognize whether or not this effect is evident in the EHR data, i.e. whether the transition function reflects an underlying homogeneous (if $g_{ij}(s)$ is independent of $s$) or duration-dependent transitions by learning the parameter $\beta_{ij}$. Moreover, the parameter $\beta_{ij}$ is defined per state; the HASMM model can capture scenarios where transitions are duration-independent from some states, but are duration-dependent from others.}. If $\beta_{ij} = 0,$ then we have that $g_{ij}(s) = g_{ij}(0) = \frac{e^{\eta_{ij}}}{\sum_{k=1}^{N} e^{\eta_{ik}}}, \forall s \in \mathbb{R}_{+},$ i.e. the transition probability out of state $i$ remains constant irrespective of the sojourn time in that state. In the limit when $s$ goes to infinity, the parameter $\beta_{ij}$ dominates the functional form in (\ref{eq6}). Figure \ref{FigQ} depicts exemplary transition functions $(g_{ij}(s))_{i,j}$ for a 4-state HASMM.\\ 

\begin{figure*}[t!]
        \centering
        \includegraphics[width=5.5in]{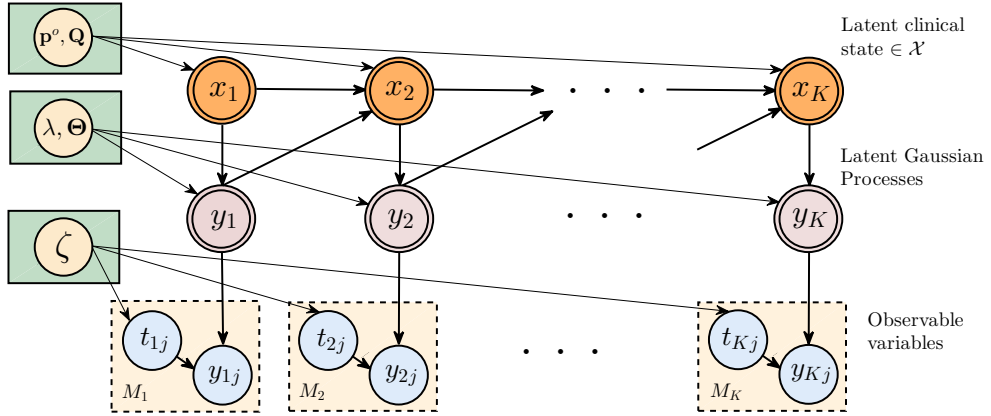}
				\captionsetup{font= small}
        \caption{A basic graphical model for the HASMM.}
\label{FigQ4xs}
\end{figure*}

\subsubsection{Distributional specifications for the observable variables}
As explained in Subsection \ref{sec2.1}, the observable process $Y(t)$ can be decomposed as $Y(t) = \sum^{K}_{n=1} Y_{n}(t) \cdot {\bf 1}_{\left\{\tau_{n}\leq t < \tau_{n+1}\right\}}$, where the paths $(Y_n(t))_{n=1}^{K}$ are conditionally independent given the state sequence $\{X_n\}_{n=1}^{K}$. Since observations are drawn from $Y(t)$ at arbitrarily, and irregularly spaced time instances $\mathcal{T}$, we have to model the distributional properties of $Y(t)$ in continuous time. We model every path $Y_n(t)$ defined over $[\tau_{n}, \tau_{n+1})$ as a segment drawn from a multi-task Gaussian Process (GP), with a hyper-parameter set $\Theta_{i}$ that depends on the corresponding latent state $X_n = i$ (\cite{rasmussen2006gaussian, bonilla2007multi}). The input to the multi-task GP is the time variable and the output is the set of physiological variables at a certain point of time. The GP associated with every state $X_n = i$ is parametrized by a constant mean function $m_{i}(t) = m_i$, a {\it squared-exponential} covariance kernel $k_{i}(t,t^{'}) = \sigma_{i}^{2} \, e^{-\frac{1}{2\ell_i^{2}}\,||t-t^{'}||^{2}}$, and a ``free-form" covariance matrix $\Sigma_i$ between the different physiological measurements (\cite{bonilla2007multi}). Thus, for a $Q$-dimensional physiological stream $Y(t) = (Y^1(t),.\,.\,., Y^Q(t))$, the observations for state $i$ are generated as follows 
\[\left<Y^l_i(t) \cdot Y^v_i(t^{'})\right> = \Sigma_i(l,v) \,\cdot\, k_i(t, t^{'}) \,\,\,\,\,\,\,\,\,\,\,\,\,\, \{Y^l_i(t)\}_{t \in \mathcal{T}, 1\leq l \leq Q} \sim \mathcal{N}(m_i(t), {\bf \Sigma}_i),\]
where ${\bf \Sigma}_i(l,v,t,t^{'}) = \left<Y^l_i(t) \cdot Y^v_i(t^{'})\right>$. The GP hyper-parameters associated with state $i$ are given by $\Theta_i = (m_i, \sigma_{i}, \Sigma_i, \ell_i)$, i.e. $Y_n(t)|X_n = i \sim \mathcal{GP}(\Theta_i)$.\\
\\
We note that the HASMM model is a {\it segment model} (\cite{ostendorf1996hmm,murphy2002hidden, yu2010hidden, guedon2007exploring}), i.e. observation samples that are defined within the sojourn time of the same state are correlated, but observation samples in different states are independent. The model can also be viewed as a state-switching model, but for which the transition dynamics do not need to be linear as in (\cite{georgatzis2016input,fox2011bayesian}), but rather depend on the covariance kernel $k_{i}(t,t^{'})$. Figure \ref{FigQ2} depicts the correlation structure of the observable variables in terms of the covariance matrix of a discrete version of $Y(t)$ generated under a specific hidden state sequence. We can see that conditioned on the hidden state sequence, the covariance matrix is a block diagonal matrix, where the sizes of the blocks are random and are determined by the hidden states' sojourn times. \\
\\
The sampling times in $\mathcal{T}$ are generated by the point process $\Phi(\zeta)$, which for the sake of completeness of the model description, we specify as a Poisson process with an intensity parameter $\zeta$. Note though that since we assume the sampling times are uninformative of the latent states path $X(t)$, the distributional specification of $\Phi(\zeta)$ is ancillary to the inference and learning algorithms developed in Sections \ref{sec3} and \ref{sec3.3}.  

\begin{figure*}[t]
        \centering
        \includegraphics[width=5.75in]{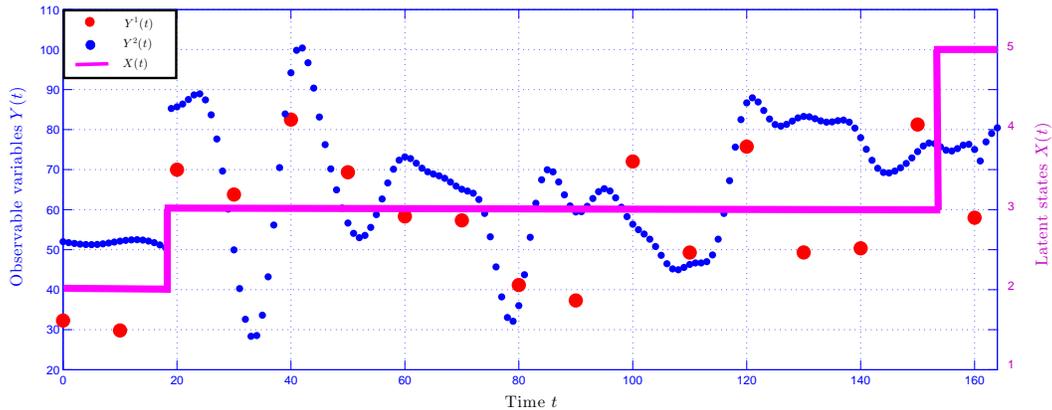}
				\captionsetup{font= small}
        \caption{An episode generated by \texttt{GenerateHASMM($\Gamma$)} with $N=5$. The realized hidden state sequence (upper) is $\{2,3,5\}$, and is absorbed in state 5. The physiological stream $(Y^1(t),Y^2(t))$ is 2-dimensional and stream $Y^2(t)$ is sampled more intensely than $Y^1(t)$.}
\label{FigQ3}
\end{figure*}

\subsubsection{Sampling episodes from an HASMM}
We conclude this Section by presenting an algorithm for sampling episodes from an HASMM with a hyper-parameter set $\Gamma$. Algorithm \ref{alg1} (\texttt{GenerateHASMM($\Gamma$)})\footnote{Our HASMM toolbox for Matlab is available at https://github.com/ahmedmalaa/HASMMtoolbox.} samples a patient's episodes by first sampling an initial state from $\mathcal{X},$ and then sequentially samples sojourn times $s$ from the Gamma distribution, and new states using the semi-Makrov kernel ${\bf Q}(s)$, until an absorbing state is drawn. Figure \ref{FigQ3} depicts an episode sampled via Algorithm \ref{alg1}.     
\begin{algorithm}
  \caption{Sampling episodes from an HASMM}\label{alg1}
  \begin{algorithmic}[1]
    \Procedure{\texttt{GenerateHASMM}}{$\Gamma$}
		  \State {\bfseries Input:} HASMM model parameters $\Gamma = (N, {\bf \lambda}, {\bf p}^{o}, {\bf Q}(s), {\bf \Theta}, \zeta)$
	    \State {\bfseries Output:} An episode $(\{X_n\}_{n=1}^{K},\{\tau_n\}_{n=1}^{K}, \{Y(t_m)\}_{m=1}^{M}, \{t_m\}_{m=1}^{M})$
      \State $\tau_1 \gets 0$, $k \gets 1$, $\mathcal{T} \sim \mbox{Poisson}(\zeta)$  \Comment{Initializations}
			\State $x_1 \sim \mbox{Multinomial}(p^{o}_1,p^{o}_2,.\,.\,.,p^{o}_N)$ \Comment{Sample an initial latent state}
			\State $s_1 \sim \mbox{Gamma}(\lambda_{x_1,s}, \lambda_{x_1,r}),$ $\tau_2 \gets \tau_1 + s_1$
			\State $\mathcal{T}_{1} = \{t \in \mathcal{T}: \tau_{1} \leq t \leq \tau_{2}\}$
      \While{$x_k \notin \{1,N\}$} \Comment{Sample latent states until absorption}
				\State $x_{k+1} \sim \mbox{Multinomial}(g_{x_{k}1}(s_{k}),g_{x_{k}2}(s_{k}),.\,.\,.,g_{x_{k}N}(s_{k}))$
        \State $s_{k+1} \sim \mbox{Gamma}\left(\lambda_{x_{k+1},s}, \lambda_{x_{k+1},r}\right),$ $\tau_{k+2} \gets \tau_{k+1} + s_{k+1}$
				\State $\mathcal{T}_{k+1} = \{t \in \mathcal{T}: \tau_{k+1} \leq t \leq \tau_{k+2}\}$
				\State $\{y(t_m)\}_{t_m \in \mathcal{T}_{k+1}} \sim \mathcal{GP}(\Theta_{x_{k+1}})$ \Comment{Sample observations from a Gaussian Process}
				\State $k \gets k+1$
      \EndWhile\label{euclidendwhile}
      \State \textbf{return} $(\{x_n\}_{n=1}^{K},\{\tau_n\}_{n=1}^{K}, \{y(t_m)\}_{m=1}^{M}, \{t_m\}_{m=1}^{M})$
    \EndProcedure
  \end{algorithmic}
\end{algorithm}

\section{Inference in Hidden Absorbing Semi-Markov Models}
\label{sec3}
In this Section, we develop an online algorithm that carries out diagnostic and prognostic inferences for a monitored patient's episode in real-time. Given an ongoing realization of an episode $\{y(t_1),y(t_2),.\,.\,.,y(t_m)\}$ at time $t_m$ (before the censoring time $T_c$), and the HASMM model parameter $\Gamma$ that has generated this realization (i.e. $\{y(t_1),y(t_2),.\,.\,.,y(t_m)\}$ is sampled via the algorithm \texttt{GenerateHASMM($\Gamma$)}), we aim at carrying out the following inference tasks:
\begin{itemize}
\item {\bf Diagnosis:} Infer the patient's current clinical state, i.e. compute
\[\mathbb{P}(X(t_m) = j \left|\, Y(t_1) = y(t_1),.\,.\,., Y(t_m) = y(t_m), \Gamma\right.), \, \forall j \in \mathcal{X}.\]
\item {\bf Prognostic Risk Scoring:} Compute the patient's risk of absorption in the catastrophic state, i.e. 
\[\mathbb{P}(\mathcal{A}_{N} \left|\, Y(t_1) = y(t_1),.\,.\,., Y(t_m) = y(t_m), \Gamma\right.).\]
\end{itemize}
In the rest of this Section, we drop the conditioning on $\Gamma$ for notational brevity. The first inference task corresponds to disease severity estimation for patients with chronic disease, or clinical acuity assessment for critical care patients. The second task corresponds to risk scoring for future adverse events for patients who have been monitored for some period of time, i.e. the risk of developing a future preclinical or clinical breast cancer state (\cite{gail2010comparing}), the risk of clinical deterioration for post-operative patients in wards (\cite{rothman2013development}), the risk of mortality for ICU patients (\cite{knaus1985apache}), etc.    

\begin{figure*}[t!]
        \centering
				\includegraphics[width=5.5in]{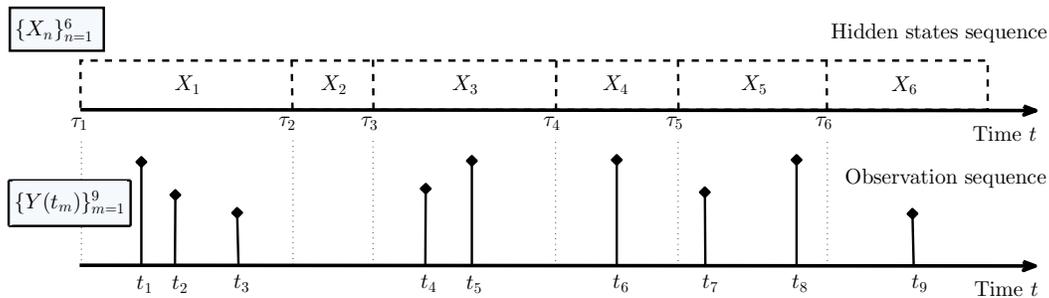}
				\captionsetup{font= small}
        \caption{An exemplary HASMM episode with 6 hidden state realizations and 9 observed samples.}
\label{FigQ4}
\end{figure*}

\subsection{Challenges facing the HASMM Inference Tasks}
\label{sec3.2}
The inference tasks discussed in the previous Subsection are confronted with 3 main challenges --listed hereunder-- that hinder the direct deployment of classical forward-backward message-passing routines.
\begin{enumerate}
\item In addition to the clinical states $\{X_n\}^{K}_{n=1}$ being unobserved, the transition times among the states, $\{\tau_n\}^{K}_{n=1}$, are also unobserved (i.e. we do not know the time at which the patient's state changed). Thus, unlike the discrete-time models in (\cite{murphy2002hidden, johnson2013bayesian, yu2010hidden, dewar2012inference, guedon2007exploring}), in which we know that the underlying states switch sequentially in a (known) one-to-one correspondence with the observations, in an HASMM the association between states and observations is unknown. Figure \ref{FigQ4} depicts an exemplary HASMM episode with 6 realized states and 9 observations samples; in this realization, the association between the observations $\{Y(t_1), Y(t_2), Y(t_3)\}$ and state $X_1$ is hidden. The importance of reasoning about the hidden transition times is magnified by the duration-dependence of the transition probabilities that govern the sequence $\{X_n\}^{K}_{n=1}$. 
\item Since observations are made at random and arbitrary time instances, some transitions may not be associated with any evidential data. That is, as it is the case for state $X_2$ in Figure \ref{FigQ4}, there is no guarantee that for every state $X_n$, an observation is drawn during its occupancy, i.e. $[\tau_n,\tau_{n+1})$. In a practical setting, the inference algorithm should be able to reason about the state trajectories even in silence periods that come with no observations (recall the example in Figure \ref{Fiq2} where observations of a critical care patient's systolic blood pressure stop for an entire day). Hence, one cannot directly discretize the time variable and use the discrete-time HMM inference algorithms (e.g. the algorithms in (\cite{rabiner1989tutorial})) since in that case we would exhibit time steps that come with no associated observations, and with potential state transitions. 
\item The HASMM model assumes that observations that belong to the same state are correlated (e.g. in Figure \ref{FigQ4}, each of the subset of observations $\{Y(t_1),Y(t_2),Y(t_3)\}$, $\{Y(t_4),Y(t_5)\}$ and $\{Y(t_7),Y(t_8)\}$ are not drawn independently conditioned on the latent state since they are sampled from a GP), thus we cannot use the variable-duration and explicit-duration HSMM inference algorithms in (\cite{murphy2002hidden, johnson2013bayesian, yu2010hidden,guedon2007exploring}), as those assume that all observations are conditionally independent given the latent states. Our model is closer to a segment-HSMM model (\cite{yu2010hidden,guedon2007exploring}), but with irregularly spaced observations and an underlying duration-dependent state evolution process, which requires a different construction of the forward messages.
\end{enumerate}
In the following Subsection, we develop a forward filtering algorithm that deal with episodes generated from an HASMM and address the above challenges.\\
\subsection{The HASMM Forward Filtering Algorithm}
Given a realization of an episode $\{y(t_1),y(t_2),.\,.\,.,y(t_m)\}$ at time $t_m$, the posterior probability of the patient's current clinical state $X(t_m)$ is given by\footnote{We use the notation $d\mathbb{P}$ to denote a probability density defined with respect to $(\Omega, \mathcal{F}, \mathbb{P})$.}
\begin{align}
\mathbb{P}(X(t_m) = j \left|\, y(t_1),.\,.\,., y(t_m), \mathcal{T}\right.) &= \frac{d\mathbb{P}(X(t_m) = j, y(t_1),.\,.\,., y(t_m)\,|\, \mathcal{T})}{d\mathbb{P}(y(t_1),.\,.\,., y(t_m)|\mathcal{T})} \nonumber \\
 &= \frac{d\mathbb{P}(X(t_m) = j, y(t_1),.\,.\,., y(t_m)\,|\,\mathcal{T})}{\sum_{j=1}^{N} d\mathbb{P}(X(t_m) = j, y(t_1),.\,.\,., y(t_m)\,|\,\mathcal{T})}.
\label{eqqInf}
\end{align}
The above application of Bayes' rule implies that, given the observation times $\mathcal{T}$, computing the joint probability density $d\mathbb{P}(X(t_m) = j, y(t_1),.\,.\,., y(t_m)\,|\,\mathcal{T})$ suffices for computing the posterior probability of the patient's clinical states. As it is the case for the conventional HMM setting, we denote these joint probabilities as the {\it forward messages} $\alpha_{m}(j \,|\,\mathcal{T}) = d\mathbb{P}(X(t_m) = j, y(t_1),.\,.\,., y(t_m)\,|\,\mathcal{T})$. \\ 

Since the HASMM is a segment model, the conventional notion of the forward messages $\alpha_{m}(j\,|\,\mathcal{T})$ does not suffice for constructing the forward filtering algorithm since we need to account for the latent correlation structures between the (conditionally-dependent) observations (\cite{murphy2002hidden}). To that end, we define $\alpha_m(j, w \,|\,\mathcal{T})$ as the forward message for the $j^{th}$ state at the $m^{th}$ observation time (i.e. $t_m$) {\it with a lag $w$} as follows
\begin{align}
\alpha_m(j, w\,|\,\mathcal{T}) = d\mathbb{P}(X(t_m) = j, \{t_u\}_{u=m-w+1}^{m} \in \mathcal{T}_n, t_{m-w} \in \mathcal{T}_{n^{'}}, \{y(t_u)\}_{u=1}^{m}\,|\,\mathcal{T}),
\label{eqqInf333}
\end{align}
for some $n, n^{'} \in \mathbb{N}_{+}$, and $n \neq n^{'}$. That is, the forward message $\alpha_m(j, w\,|\,\mathcal{T})$ is simply the joint probability that the current state is $j$, that the associated observations are $(y(t_1),.\,.\,., y(t_m)),$ and that the current state has lasted for the last $w$ measurements. For notational brevity, denote the event $\left\{\{t_u\}_{u=m-w+1}^{m} \in \mathcal{T}_n, t_{m-w} \in \mathcal{T}_{n^{'}}\right\}$ as $\psi(m, w)$. Thus, $\alpha_m(j, w\,|\,\mathcal{T})$ can be written as
\begin{align} 
\alpha_m(j, w\,|\,\mathcal{T}) = \sum_{i=1}^{N}\sum_{w^{'}=1}^{m-w} d\mathbb{P}(X(t_m) = j, \psi(m, w), X(t_{m-w}) = i, \psi(m-w, w^{'}), \{y(t_u)\}_{u=1}^{m}\,|\,\mathcal{T}),\nonumber 
\end{align}
which can be decomposed using the conditional independence properties of the states, observable variables and sojourn times as follows
\[d\mathbb{P}(X(t_m) = j, \psi(m, w), X(t_{m-w}) = i, \psi(m-w, w^{'}), \{y(t_u)\}_{u=1}^{m}) = \]
\begin{align}
& d\mathbb{P}(\{y(t_u)\}_{u=m-w+1}^{m}\, | \, X(t_m) = j, \psi(m, w)) \, \times \, \underbrace{\mathbb{P}(X(t_m) = j \, | \, X(t_{m-w}) = i, \psi(m-w, w^{'}))}_{p_{ij}(t_m-t_{m-w}, \psi(m-w, w^{'}))} \, \times \, \nonumber \\
& \underbrace{\mathbb{P}(\psi(m, w) \, | \, X(t_m) = j)}_{V_j(t_m-t_{m-w}|\lambda_j)-V_j(t_m-t_{m-w+1}|\lambda_j)}  \, \times \, \underbrace{d\mathbb{P}(X(t_{m-w}) = i, \psi(m-w, w^{'}), \{y(t_u)\}_{u=1}^{m-w})}_{\alpha_{m-w}(i, w^{'})},
\label{eqqInf3}
\end{align}
where we have dropped the conditioning on $\mathcal{T}$ for notational brevity. The first term, $d\mathbb{P}(\{y(t_u)\}_{u=m-w+1}^{m}\, | \, X(t_m) = j, \psi(m, w))$, is the probability density of the observable variables in $\{y(t_u)\}_{u=m-w+1}^{m}$ conditioned on the hidden state being $X(t_m) = j$ and that the time instances $\{t_u\}_{u=m-w+1}^{m}$ reside in the sojourn time of $X(t_m) = j$. The second term, $p_{ij}(t_m-t_{m-w}, \psi(m-w, w^{'}))$, is the {\it interval transition probability}, i.e. the probability that the hidden state sequence transits to state $j$ after a period of $t_m-t_{m-w}$, given that its sojourn time in state $X(t_{m-w}) = i$ at time $t_{m}$ is at least $t_{m}-t_{m-w+1}$, and at most $t_{m}-t_{m-w-w^{'}}$. The third term is the probability that the sojourn time in state $X(t_m) = j$ is between $t_{m}-t_{m-w+1}$ and $t_{m}-t_{m-w}$, whereas the fourth term, $\alpha_{m-w}(i, w^{'})$, is the $(m-w)^{th}$ forward message with a lag of $w^{'}$. Thus, we can write the $m^{th}$ forward message with a lag $w$ as follows 
\[\alpha_m(j, w) = d\mathbb{P}(\{y(t_u)\}_{u=m-w+1}^{m}\, | \, X(t_m) = j) \times\]
\begin{align} 
\sum_{i=1}^{N}\sum_{w^{'}=1}^{m-w} p_{ij}(t_m-t_{m-w}, \psi(m-w, w^{'})) \cdot \left(V_j(t_m-t_{m-w}|\lambda_j)-V_j(t_m-t_{m-w+1}|\lambda_j)\right) \cdot \alpha_{m-w}(i, w^{'}). 
\label{eqqInf4}
\end{align}
As we can see in (\ref{eqqInf4}), one can express $\alpha_m(j, w)$ using a recursive formula that makes use of the older forward messages $\{\alpha_{m-w}(i, w^{'})\}_{w=1}^{m},$ where $\alpha_{o}(i, w^{'}) = 0$, which allows for an efficient dynamic programming algorithm to infer the patient's clinical state in real-time; this is important in critical care settings where prompt risk assessments are crucial for timely clinical intervention.\\ 

The construction of the forward messages in (\ref{eqqInf4}) parallels the structure of forward message-passing in segment-HSMM (See Section 1.2 in (\cite{murphy2002hidden}) and Section 4.2.2 in (\cite{yu2010hidden})), but with the following differences. In (\ref{eqqInf4}), the time interval between every two observation samples is irregular, which reflects in the correlation between the observations in $\{y(t_u)\}_{u=m-w+1}^{m}$ (depends on the covariance kernel of the GP, and the probability of the current latent state's sojourn time being encompassing the most recent $w$ samples, i.e. $\left(V_j(t_m-t_{m-w}|\lambda_j)-V_j(t_m-t_{m-w+1}|\lambda_j)\right)$. However, the most challenging ingredient of the forward message is the interval transition probability $p_{ij}(t_m-t_{m-w}, \psi(m-w, w^{'}))$. This is because unlike the discrete-time HSMM models in (\cite{murphy2002hidden, yu2010hidden}), which exhibit transitions only at discrete time steps that are always accompanied with evidential observations, i.e. no hidden transitions can occur between observation samples, and the transitions among hidden states are duration-independent, in an HASMM, transitions can occur at arbitrary time instances, multiple transitions can occur between two observation samples, and transitions are duration-dependent.\\

In order to evaluate the term $p_{ij}(t_m-t_{m-w}, \psi(m-w, w^{'}))$, we construct a virtual (discrete-time) trivariate {\it embedded Markov chain} $\{X(t_m),  t_{m-w}, t_{m-w+1}\}$, the transition probabilities of which are equal to the interval transition probabilities. In the recent work in (\cite{liu2015efficient}), a similar embedded Markov chain analysis was conducted for a CT-HMM, but for which the underlying state evolution process was assumed to be a duration-independent ordinary Markov chain for which the expressions for the interval transition probabilities are readily available by virtue of the exponential distributions of the memoryless state sojourn times.\\

Recall from Subsection \ref{sec2.1.1} that the semi-Markov kernel of the hidden state sequence $\{X_n\}_{n=1}^{K}$ is defined as $Q_{ij}(\tau) = \mathbb{P}(X_{n+1} = j, S_n \leq \tau | X_{n} = i),$ i.e. the probability that the sequence transits from state $i$ to state $j$ given that the sojourn time in $i$ is less than or equal to $\tau$. Theorem \ref{thm1} establishes the methodology for computing the interval transition probabilities $p_{ij}(t_m-t_{m-w}, \psi(m-w, w^{'}))$ using the parameters of an HASMM. In Theorem \ref{thm1}, we define ${\bf \tilde{P}}(\tau, \munderbar{s}, \bar{s})$ as a matrix-valued function ${\bf \tilde{P}}: \mathcal{S} \rightarrow [0,1]^{N \times N},\, \mathcal{S} = \left\{(\tau,\munderbar{s}, \bar{s}): \tau \in \mathbb{R}_{+}, \bar{s} \in \mathbb{R}_{+},  \munderbar{s} \leq \bar{s}\right\},$ the entries of which are given by
\begin{align} 
{\bf \tilde{P}}(\tau, \munderbar{s}, \bar{s}) = \underbrace{\left[
\begin{array}{c|c|c|c}
\tilde{p}_{11}(\tau, \munderbar{s}, \bar{s}) & \tilde{p}_{21}(\tau, \munderbar{s}, \bar{s}) & \cdots & \tilde{p}_{N1}(\tau, \munderbar{s}, \bar{s})\\ 
\tilde{p}_{12}(\tau, \munderbar{s}, \bar{s}) & \tilde{p}_{22}(\tau, \munderbar{s}, \bar{s}) & \cdots & \tilde{p}_{N2}(\tau, \munderbar{s}, \bar{s})\\ 
\vdots & \vdots &  & \vdots \\
\tilde{p}_{1N}(\tau, \munderbar{s}, \bar{s}) & \tilde{p}_{2N}(\tau, \munderbar{s}, \bar{s}) & \cdots & \tilde{p}_{NN}(\tau, \munderbar{s}, \bar{s}) 
\end{array}\right]}_{\small \mbox{Size}\, N \times N \, \small \mbox{matrix}}. \nonumber 
\label{eqqInf6xx}
\end{align}
In addition, we define a {\it truncated semi-Markov kernel} as 
\[\bar{Q}_{ij}(\tau, \munderbar{s}, \bar{s}) = \int_{s = \munderbar{s}}^{\bar{s}} (\bar{g}_{ij}(\tau+s)-\bar{g}_{ij}(s))\, \cdot \, \frac{V_{i}(\tau+s|\lambda_i)-V_{i}(s|\lambda_i)}{1-V_{i}(s|\lambda_i)} \,\cdot\, dV_{i}(s|\lambda_i),\]
a scalar-valued function $\bar{Q}_{i}(\tau, \munderbar{s}, \bar{s}) = \sum_{j \in \mathcal{X}\setminus\{i\}} \bar{Q}_{ij}(\tau, \munderbar{s}, \bar{s})$, and a matrix-valued function 
\begin{align} 
{\bf \bar{Q}}(\tau, \munderbar{s}, \bar{s}) = \underbrace{\left[
\begin{array}{c|c|c|c}
0 & \bar{Q}_{21}(\tau, \munderbar{s}, \bar{s}) & \cdots & \bar{Q}_{N1}(\tau, \munderbar{s}, \bar{s})\\ 
\bar{Q}_{12}(\tau, \munderbar{s}, \bar{s}) & 0 & \cdots & \bar{Q}_{N2}(\tau, \munderbar{s}, \bar{s})\\
\vdots & \vdots &  & \vdots \\
\bar{Q}_{1N}(\tau, \munderbar{s}, \bar{s}) & \bar{Q}_{2N}(\tau, \munderbar{s}, \bar{s}) & \cdots & 0
\end{array}\right]}_{\small \mbox{Size}\, N \times N \, \small \mbox{matrix}}. \nonumber 
\label{eqqInf6xx222}
\end{align}

\begin{theorem}[Interval transition probabilities]
\label{thm1} 
Let ${\bf \tilde{P}}(\tau, \munderbar{s}, \bar{s})$ be the solution to the following integral equation 
\begin{equation}
{\bf \tilde{P}}(\tau, \munderbar{s}, \bar{s}) = {\bf I}_{N \times N} - \mbox{\textnormal{diag}}\left(\bar{Q}_{1}(\tau, \munderbar{s}, \bar{s}), \dots, \bar{Q}_{N}(\tau, \munderbar{s}, \bar{s})\right) + \int_{u=0}^{\tau} \frac{\partial{\bf \bar{Q}}(u, \munderbar{s}, \bar{s})}{\partial u} \,\times\, {\bf \tilde{P}}(\tau-u, 0, 0)\, du,  
\label{eqqInf6}
\end{equation}
for the three independent variables $(\tau, \munderbar{s}, \bar{s}) \in \mathcal{S}$. Then, the interval transition probability $p_{ij}$ is given by
\[p_{ij}(t_m-t_{m-w}, \psi(m-w, w^{'})) = \tilde{p}_{ij}(\tau, \munderbar{s}, \bar{s}), \forall i, j \in \mathcal{X},\] 
at $\tau = t_m-t_{m-w},$ $\munderbar{s} = t_m-t_{m-w+1}$, and $\bar{s} = t_m-t_{m-w+w^{'}}$.
\begin{proof}
See Appendix \ref{appA}. 
\end{proof}
\end{theorem}
Theorem \ref{thm1} follows from a {\it first-step analysis} that is akin to the derivation of the conventional Chapman-Kolmogorov equations in ordinary Markov chains (\cite{kulkarni1996modeling}). The integral equation in (\ref{eqqInf6}) is a (matrix-valued) non-homogeneous {\it Volterra integral equation of the second kind} (\cite{polyanin2008handbook}).\\
\\
It can be easily demonstrated that a closed-form solution that hinges on conventional kernel methods cannot be obtained. Hence, we resort to a numerical method in order to solve (\ref{eqqInf6}) for ${\bf \tilde{P}}(\tau, \munderbar{s}, \bar{s}), \, \forall (\tau, \munderbar{s}, \bar{s}) \in \mathcal{S}$. Before presenting the numerical method, we reformulate (\ref{eqqInf6}) as follows    
\begin{equation}
{\bf \tilde{P}}(\tau, \munderbar{s}, \bar{s}) = {\bf I}_{N \times N} - \mbox{\textnormal{diag}}\left(\bar{Q}_{1}(\tau, \munderbar{s}, \bar{s}), \dots, \bar{Q}_{N}(\tau, \munderbar{s}, \bar{s})\right) + \left(\frac{\partial{\bf \bar{Q}}(., \munderbar{s}, \bar{s})}{\partial u} \,\star\, {\bf \tilde{P}}(., 0, 0)\right)(\tau),  
\label{eqqInf77}
\end{equation}
where $\star$ is an element-wise convolution operator. (\ref{eqqInf77}) follows from (\ref{eqqInf6}) by the fact that the integral in (\ref{eqqInf6}) is a convolution integral; (\ref{eqqInf77}) can be expressed as follows   
\begin{equation}
{\bf \tilde{P}}(\tau, \munderbar{s}, \bar{s}) = \mathcal{B}\{{\bf \bar{Q}}(\tau, \munderbar{s}, \bar{s})\}({\bf \tilde{P}}(\tau, \munderbar{s}, \bar{s})),  
\label{eqqInf77x1}
\end{equation}
where the (functional) operator $\mathcal{B}\{{\bf Q}\}({\bf \tilde{P}})$ is given by
\begin{align}
&\mathcal{B}\{{\bf \bar{Q}}(\tau, \munderbar{s}, \bar{s})\}({\bf \tilde{P}}(\tau, \munderbar{s}, \bar{s})) = \nonumber \\ 
&{\bf I}_{N \times N} - \mbox{\textnormal{diag}}\left(\bar{Q}_{1}(\tau, \munderbar{s}, \bar{s}), \dots, \bar{Q}_{N}(\tau, \munderbar{s}, \bar{s})\right) + \mathscr{F}^{-1}\left\{\mathscr{F}\left\{\frac{\partial{\bf \bar{Q}}(\tau, \munderbar{s}, \bar{s})}{\partial \tau}\right\} \,\cdot\, \mathscr{F}\left\{{\bf \tilde{P}}(\tau, 0, 0)\right\}\right\},  
\label{eqqInf77x2}
\end{align}
where $\mathscr{F}$ is the Fourier transform operator, and the transforms in (\ref{eqqInf77x2}) are all taken with respect to $\tau$.\\

The solution to (\ref{eqqInf77x1}) can be obtained via the {\it successive approximation} method (\cite{opial1967weak}) as follows. We initialize the function ${\bf \tilde{P}}(\tau, \munderbar{s}, \bar{s})$ with the truncated semi-Markov kernel\footnote{This is a reasonable initialization since the entries of the semi-Markov kernel correspond to interval transition probabilities conditioned on there being no intermediate transitions on the way from state $i$ to state $j$.} ${\bf \bar{Q}}(\tau, \munderbar{s}, \bar{s})$, and then iteratively apply the operator $\mathcal{B}(.)$ to obtain a new value for ${\bf \tilde{P}}(\tau, \munderbar{s}, \bar{s})$ until convergence. That is, the successive approximation procedure goes as follows 
\begin{align}
&{\bf \tilde{P}}^{o}(\tau, \munderbar{s}, \bar{s}) = {\bf \bar{Q}}(\tau, \munderbar{s}, \bar{s}) \nonumber \\
&\mbox{While}\,\, \norm{\,{\bf \tilde{P}}^{z}(\tau, \munderbar{s}, \bar{s})-{\bf \tilde{P}}^{z-1}(\tau, \munderbar{s}, \bar{s})\,}_{\infty} > \epsilon \nonumber \\
&{\bf \tilde{P}}^{z}(\tau, \munderbar{s}, \bar{s}) = \mathcal{B}\{{\bf \bar{Q}}(\tau, \munderbar{s}, \bar{s})\}({\bf \tilde{P}}^{z-1}(\tau, \munderbar{s}, \bar{s})). \nonumber \\
\label{eqqInf77x3}
\end{align}
The following Theorem establishes the validity of the procedure in (\ref{eqqInf77x3}) as a solver for (\ref{eqqInf77x1}). Before presenting the statement of Theorem \ref{thm2}, we define the function space $\mathcal{P}$ as follows
\[\mathcal{P} = \left\{{\bf \tilde{P}}(\tau, \munderbar{s}, \bar{s}):\,\, \tilde{p}_{ij}(\tau, \munderbar{s}, \bar{s}) \in [0,1], \sum_j \tilde{p}_{ij}(\tau, \munderbar{s}, \bar{s}) = 1, \tilde{p}_{ij}(0, \munderbar{s}, \bar{s}) = \delta_{ij}, (\tau, \munderbar{s}, \bar{s}) \in \mathcal{S}\right\},\]
where $\delta_{ij}$ is the Kronecker delta function.  
\begin{theorem}[Convergence of successive approximations]
\label{thm2} 
The functional $\mathcal{B}\{{\bf \bar{Q}}\}({\bf \tilde{P}})$ has a unique fixed-point ${\bf \tilde{P}}^{*}$ in $\mathcal{P},$ and the successive approximation procedure in (\ref{eqqInf77x3}) always converges to the fixed point, i.e. ${\bf \tilde{P}}^{\infty}(\tau, \munderbar{s}, \bar{s}) = {\bf \tilde{P}}^{*}(\tau, \munderbar{s}, \bar{s})$, starting from any initial value ${\bf \tilde{P}}^{o}(\tau, \munderbar{s}, \bar{s}) \in \mathcal{P}$.  
\begin{proof}
See Appendix \ref{appB}. 
\end{proof}
\end{theorem} 
It is important to note that we do not need to solve for ${\bf \tilde{P}}(\tau, \munderbar{s}, \bar{s})$ during real-time inference. Instead, we create a look-up table comprising a discretized version of ${\bf \tilde{P}}(\tau, \munderbar{s}, \bar{s}) = \left[\tilde{p}_{ij}(a \Delta \tau, b \Delta \munderbar{s}, c \Delta \bar{s}))\right]_{i,j, a, b,c}$, and then we query this table when performing real-time inference for monitored patients. Hence, efficient and fast inferences can be provided for critical care patients for whom prompt diagnoses are necessary for the efficacy of clinical interventions. Algorithm \ref{alg2} shows a pseudocode for constructing a look-up table of interval transition probabilities, \texttt{TransitionLookUp}($\Gamma$, $\epsilon$), which takes as an input the parameter set $\Gamma$ and a precision level $\epsilon$ (to control the termination of the successive approximation iterations), and outputs the interval transitions look-up table ${\bf \tilde{P}}(\tau, \munderbar{s}, \bar{s})$. In Algorithm \ref{alg2}, FFT and IFFT refer to the fast Fourier transform operation and its inverse, respectively, and ``diff(.)" refers to a numerical differentiation operation. \\
\\
\begin{algorithm}[t]
  \caption{Constructing a look-up table of interval transition probabilities}\label{alg2}
  \begin{algorithmic}[1]  
    \Procedure{\texttt{TransitionLookUp}}{$\Gamma$, $\epsilon$}
		  \State {\bfseries Input:} HASMM parameters $\Gamma$ and precision $\epsilon$ 
	    \State {\bfseries Output:} A look-up table $\left[\, \tilde{p}_{ij}(a \Delta \tau, b \Delta \munderbar{s}, c \Delta \bar{s})\, \right]_{i,j, a, b,c}$ 
      \State Set the values of $A, B$ and $C$ (number of steps), $\Delta \tau$ (step sizes)
			\For{$a$ = 1 to $A$, $b$ = 1 to $B$, $c$ = 1 to $C$}
        \State $g^{\tau}_{ij}(a \Delta \tau) \gets \sum_{x=1}^{a}\frac{e^{(\eta_{ij}+\beta_{ij} x \Delta \tau)}}{\sum_{k=1}^{N} e^{(\eta_{ik}+\beta_{ik} x \Delta \tau)}} \, \left(\frac{1}{\Gamma(\lambda_{i,s})\,\lambda_{i,r}^{\lambda_{i,s}}}\, (x \Delta \tau)^{\lambda_{i,s}-1}\,e^{-\frac{x \Delta \tau}{\lambda_{i,r}}}\right)\, \Delta \tau$ 
				\State $g^{s}_{ij}(a \Delta s) \gets \sum_{x=1}^{a}\frac{e^{(\eta_{ij}+\beta_{ij} x \Delta s)}}{\sum_{k=1}^{N} e^{(\eta_{ik}+\beta_{ik} x \Delta s)}} \, \left(\frac{1}{\Gamma(\lambda_{i,s})\,\lambda_{i,r}^{\lambda_{i,s}}}\, (x \Delta s)^{\lambda_{i,s}-1}\,e^{-\frac{x \Delta s}{\lambda_{i,r}}}\right)\, \Delta s$ 
				\State $\bar{Q}_{ij}(a \Delta \tau, b \Delta \munderbar{s}, c \Delta \bar{s}) \gets \sum_{x=b}^{c} \frac{(g^{\tau}_{ij}(a \Delta \tau)-g^{s}_{ij}(x \Delta s))\,(V_i(a \Delta \tau |\lambda_i)-V_i(x \Delta s|\lambda_i))}{1-V_i(x \Delta s|\lambda_i)}\, v_i(x \Delta s|\lambda_i)$
      \EndFor	
			  \State $e = \epsilon + 1$
				\State $z \gets 1$
				\State $\tilde{p}^{(o)}_{ij}(a \Delta \tau, b \Delta \munderbar{s}, c \Delta \bar{s}) \gets \bar{Q}_{ij}(a \Delta \tau, b \Delta \munderbar{s}, c \Delta \bar{s}),\, \forall a, b, c, i, j.$
			\While{$e > \epsilon$} 
			  \State $CQ_{i,j,k}(a \Delta \tau, b \Delta \munderbar{s}, c \Delta \bar{s}) \gets$ \\
				$\,\,\,\,\,\,\,\,\,\,\,\,\,\,\,\,\,\,\,\,\,\,\,\,\,\,\,\,\,\,\,\,\,\,\, \mbox{IFFT}\left(\mbox{FFT}\left(\mbox{diff}\left(\bar{Q}_{ik}(a \Delta \tau, b \Delta \munderbar{s}, c \Delta \bar{s})\right)\right), \mbox{FFT}\left(\tilde{p}^{(z-1)}_{jk}(a \Delta \tau, b \Delta \munderbar{s}, c \Delta \bar{s})\right)\right),$ 
				\State $\tilde{p}^{(z)}_{ij}(a \Delta \tau, b \Delta \munderbar{s}, c \Delta \bar{s}) \gets \delta_{ij}\, \bar{Q}_{ij}(a \Delta \tau, b \Delta \munderbar{s}, c \Delta \bar{s}) + \sum_{k=1}^{N} CQ_{i,j,k}(a \Delta \tau, b \Delta \munderbar{s}, c \Delta \bar{s})$
				\State ${\bf \tilde{P}}^{(z)}(a \Delta \tau, b \Delta \munderbar{s}, c \Delta \bar{s}) = \left[\tilde{p}^{(z)}_{ij}(a \Delta \tau, b \Delta \munderbar{s}, c \Delta \bar{s}))\right]_{i,j, a, b,c}$
				\State $e \gets \norm{\,{\bf \tilde{P}}^{(z)}(a \Delta \tau, b \Delta \munderbar{s}, c \Delta \bar{s})-{\bf \tilde{P}}^{(z-1)}(a \Delta \tau, b \Delta \munderbar{s}, c \Delta \bar{s})\,}_{\infty}$
				\State $z \gets z + 1$
      \EndWhile\label{euclidendwhile}
      \State \textbf{return} ${\bf \tilde{P}}^{(z)}(a \Delta \tau, b \Delta \munderbar{s}, c \Delta \bar{s})$
    \EndProcedure
  \end{algorithmic}
\end{algorithm}
Now that we have constructed the algorithm \texttt{TransitionLookUp} to compute the interval transition probabilities in the look-up table ${\bf \tilde{P}}(a \Delta \tau, b \Delta \munderbar{s}, c \Delta \bar{s})$, we can implement a forward-filtering inference algorithm using dynamic programming (by virtue of the recursive formula in (\ref{eqqInf4})). In particular, the posterior probability of the patient's current clinical state in terms of the forward messages can be written as
\begin{align}
\mathbb{P}(X(t_m) = j \left|\, y(t_1),.\,.\,., y(t_m)\right.) &= \frac{\sum_{w=1}^{m} \alpha_{m}(j, w)}{\sum_{k=1}^{N}\sum_{w=1}^{m} \alpha_{m}(k, w)}.
\label{eqfilter}
\end{align}
Algorithm \ref{alg3}, \texttt{ForwardFilter}, implements real-time inference of a patient's clinical state given a sequence of measurements $\{y(t_1),.\,.\,., y(t_m)\}$. In Algorithm \ref{alg3}, we invoke \texttt{TransitionLookUp} initially to construct the look-up table of transition probabilities, but in practice, the look-up table can be constructed in an offline stage once the HASMM parameter set $\Gamma$ is known. The number of computations can be reduced by limiting the lags $w$ for every forward message $\alpha_{m}(j, w)$ to the samples in $\mathcal{T}$ that reside in a period $t_m-T_{max}$, where $T_{max}$ is derived from the Gamma distribution of the sojourn time (e.g. $T_{max}$ can be selected such that $V_{i}(T_{max}|\lambda_{i}) = 90\%$). The complexity of \texttt{ForwardFilter} is similar to the conventional forward algorithms in (\cite{rabiner1989tutorial}), i.e. $\mathcal{O}(Nm^{2})$.

\begin{algorithm}[t]
  \caption{Forward filtering inference}\label{alg3}
  \begin{algorithmic}[1]
    \Procedure{\texttt{ForwardFilter}}{$\Gamma$, $\{y(t_w)\}_{w=1}^{m}$, $\epsilon$}
		  \State {\bfseries Input:} Observed samples $\{y(t_w)\}_{w=1}^{m}$, HASMM parameters $\Gamma$, and precision $\epsilon$
	    \State {\bfseries Output:} The posterior state distribution $\left\{\mathbb{P}(X(t_m) = j\,|\,\{y(t_w)\}_{w=1}^{m})\right\}_{j=1}^{N}$ 
      \State ${\bf \tilde{P}}(a \Delta \tau, b \Delta \munderbar{s}, c \Delta \bar{s})  \gets$ \texttt{TransitionLookUp}($\Gamma$, $\epsilon$)
			\State $\alpha_{1}(j, 1) = d\mathbb{P}(y(t_1)\, | \, X(t_1) = j) \, \sum_{i=1}^{N}\tilde{p}_{ij}(t_1, 0, 0) \, \cdot \, p^{o}_{i}, \,\, \forall j \in \mathcal{X}$
			\For{$z$ = 2 to $m$}
			\For{$w$ = 1 to $z$}
			  \State $a^{*}(z,w) = \mbox{arg}\,\mbox{min}_{a} \left|t_z-t_{z-w}-a\Delta \tau\right|$
				\State $b^{*}(z,w) = \mbox{arg}\,\mbox{min}_{b} \left|t_{z}-t_{z-w+1}-b\Delta \munderbar{s}\right|$
				\State $c^{*}(z,w,w^{'}) = \mbox{arg}\,\mbox{min}_{c} \left|t_{z}-t_{z-w-w^{'}}- c\Delta \bar{s}\right|$
        \State $\alpha_{z}(j, w) = d\mathbb{P}(\{y(t_u)\}_{u=z-w+1}^{z}\, | \, X(t_z) = j)\, \sum_{i=1}^{N}\sum_{w^{'}=1}^{z-w} \alpha_{z-w}(i, w^{'})\, \times$ 
				\begin{align} 
				\tilde{p}_{ij}(a^{*}(z,w)\Delta \tau, b^{*}(z,w)\Delta \munderbar{s}, c^{*}(z,w,w^{'}) \Delta \bar{s})\, \times \left(V_j(t_z-t_{z-w}|\lambda_j)-V_j(t_z-t_{z-w+1}|\lambda_j)\right)  \nonumber
\end{align}
			\EndFor
      \EndFor	
			\State $\mathbb{P}(X(t_m) = j \left|\, \{y(t_u)\}_{u=1}^{m}\right.) = \frac{\sum_{w=1}^{m} \alpha_{m}(j, w)}{\sum_{k=1}^{N}\sum_{w=1}^{m} \alpha_{m}(k, w)}$
      \State \textbf{return} $\{\mathbb{P}(X(t_m) = j\,|\,\{y(t_w)\}_{w=1}^{m})\}_{j=1}^{N}$
    \EndProcedure
  \end{algorithmic}
\end{algorithm}  

\subsection{Prognostic risk scoring using an HASMM}
\label{sec3.33}
Diagnostic inference, e.g. estimating the patient's current state after a screening test, can be conducted by a direct application of the forward filtering algorithm presented in the previous Subsection. In this Subsection, we now explain how prognostic inferences can also be conducted using the algorithms presented in the previous Subsection. Prognostic risk scoring plays an important role in designing screening guidelines (\cite{gail2010comparing}), acute care interventions (\cite{knaus1985apache}) and surgical decisions (\cite{foucher2007semi}). A risk score is a measure for the patient's risk of encountering an adverse event (abstracted as state $N$ in our model) at any future time step starting from time $t_m$. That is, the patient's risk score at time $t_m$ can be formulated as 
\begin{align}
R(t_m) &= \mathbb{P}(\mathcal{A}_{N} \, | \, \{y(t_u)\}_{u=1}^{m}, \Gamma)  \nonumber \\
&= 1-\mathbb{P}(X(\infty) = N \, | \, \{y(t_u)\}_{u=1}^{m}, \Gamma),
\label{riskeq}
\end{align}
which can be computed using the outputs of \texttt{TransitionLookUp} and \texttt{ForwardFilter} as follows 
\begin{align}
R(t_m) &= \sum_{j=1}^{N} \tilde{p}_{jN}(A, 0, 0) \, \cdot \, \frac{\sum_{w=1}^{m} \alpha_{m}(j, w)}{\sum_{k=1}^{N}\sum_{w=1}^{m} \alpha_{m}(k, w)}. 
\label{riskeq4}
\end{align}
Therefore, the procedures \texttt{TransitionLookUp} and \texttt{ForwardFilter} suffice for executing both the diagnostic and prognostic inference tasks. 

\section{Learning Hidden Absorbing Semi-Markov Models}
\label{sec3.3}

In Section \ref{sec3.2}, we developed an inference algorithm that can handle diagnostic and prognostic tasks for patients in real-time assuming that the true HASMM parameter set $\Gamma$ is known. In practice, the parameter set $\Gamma$ is not known, and has to be learned from an offline EHR dataset $\mathcal{D}$ that comprises $D$ episodes for previously hospitalized or monitored patients, i.e.
\[\mathcal{D} = \left\{\{y^{d}_{m}, t^{d}_{m}\}_{m=1}^{M^{d}}, T^{d}_c, l^{d}\right\}_{d=1}^{D},\]  
where $\{y^{d}_{m}, t^{d}_{m}\}_{m=1}^{M^{d}}$ are the observable variables and their respective sampling times for the $d^{th}$ episode, $T^{d}_c$ is the episode's censoring time, and $l^d \in \{1, N\}$ is a label for the realized absorbing state. \\

We note that unlike the conventional HMM learning setting (\cite{rabiner1989tutorial, zhang2001segmentation, nodelman2012expectation}), the episodes are not of equal-length as the observations for every episode stop at a random, but informative, censoring time. Thus, the patient's state trajectory does not manifest only in the observable time series, i.e. $\{y^{d}_{m}, t^{d}_{m}\}_{m=1}^{M^{d}}$, but also in the episode's censoring variables $\{T^{d}_c, l^{d}\}$. In this Section, we develop an efficient algorithm, which we call the {\it forward-filtering backward-sampling Monte Carlo EM} (FFBS-MCEM) algorithm, that computes the Maximum Likelihood (ML) estimate of $\Gamma$ given an informatively censored dataset $\mathcal{D}$, i.e. $\Gamma^{*} = \arg\,\max_{\Gamma}\, \Lambda(\mathcal{D}\,|\,\Gamma),$ where $\Lambda(\mathcal{D}\,|\, \Gamma) = d\mathbb{P}(\mathcal{D}\,|\, \Gamma)$ is the likelihood of the dataset $\mathcal{D}$ given the parameter set $\Gamma$. We start by presenting the learning setup in Section \ref{sec3.31}, and then we present the FFBS-MCEM algorithm Section \ref{sec3.32}.  

\subsection{The Learning Setup}
\label{sec3.31}
We focus on the challenging scenario when no domain knowledge or diagnostic assessments for the patients' latent states are provided in the dataset \footnote{For some settings, such as chronic kidney disease progression estimation (\cite{eddy2006chronic}), the EHR records may include some anchors or assessments to the latent states over time. A simpler version of the learning algorithm proposed in this Section can be used to deal with such datasets. In critical care settings, it is more common that the EHR records are not labeled with any clinical state assessments over time (\cite{ForecastICU}).} $\mathcal{D}$ (with the exception of the absorbing state which is declared by the variable $l^{d}$), i.e. the learning setup is an {\it unsupervised} one. For such a scenario, the main challenge in constructing the ML estimator $\Gamma^{*}$ resides in the hiddenness of the patients' state trajectories in the training dataset $\mathcal{D}$; the dataset $\mathcal{D}$ contains only the sequence of observable variables, their respective observation times, the episode’s censoring time and the state in which the trajectory was absorbed. If the patients' latent state trajectories $(X(t))_{t\in\mathbb{R}_{+}}$ were observed in $\mathcal{D}$, the ML estimation problem $\Gamma^{*} = \mbox{arg}\,\mbox{max}_{\Gamma}\, \mathbb{P}(\mathcal{D}\,|\,\Gamma)$ would have been straightforward; the hiddenness of $(X(t))_{t\in\mathbb{R}_{+}}$ entails the need for marginalizing over the space of all possible latent trajectories conditioned on the observed variables, which is a hard task even for conventional continuous-time HMM models (\cite{liu2015efficient, nodelman2012expectation, leiva2011visualization, metzner2007generator}). \\   

In order to construct the ML estimator for $\Gamma,$ we start by writing the complete likelihood, i.e. the likelihood of an HASMM with a parameter set $\Gamma$ to generate both the hidden states trajectory $\{x^{d}_n, s^{d}_n\}_{n=1}^{k^{d}}$ and the observable variables $\{y^{d}_m, t^{d}_m\}_{m=1}^{M^{d}}$ for the $d^{th}$ episode in the dataset $\mathcal{D}$ as follows
\begin{align}
&d\mathbb{P}\left(\left.\{x^{d}_n, s^{d}_n\}_{n=1}^{k^{d}}, \{y^{d}_m, t^{d}_m\}_{m=1}^{M^{d}}\,\right|\,\Gamma \right) = \mathbb{P}(x^{d}_1 |\Gamma) \, \cdot \, d\mathbb{P}(s^{d}_{1} | x^{d}_{1}, \Gamma) \, \cdot\,  d\mathbb{P}(\{y^{d}_m, t^{d}_m\}_{t^{d}_m \in \mathcal{T}^{d}_{1}} | x^{d}_{1}, \Gamma)\, \times \nonumber \\
&\prod_{n=2}^{k^{d}}\mathbb{P}(x^{d}_{n} \,|\, x^{d}_{n-1}, s^{d}_{n-1}, \Gamma) \, \cdot \, d\mathbb{P}(s^{d}_{n} \,|\, x^{d}_{n}, \Gamma) \, \cdot \, d\mathbb{P}(\{y^{d}_m, t^{d}_m\}_{t^{d}_m \in \mathcal{T}^{d}_{n}} \,|\, x^{d}_{n}, \Gamma),    
\label{eqLkl1}
\end{align}
where $k^d$ is the number of states that realized in episode $d$ from $t=0$ until absorption. The factorization in (\ref{eqLkl1}) follows from the conditional independence properties of the HASMM variables (see Figure \ref{FigQ4xs}). Since we cannot observe the latent states trajectory $\{x^{d}_n, s^{d}_n\}_{n=1}^{k^{d}}$, the ML estimator deals with the expected likelihood $\Lambda(\mathcal{D}\,|\, \Gamma),$ which is evaluated by marginalizing the complete likelihood over the latent states trajectories, i.e.
\begin{align}
\Lambda(\mathcal{D}\,|\, \Gamma) &= \mathbb{E}_{x(t)|\mathcal{D},\, \Gamma}\left[\,\,\prod_{d=1}^{D}\, d\mathbb{P}(\{x^{d}_n, s^{d}_n\}_{n=1}^{k^{d}}, \{y^{d}_m, t^{d}_m\}_{m=1}^{M^{d}}\,|\,\Gamma)\right] \nonumber \\
&= \prod_{d=1}^{D}\,\mathbb{E}_{x^d(t)|\mathcal{D},\, \Gamma}\left[\,\,d\mathbb{P}(\{x^{d}_n, s^{d}_n\}_{n=1}^{k^{d}}, \{y^{d}_m, t^{d}_m\}_{m=1}^{M^{d}}\,|\,\Gamma)\,\right] \nonumber \\
&= \prod_{d=1}^{D}\, \int d\mathbb{P}(\{x^{d}_n, s^{d}_n\}_{n=1}^{k^{d}}, \{y^{d}_m, t^{d}_m\}_{m=1}^{M^{d}}\,|\,\Gamma) \, \cdot \, d\mathbb{P}(\{x^{d}_n, s^{d}_n\}_{n=1}^{k^{d}}\,|\,\mathcal{D},\, \Gamma),
\label{eqLklx2} 
\end{align}
where we have assumed that the episodes in $\mathcal{D}$ are independent and identically distributed. The integral in (\ref{eqLklx2}) can be further decomposed as follows
\begin{align}
\Lambda(\mathcal{D}\,|\, \Gamma) &= \prod_{d=1}^{D} \int \mathbb{P}(x^{d}_1 |\Gamma) \, \cdot \, d\mathbb{P}(s^{d}_{1} | x^{d}_{1}, \Gamma) \, \cdot \, d\mathbb{P}(\{y^{d}_m, t^{d}_m\}_{t^{d}_m \in \mathcal{T}^{d}_{1}} | x^{d}_{1}, \Gamma) \,\cdot\, d\mathbb{P}(\{x^{d}_n, s^{d}_n\}_{n=1}^{k^{d}}\,|\,\mathcal{D},\, \Gamma) \nonumber \\ 
&\, \times \, \prod_{n=2}^{k^{d}}\mathbb{P}(x^{d}_{n} \,|\, x^{d}_{n-1}, s^{d}_{n-1}, \Gamma) \, \cdot \, d\mathbb{P}(s^{d}_{n} \,|\, x^{d}_{n}, \Gamma) \, \cdot \, d\mathbb{P}(\{y^{d}_m, t^{d}_m\}_{t^{d}_m \in \mathcal{T}^{d}_{n}} \,|\, x^{d}_{n}, \Gamma).    
\label{eqLkl2xD}
\end{align}

\subsection{Challenges Facing the HASMM Learning Task}
\label{sec3.301}
The problem of learning the HASMM parameters by maximizing the likelihood function in (\ref{eqLkl2xD}) is obstructed by various obstacles that hinder the deployment of off-the-shelf learning algorithms; we list these challenges hereunder.
\begin{enumerate}
\item Finding the ML estimate $\Gamma^{*}$ by direct maximization of $\Lambda(\mathcal{D}\,|\, \Gamma)$ is not viable due to the intractability of the integral in (\ref{eqLkl2xD}), i.e. $\Lambda(\mathcal{D}\,|\, \Gamma)$ has no analytic maximizer. The difficulty of evaluating the expected likelihood $\Lambda(\mathcal{D}\,|\, \Gamma)$ follows from the need to average the complete likelihood over a complicated posterior density function for the latent state trajectory. 
\item Direct adoption of the conventional Baum-Welch implementation of the EM algorithm as a solution to the intractable problem of maximizing the expected likelihood in (\ref{eqLkl2xD}) --as has been applied in HMMs (\cite{rabiner1989tutorial}), HSMMs (\cite{murphy2002hidden}), EDHMMs and VDHMMs (\cite{yu2010hidden}-- is not possible for the HASMM setting. This is due to the intractability of the integral involved in the E-step; a problem that is also faced by other continuous-time models (\cite{liu2015efficient, nodelman2012expectation}). However, these models assumed Markovian state trajectories, in which case the implementation of the E-step boils down to computing the expected state durations and transition counts as sufficient statistics for estimating the latent trajectories\footnote{Different approaches have been developed in the literature for computing these quantities: (\cite{wang2014unsupervised}) assumes that the transition rate matrix is diagonalizable, and hence utilize a closed-form estimator for the transition rates, whereas (\cite{liu2015efficient}) uses the {\it Expm} and {\it Unif} methods (originally developed in (\cite{hobolth2011summary})) to evaluate the integrals of the transition matrix exponential. Unfortunately, none of these methods could be utilized for computing the proximal log-likelihood of an HASMM due to the semi-Markovianity of the state trajectory (i.e. state-durations are not exponentially distributed as it is the case in (\cite{liu2015efficient, nodelman2012expectation, hobolth2011summary, wang2014unsupervised})).} (e.g. see Equations (12) and (13) in (\cite{liu2015efficient})). This simplification, which follows from the plausible properties of the Markov chain's transition rate matrix, does not materialize for semi-Markovian transitions. Further complications are introduced by the duration-dependence of the state-transitions and the segmental nature of the observables.
\item Learning an informatively censored dataset would naturally benefit from the information conveyed in the censoring variables $\{T_c^d, l^d\}$. However, the availability of censoring information leads to more complicated posterior density expressions for the latent state trajectories, which complicates the job of any analytic, variational or Monte Carlo based inference method one would use to infer the latent state trajectories\footnote{Note that the censoring information are only (causally) available in the model training (learning) phase since we deal with an offline batch of data through which we can see the full patients' episodes, whereas real-time inference, discussed in the previous Section, does not take advantage of any external censoring information since this information cannot be causally provided to our algorithms.}.   
\end{enumerate}
In the following Section, we present a learning algorithm that addresses the above challenges, and provides insights into general settings in which informatively censored time series data are to be dealt with.

\subsection{The Forward-filtering Backward-sampling Monte Carlo EM Algorithm}
\label{sec3.32}
\subsubsection{Expectation-Maximization}
\begin{figure*}[t]
        \centering
        \includegraphics[width=4.5in]{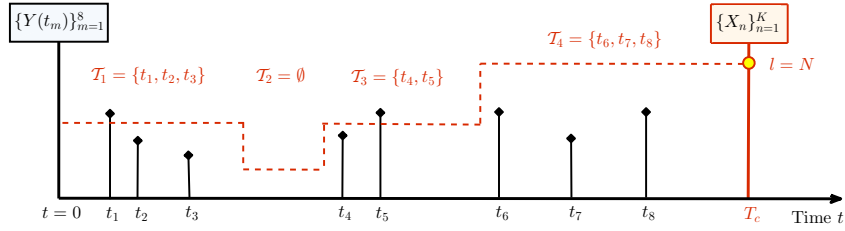}
				\captionsetup{font= small}
        \caption{An episode that comprised 8 observable samples, censored at time $T_c$, and absorbed in state $N$ (catastrophic state). The dashed state trajectory is a trajectory that could have generated the observables with a positive probability. Computing the proximal log-likelihood requires averaging over infinitely many paths that could have generated the observables with a positive probability.}
\label{Figpaths}
\end{figure*}

As in the case of classical discrete and continuous-time HMMs, we address the first challenge stated in Section \ref{sec3.301} by using the EM algorithm (\cite{liu2015efficient, nodelman2012expectation}). The iterative EM algorithm starts with an initial guess $\hat{\Gamma}^{o}$ for the parameter set, and maximizes a proxy for the log-likelihood in the $z^{th}$ iteration as follows:
\begin{itemize}
			\item {\bf E-step:}\,\, $Q(\Gamma; \hat{\Gamma}^{z-1}) = \sum_{d=1}^{D}\mathbb{E}\left[\left.\log(\mathbb{P}(\{x^{d}_n, s^{d}_n\}_{n=1}^{k^{d}}, \{y^{d}_m, t^{d}_m\}_{m=1}^{M^{d}}\,|\,\Gamma))\right|\, \mathcal{D}, \hat{\Gamma}^{z-1}\right].$ 
			\item {\bf M-step:}\,\, $\hat{\Gamma}^{z} = \arg\,\max_{\Gamma}\, Q(\Gamma; \hat{\Gamma}^{z-1}).$
\end{itemize}
The E-step computes the {\it proximal expected log-likelihood} $Q(\Gamma; \hat{\Gamma}^{z-1})$, which entails evaluating the following integral
\begin{align}
Q(\Gamma; \hat{\Gamma}^{z-1}) = \sum_{d=1}^{D}\int \log(d\mathbb{P}(\{x^{d}_n, s^{d}_n\}_{n=1}^{k^{d}}, \{y^{d}_m, t^{d}_m\}_{m=1}^{M^{d}}\,|\,\Gamma)) \, \cdot \, d\mathbb{P}(\{x^{d}_n, s^{d}_n\}_{n=1}^{k^{d}} \,|\, \mathcal{D}, \hat{\Gamma}^{z-1}), \nonumber 
\end{align}
where $d\mathbb{P}(\{x^{d}_n, s^{d}_n\}_{n=1}^{k^{d}} \,|\, \mathcal{D}, \hat{\Gamma}^{z-1}) = d\mathbb{P}(\{x^{d}_n, s^{d}_n\}_{n=1}^{k^{d}} \,|\, \{y^{d}_m, t^{d}_m\}_{m=1}^{M^{d}}, x^{d}(T^{d}_c) = l^{d}, \hat{\Gamma}^{z-1}).$ That is, the proximal expected log-likelihood $Q(\Gamma; \hat{\Gamma}^{z-1})$ is computed by marginalizing the likelihood of the observed samples of the $d^{th}$ episodes $\{y^{d}_m, t^{d}_m\}_{m=1}^{M^{d}}$ over all potential latent paths $(x^{d}(t))_{t\in\mathbb{R}_{+}}$ that are censored at time $T^{d}_c$ and absorbed in state $l^{d}$. Figure \ref{Figpaths} depicts a set of observables $(\{y^{d}_m, t^{d}_m\}_{m=1}^{M^{d}}, x^d(T^d_c) = l^d)$ for one episode, and a potential latent path $\{x^d_n, s^d_n\}_{n=1}^{k^d}$ that could have generated such observables. Computing $Q(\Gamma; \hat{\Gamma}^{z-1})$ requires averaging over the posterior density of the latent paths conditional on an observable episode.   

\subsubsection{\small ``The Only Good Monte Carlo is a Dead Monte Carlo"}
Since computing $Q(\Gamma; \hat{\Gamma}^{z-1})$ does not admit a closed-form solution, as mentioned earlier in the second challenge stated in Section \ref{sec3.301}, we resort to a Monte Carlo approach for approximating the integral involved in the E-step (\cite{caffo2005ascent}). That is, in the $z^{th}$ iteration of the EM algorithm, we draw $G$ random trajectories $(\{x^{d,g}_n, s^{d,g}_n\}_{n=1}^{k^{d,g}})_{g=1}^{G}$ for every episode $d$, and use those trajectories to construct a Monte Carlo approximation $\hat{Q}_{G}(\Gamma; \hat{\Gamma}^{z-1})$ for the proximal log-likelihood function $Q(\Gamma; \hat{\Gamma}^{z-1})$. Sample trajectories are drawn from the posterior density of the latent states' trajectory conditional on the the observable variables (including the censoring information). That is to say, the $g^{th}$ sample trajectory $\{x^{d,g}_n, s^{d,g}_n\}_{n=1}^{k^{d,g}}$ is drawn as follows
\begin{align}
\{x^{d,g}_n, s^{d,g}_n\}_{n=1}^{k^{d,g}} \sim d\mathbb{P}(\{x^{d}_n, s^{d}_n\}_{n=1}^{k^{d}} \,|\, \{y^{d}_m, t^{d}_m\}_{m=1}^{M^{d}}, x^{d}(T^{d}_c) = l^{d}, \hat{\Gamma}^{z-1}),
\label{LLN0}
\end{align}
for $g \in \{1,.\,.\,.,G\}$. Hence, the proximal log-likelihood $Q(\Gamma; \hat{\Gamma}^{z-1})$ can be approximated via a Monte Carlo estimate $\hat{Q}_{G}(\Gamma; \hat{\Gamma}^{z-1})$ as follows 
\begin{align}
\hat{Q}_{G}(\Gamma; \hat{\Gamma}^{z-1}) \triangleq \sum_{d=1}^{D} \frac{1}{G} \sum_{g=1}^{G} \log(d\mathbb{P}(\{x^{d, g}_n, s^{d, g}_n\}_{n=1}^{k^{d, g}}, \{y^{d}_m, t^{d}_m\}_{m=1}^{M^{d}}\,|\,\Gamma)).
\label{LLN1}
\end{align}
It follows from the {\it Glivenko-Cantelli} Theorem (\cite{durrett2010probability}) that 
\[||\, Q(\Gamma; \hat{\Gamma}^{z-1})-\hat{Q}_{G}(\Gamma; \hat{\Gamma}^{z-1}) \,||_{\infty} = \sup_{\Gamma}|\, Q(\Gamma; \hat{\Gamma}^{z-1})-\hat{Q}_{G}(\Gamma; \hat{\Gamma}^{z-1})\, | \rightarrow 0 \,\,\,\, \mbox{a.s.},\]
and hence the Monte Carlo implementation of the E-step becomes more accurate as the sample size $G$ increases. Sampling trajectories from the posterior distribution specified in (\ref{LLN0}) in order to obtain a Monte Carlo estimate for $Q(\Gamma; \hat{\Gamma}^{z-1})$ is not a straight forward task; the sampler needs to jointly sample the states and their sojourn times taking into account the duration-dependent transitions among states, and that the number of variables sampled (number of states) $k^{d, g}$ in each trajectory is itself random. \\

Since there is no straightforward method that can generate samples for the random state trajectory $\{x^d_n, s^d_n\}_{n=1}^{k^d}$ from the joint posterior density in (\ref{LLN0}), the normative solution for such a problem is to resort to a Markov Chain Monte Carlo (MCMC) method such as Metropolis-Hastings or Gibbs sampling (\cite{carter1994gibbs}). Since the number of state and sojourn time variables, $k^d$, is itself random, one can even resort to a reversible jump MCMC method (\cite{green2009reversible}) in order to generate the samples for $\{x^d_n, s^d_n\}_{n=1}^{k^d}$. At this point of our analysis, we invoke the classical aphorism with which we titled this Subsection: {\it ``The Only Good Monte Carlo is a Dead Monte Carlo"} (\cite{trotter1956conditional}). By this quote, Trotter meant to advocate the view that sophisticated Monte Carlo methods should be avoided whenever possible; whenever an integral is analytically tractable, or whenever some analytic insights can be exploited to built simpler samplers, doing so should be preferred to an expensive Monte Carlo method. MCMCs are indeed expensive: they mix very slowly and they generate correlated samples. Adopting an MCMC to generate random state trajectories in every iteration of the EM algorithm and for every episode in $\mathcal{D}$ is beyond affordable. Fortunately, in the rest of this Section we show that an efficient sampler that generates independent samples of $\{x^d_n, s^d_n\}_{n=1}^{k^d}$ and for which the run-time is geometrically distributed can be constructed by capitalizing on the censoring information and utilizing some insights from the literature on sequential Monte Carlo smoothing (\cite{godsill2012monte}). \\

\subsubsection{The Forward-filtering Backward-sampling Recipe}
The availability of the censoring information (censoring time $T_c$ and absorbing state $x^{d}(T_c) = l^{d}$) for every episode $d$ in $\mathcal{D}$, together with the inherent non-linearity of the semi-Markovian transition dynamics encourage the development of a {\it forward-filtering backward-sampling} (FFBS) Monte Carlo algorithm\footnote{The methods used in this Section are also known in the literature as {\it sequential Monte Carlo} or {\it particle filtering} methods (\cite{godsill2012monte}).} that goes in the reverse-time direction of every episode by starting from the censoring instance, and sequentially sampling the latent states conditioned on the (sampled) future trajectory (\cite{godsill2012monte}). That is, unlike the {\it generative process} (described by the routine \texttt{GenerateHASMM}($\Gamma$)) which uses the knowledge of $\Gamma$ to generate sample trajectories by drawing an initial state and then sequentially going forward in time and sampling future states until absorption, the {\it inferential process} naturally goes the other way around: it exploits informative censoring by starting from the (known) final absorbing state (and censoring time), and sequentially samples a trajectory by traversing backwards in time and conditioning on the future. \\

We start constructing our forward-filter backward-sampler by first formulating the posterior density of the latent trajectory $\{x^{d}_n, s^{d}_n\}_{n=1}^{k^{d}}$ (from which we sample the $G$ trajectories in the $z^{th}$ iteration of the FFBS-MCEM algorithm as shown in (\ref{LLN0})) as follows  
\begin{align}
&d\mathbb{P}(\{x^{d}_n, s^{d}_n\}_{n=1}^{k^{d}} \,|\, \{y^{d}_m, t^{d}_m\}_{m=1}^{M^{d}},\, x^{d}(T^{d}_c) = l^{d},\, \hat{\Gamma}^{z-1}) = \nonumber \\ 
& d\mathbb{P}(s^{d}_{k^{d}} \,|\, \{y^{d}_m, t^{d}_m\}_{m=1}^{M^{d}},\, x^{d}(T^{d}_c) = l^{d}) \, \cdot\, \prod_{n=1}^{k^{d}-1} d\mathbb{P}(x^{d}_{n}, s^{d}_{n} \,|\, \underbrace{\{x^{d}_{n^{'}}, s^{d}_{n^{'}}\}_{n^{'}=n+1}^{k^{d}}}_{\small \mbox{Future trajectory}},\, \{y^{d}_m, t^{d}_m\}_{m=1}^{M^{d}},\, T^d_c), 
\label{eqqq1}
\end{align}
where the conditioning on the $(z-1)^{th}$ guess of the parameter set, $\hat{\Gamma}^{z-1}$, is suppressed for notational convenience. The formulation in (\ref{eqqq1}) decomposes the posterior density of the latent trajectory $\{x^{d}_n, s^{d}_n\}_{n=1}^{k^{d}}$ into factors in which the likelihood of every state $n$ is conditioned on the future trajectory starting from $n$ (i.e. the states $x^d_{n+1}$ up to the absorbing states, together with their corresponding sojourn times). The posterior density in (\ref{eqqq1}) can be further decomposed as follows
\begin{align}
&d\mathbb{P}(\{x^{d}_n, s^{d}_n\}_{n=1}^{k^{d}} \,|\, \{y^{d}_m, t^{d}_m\}_{m=1}^{M^{d}},\, x^{d}(T^{d}_c) = l^{d}) = d\mathbb{P}(s^{d}_{k^{d}} \,|\, x^{d}_{k^{d}} = l^{d}, s^{d}_{k^{d}} < T^d_c) \, \times \,\nonumber \\
&\prod_{n=1}^{k^{d}-1} d\mathbb{P}(x^{d}_{n}, s^{d}_{n} \,|\, \{x^{d}_{n^{'}}, s^{d}_{n^{'}}\}_{n^{'}=n+1}^{k^{d}}, s^{d}_{n} < \underbrace{T^d_c - (s^{d}_{n+1}+\,.\,.\,.\,+s^{d}_{k^d})}_{\small \mbox{Elapsed time in the episode}}, \{y^{d}_m, t^{d}_m\}_{m=1}^{M^{d}}), 
\label{eqqq2}
\end{align}
which, using the conditional independence properties of the HASMM (see Figure \ref{FigQ4xs}), can be simplified as follows
\begin{align}
&d\mathbb{P}(\{x^{d}_n, s^{d}_n\}_{n=1}^{k^{d}} \,|\, \{y^{d}_m, t^{d}_m\}_{m=1}^{M^{d}},\, x^{d}(T^{d}_c) = l^{d}) = \nonumber \\
&d\mathbb{P}(s^{d}_{k^{d}} \,|\, x^{d}_{k^{d}} = l^{d},\, s^{d}_{k^{d}} < T^d_c) \,\cdot\, \mathbb{P}(x^{d}_{1} \,|\, x^{d}_{2},\, s^{d}_{1} = T^d_c - (s^{d}_{2}+\,.\,.\,.\,+s^{d}_{k^d}),\, \{(y^{d}_m, t^{d}_m): t^{d}_m \in \mathcal{T}^{d}_1\}) \, \times \,\nonumber \\
&\prod_{n=2}^{k^{d}-1} d\mathbb{P}(x^{d}_{n},\, s^{d}_{n} \,|\, x^{d}_{n+1},\, s^{d}_{n} < \underbrace{T^d_c - (s^{d}_{n+1}+\,.\,.\,.\,+s^{d}_{k^d})}_{\small \mbox{Elapsed time in the episode}}, \underbrace{\{(y^{d}_m, t^{d}_m): t^{d}_m \in \mathcal{T}^{d}/\cup_{n^{'}=n+1}^{k^{d}}\mathcal{T}^d_{n^{'}}\}}_{\small \mbox{Observable variables up to state $n$}}). 
\label{LLN2}
\end{align}
From (\ref{LLN2}), we can see that for the last state in every episode $d$, i.e. state $k^d$, we already know that $x^d_{k^d} = l^{d}$, and hence the randomness is only in the last state's sojourn time $s^d_{k^d} = l^{d}$. Contrarily, for the first state, we know that conditioned on the sojourn times of the ``future states" $(s^{d}_{2},\,.\,.\,.\,,\, s^{d}_{k^d})$, the sojourn time of state $x^d_1$ is equal to $T_c^d-\sum_{n^{'}=2}^{k^d}s^{d}_{n^{'}}$ almost surely, and hence the randomness is only in the initial state realization $x^d_1$. Generally, (\ref{LLN2}) says that a sufficient statistic for the $n^{th}$ state and sojourn time is the future trajectory (starting from state $n+1$) summarized by: the next state, i.e. $x^d_{n+1}$, the observable variables up to state $n$, and the time elapsed in the episode up to state $n$, i.e. the duration of state $n$ cannot exceed the difference between the censoring time $T^d_c$ and the sojourn time of the future trajectory that stems from state $n+1$. This is captured by the last factor in (\ref{LLN2}), which explicitly specifies the likelihood of a joint realization for a state and its sojourn time conditioned on the future trajectory. Using Bayes' rule, we can further represent the last factor in (\ref{LLN2}) in terms of familiar quantities that are directly derived from the HASMM model parameters as follows 
\begin{align}
&d\mathbb{P}(x^{d}_{n}, s^{d}_{n} \,|\, x^{d}_{n+1}, s^{d}_{n} < T^d_c - (s^{d}_{n+1}+\,.\,.\,.\,+s^{d}_{K^d}), \{(y^{d}_m, t^{d}_m): t^{d}_m \in \mathcal{T}^{d}/\cup_{n^{'}=n+1}^{k^{d}}\mathcal{T}^d_{n^{'}}\}) \nonumber \\
& \propto \underbrace{\mathbb{P}(x^{d}_{n} \,|\, \{(y^{d}_m, t^{d}_m): t^{d}_m \in \mathcal{T}^{d}/\cup_{n^{'}=n+1}^{k^{d}}\mathcal{T}^d_{n^{'}}\})}_{\small \mbox{Forward message}} \, \times \, \underbrace{\mathbb{P}(x^{d}_{n+1} \,|\, x^{d}_{n}, s^{d}_{n})}_{\small \mbox{Transition function}}\, \times \, \nonumber \\
&\,\,\,\,\,\,\,\, \,\,\, \underbrace{d\mathbb{P}(s^{d}_{n} \,|\, x^{d}_{n},  s^{d}_{n} < T^d_c - (s^{d}_{n+1}+\,.\,.\,.\,+s^{d}_{k^d}))}_{\small \mbox{Truncated sojourn time distribution}}.
\label{LLN3}
\end{align}
Thus, a sampler for the latent states trajectories can be constructed using the forward messages, the HASMM's transition functions $(g_{ij}(s))_{i,j}$, and the sojourn time distributions. A compact representation for the factors in (\ref{LLN3}) is given by
\begin{align}
&\mbox{{\bf (Forward messages)}} \,\,\, \mathbb{P}(X^{d}_{n} = j \,|\, \{y^{d}_{m^{'}}, t^{d}_{m^{'}}\}^{m}_{m^{'}=1},\, \hat{\Gamma}^{z-1}) = \alpha^{d, z-1}_{m}(j),\, \forall 1 \leq m \leq M^d,\, j \in \mathcal{X}. \nonumber \\
&\mbox{{\bf (Transition functions)}} \,\,\, \mathbb{P}(X^{d}_{n+1} = j \,|\, X^{d}_{n} = i, S^{d}_{n} = s, \hat{\Gamma}^{z-1}) = g^{z-1}_{ij}(s), i, j \in \mathcal{X}, \nonumber \\ 
&\mbox{{\bf (Truncated sojourn times)}} \,\,\, d\mathbb{P}(S^{d}_{n} = s \,|\, X^{d}_{n} = j,  S^{d}_{n} < \bar{s}) = \frac{v_{j}(s|\hat{\lambda}^{z-1}_j)\, \cdot \, {\bf 1}_{\left\{s < \bar{s}\right\}}}{V_{j}(\bar{s}|\hat{\lambda}^{z-1}_j)}, j \in \mathcal{X}.\nonumber
\end{align}
Given the representations above, we can write the last factor in (\ref{LLN2}) in the $z^{th}$ iteration of the EM algorithm as follows 
\begin{align}
&d\mathbb{P}(x^{d}_{n}, s^{d}_{n}\,|\, x^{d}_{n+1}, s^{d}_{n} < \bar{s}, \{y^{d}_m, t^{d}_m\},\, \hat{\Gamma}^{z-1}) \propto \alpha^{d, z-1}_{m}(x^{d}_{n}) \, \cdot \, g^{z-1}_{x^{d}_{n},x^{d}_{n+1}}(s^{d}_{n}) \, \cdot \, \frac{v_{x^{d}_{n}}(s^{d}_{n}|\hat{\lambda}^{z-1}_{x^{d}_{n}})\, \cdot \, {\bf 1}_{\left\{s^{d}_{n} \leq \bar{s}\right\}}}{V_{x^{d}_{n}}(\bar{s}|\hat{\lambda}^{z-1}_{x^{d}_{n}})}. 
\label{eqqwqq1}
\end{align}

From the factor decomposition in (\ref{LLN3}), we can see that informative censoring allows us to construct a sampler for the latent state trajectories that operates sequentially in the reverse time direction by sampling from the posterior probability of every state $n$ given the future trajectory of states that starts from state $n+1$. From (\ref{eqqwqq1}), we note that the posterior density of the latent states conditioned on the future trajectory, from which sequential sampling is conducted, can be explicitly decomposed in terms of the HASMM parameters. A complete recipe for the forward-filtering backward-sampling procedure for sampling trajectories from the posterior density $d\mathbb{P}(\{x^{d}_n, s^{d}_n\}_{n=1}^{k^{d}} \,|\, \{y^{d}_m, t^{d}_m\}_{m=1}^{M^{d}},\, x^{d}(T^{d}_c) = l^{d},\, \hat{\Gamma}^{z-1})$ using the decomposition in (\ref{LLN3}) and the posterior density in (\ref{eqqwqq1}) is provided as follows:
\begin{itemize}
\item {\bf Forward filtering pass:}\\
\\
For every episode $d$ in $\mathcal{D}$, compute the forward messages $\{\alpha^{d, z-1}_{m}(j)\}$ for all time instances $t^{d}_m \in \mathcal{T}^{d}$ using the current estimate for the parameter set $\hat{\Gamma}^{z-1}$, i.e. invoke the routine \texttt{ForwardFilter}($\hat{\Gamma}^{z-1}, \{y^{d}_m, t^{d}_m\}^{M^{d}}_{m=1}, \epsilon$). 
\item {\bf Backward sampling pass:}\\
\\
For every episode $d$ in $\mathcal{D}$, carry out the following steps: 
\begin{enumerate}
\item Set a dummy {\it placeholder index} as $k^{\#} = 1$ and set $u_{k^{\#}} = l^d$.
\item Sample a Bernoulli random variable $B_{k^{\#}} \sim \mbox{Bernoulli}(\mathbb{P}(k^d=k^{\#}\,|\,\{u_{k},w_k\}^{k^{\#}-1}_{k=1},\, l^d))$.
\item If $B_{k^{\#}}=0$ and $k^{\#}>1$, sample a bivariate random variable $(u_{k^{\#}}, w_{k^{\#}})$ as follows
\[(u_{k^{\#}},\, w_{k^{\#}}) \sim \frac{1}{\mathcal{U}}\left(\alpha^{d, z-1}_{m}(u_{k^{\#}}) \, \cdot \, g^{z-1}_{u_{k^{\#}},u_{k^{\#}-1}}(w_{k^{\#}}) \, \cdot \, \frac{v_{u_{k^{\#}}}(w_{k^{\#}}|\hat{\lambda}^{z-1}_{u_{k^{\#}}})\, \cdot \, {\bf 1}_{\left\{w_{k^{\#}} \leq \bar{s}\right\}}}{V_{u_{k^{\#}}}(\bar{s}|\hat{\lambda}^{z-1}_{u_{k^{\#}}})}\right),\]
where $\mathcal{U} = \sum_{u}\int_{w}\alpha^{d, z-1}_{m}(u) \, \cdot \, g^{z-1}_{u,u_{k^{\#}-1}}(w) \, \cdot \, \frac{v_{u}(w|\hat{\lambda}^{z-1}_{u})\, \cdot \, {\bf 1}_{\left\{w \leq \bar{s}\right\}}}{V_{u}(\bar{s}|\hat{\lambda}^{z-1}_{u})}$, $\bar{s} = T_c^d-\sum_{n^{'}=1}^{k^{\#}-1}w_{n^{'}},$ and $m = \arg \, \max_{m^{'}}\, \{\mathcal{T}:\, t_{m^{'}} \leq \bar{s}\}$. If $B_{k^{\#}}=0$ and $k^{\#}=1$, then sample $w_{k^{\#}}$ as follows 
\[w_{k^{\#}} \sim \frac{v_{u_{k^{\#}}}(w_{k^{\#}}|\hat{\lambda}^{z-1}_{u_{k^{\#}}})\, \cdot \, {\bf 1}_{\left\{w_{k^{\#}} \leq T^d_c\right\}}}{V_{u_{k^{\#}}}(T^d_c|\hat{\lambda}^{z-1}_{u_{k^{\#}}})}.\]  
\item If $B_{k^{\#}}=1$, then set $w_{k^{\#}} = \bar{s}$. If $k^{\#}>1$, then sample $u_{k^{\#}}$ as follows 
\[u_{k^{\#}} \sim \frac{d\mathbb{P}(\{y^d_{m^{'}}, t^d_{m^{'}}\}^{m}_{m^{'}=1}\,|\, u_{k^{\#}})\, \cdot\, g_{u_{k^{\#}},u_{k^{\#}-1}}(\bar{s})\,\cdot\, v_{u_{k^{\#}}}(\bar{s}|\hat{\lambda}^{z-1}_{u_{k^{\#}}})\, \cdot\, \hat{p}^{o,z-1}_{u_{k^{\#}}}}{\sum_{u} d\mathbb{P}(\{y^d_{m^{'}}, t^d_{m^{'}}\}^{m}_{m^{'}=1}\,|\, u)\, \cdot\, g_{u,u_{k^{\#}-1}}(\bar{s})\,\cdot\, v_{u}(\bar{s}|\hat{\lambda}^{z-1}_{u})\, \cdot\, \hat{p}^{o,z-1}_{u}}.\]
\item If $B_{k^{\#}}=0$, then increment the placeholder index $k^{\#}$ and go to step 2 and repeat the consequent steps.
\item If $B_{k^{\#}}=1$, then set $k^d = k^{\#}$ and terminate the sampling process for episode $d$. Set the sampled trajectory by swapping the bivariate sequence $(u_{k^{\#}}, w_{k^{\#}})$ as follows: $(x^d_{n}, s^d_n) = (u_{k^{\#}-n+1}, w_{k^{\#}-n+1}), \forall n \in \{1,.\,.\,.,k^{\#}\}.$
\end{enumerate}
\end{itemize}

\begin{algorithm}[t]
  \caption{Truncated Rejection Sampler}\label{alg4}
  \begin{algorithmic}[1]
    \Procedure{\texttt{TRSampler}}{$\Gamma$, $u$, $\bar{s}$}
		  \State {\bfseries Input:} A parameter set $\Gamma$, a state $u$ and a truncation threshold $\bar{s}$
	    \State {\bfseries Output:} A random variable $s$ 
			\State $k \gets 0$
			\While{$k = 0$}
				\State $s \sim v_{u}(s|\lambda_u)$
				\State Accept $s$ and set $k \gets 1$ if $s < \bar{s}$. Reject $s$ otherwise. 
			\EndWhile
      \State \textbf{return} $s$
    \EndProcedure
  \end{algorithmic}
\end{algorithm}

\begin{algorithm}[t]
  \caption{Bivariate Adaptive Rejection Sampler}\label{alg5}
  \begin{algorithmic}[1]
    \Procedure{\texttt{BARSampler}}{$\{\alpha(j)\}_{j=1}^{N}$, $\Gamma$, $u^{'}$, $\bar{w}$}
		  \State {\bfseries Input:} A set of $N$ forward messages $\{\alpha(j)\}_{j=1}^{N}$, parameter set $\Gamma$, and a state $u^{'}$
	    \State {\bfseries Output:} A bivariate conditional random variable $(u,w)|u^{'}$ 
			\State $k \gets 0$
			\While{$k = 0$}
				\State $u \sim \mbox{Multinomial}(\alpha(1),.\,.\,.,\alpha(N))$
				\State $w =$ \texttt{TRSampler}$(\Gamma, u, \bar{w})$
				\State $\tilde{u} \sim \mbox{Multinomial}(g_{u1}(w),.\,.\,.,g_{uN}(w))$
				\State Accept $(u,w)$ and set $k \gets 1$ if $\tilde{u} = u^{'}$. Reject $(u,w)$ otherwise. 
			\EndWhile
      \State \textbf{return} $(u,w)$
    \EndProcedure
  \end{algorithmic}
\end{algorithm}

\begin{algorithm}[t]
  \caption{A sampler for latent state trajectories}\label{alg6}
  \begin{algorithmic}[1]
    \Procedure{\texttt{BackwardSampling}}{$\Gamma, \{\{\alpha^{d,o}_m(j)\}_{m,j}\}_d, \{y^d_m, t^d_m\}^{M_d}_{m=1}, x^d(T^d_c) = l^d$}
		  \State {\bfseries Input:} Parameter $\Gamma$, forward messages, observables, and censoring information 
	    \State {\bfseries Output:} A sampled latent state trajectory $\{x^{d}_n, s^{d}_n\}_{n=1}^{k^d}$
			\State $k^{\#} \gets 1$, $u_{k^\#} \gets l^d$, $B_{k^{\#}} \sim \mbox{Bernoulli}(\mathbb{P}(k^d = k^{\#}|\{y^d_m, t^d_m\}^{M_d}_{m=1}, x^d(T^d_c) = l^d))$
			\If{$B_{k^{\#}}=0$} 
			\State $w_{k^{\#}} =$ \texttt{TRSampler}$(\Gamma,u_{k^\#}, T^d_c)$
			\State $k^\# \gets k^\#+1$
			\Else 
			\State $w_{k^{\#}} =T^d_c$, $k^d = 1$, $\{x^d_1, s^d_1\} \gets \{u_{k^\#}, w_{k^\#}\}$
			\State Terminate \texttt{BackwardSampling}.
			\EndIf
			\While{$k^\# >0$}
			  \State $B_{k^{\#}} \sim \mbox{Bernoulli}(\mathbb{P}(k^d = k^{\#}|\{u_{k},w_{k}\}^{k^\#-1}_{k=1},\{y^d_m, t^d_m\}^{M_d}_{m=1}))$
				\State $\bar{s} = T^d_c - \sum^{k^{\#}-1}_{n^{'}=1}w_{k^{\#}}$
				\If{$B_{k^{\#}}=0$}
				\State $(u_{k^\#}, w_{k^\#}) \gets$ \texttt{BARSampler}$(\{\alpha_m^{d,o}(j)\}_j, \Gamma, u_{k^\#-1}, \bar{s})$ 
				\State $k^\# \gets k^\#+1$
				\Else
				\State Sample the initial state $u_{k^\#}$, set $w_{k^\#} \gets \bar{s}$
				\State $\{x^{d}_n,s^{d}_n\} = \{u_{k^\#-n+1}, w_{k^\#-n+1}\},\, \forall n \in \{1,.\,.\,.,k^{\#}\}$
				\State $k^\# \gets -1$
				\EndIf
			\EndWhile
      \State \textbf{return} $\{x^{d}_n, s^{d}_n\}_{n=1}^{k^d}$
    \EndProcedure
  \end{algorithmic}
\end{algorithm}

\begin{algorithm}[t]
  \caption{Forward-filtering Backward-sampling Monte Carlo EM Algorithm}\label{alg7}
  \begin{algorithmic}[1]
    \Procedure{\texttt{FFBS-MCEM}}{$\mathcal{D}$, $G$, $\epsilon$}
		  \State {\bfseries Input:} A dataset $\mathcal{D}$, number of Monte Carlo samples $G$, and a precision level $\epsilon$
	    \State {\bfseries Output:} An estimate $\hat{\Gamma}$ for the HASMM parameters
			\State Set an initial value $\hat{\Gamma}^{o}$ for the HASMM parameters 
			\State $\{\alpha^{d,o}_{m}\}^{M_d}_{m=1} =$ \texttt{ForwardFilter}$(\hat{\Gamma}^{o}, \{y^d_m, t^d_m\}^{M_d}_{m=1}, \epsilon),\, \forall 1\leq d \leq D$ \Comment{Forward pass}	
			\For{$d$ = 1 to $D$} \Comment{Backward pass: sample $G$ latent state trajectories}
			\For{$g$ = 1 to $G$}
			\State $\{x^{d,g}_n, s^{d,g}_n\}_{n=1}^{k^{d,g}} =$ \texttt{BackwardSampling}$(\hat{\Gamma}^{o}, \{y^d_m, t^d_m\}^{M_d}_{m=1}, x^d(T^d_c) = l^d)$  
			\EndFor
			\EndFor
			\State $z \gets 1$
			\State $E \gets \epsilon + 1$
			\While{$E > \epsilon$}
			  \State $I^{z-1}_{d,g} \gets d\mathbb{P}(\{x^{d,g}_n, s^{d,g}_n\}_{n=1}^{k^{d,g}}\,|\,\hat{\Gamma}^{z-1})/d\mathbb{P}(\{x^{d,g}_n, s^{d,g}_n\}_{n=1}^{k^{d,g}}\,|\,\hat{\Gamma}^{o})$ \Comment{Importance weights}
			  \State $\hat{Q}_{G}(\Gamma; \hat{\Gamma}^{z-1}) = \sum_{d, g} \log(d\mathbb{P}(\{x^{d, g}_n, s^{d, g}_n\}_{n=1}^{k^{d, g}}, \{y^{d}_m, t^{d}_m\}_{m=1}^{M^{d}}\,|\,\Gamma ))\,\cdot\,\frac{I^{z-1}_{d,g}}{G}$ \Comment{E-step}
				\State $\hat{\Gamma}^{z} = \arg \,\max_{\Gamma} \hat{Q}_{G}(\Gamma; \hat{\Gamma}^{z-1})$ \Comment{M-step}
				\State $z \gets z + 1$
			\EndWhile
      \State \textbf{return} $\hat{\Gamma} = \hat{\Gamma}^{z}$ %
    \EndProcedure
  \end{algorithmic}
\end{algorithm} 

\begin{figure*}[t]
        \centering
        \includegraphics[width=6in]{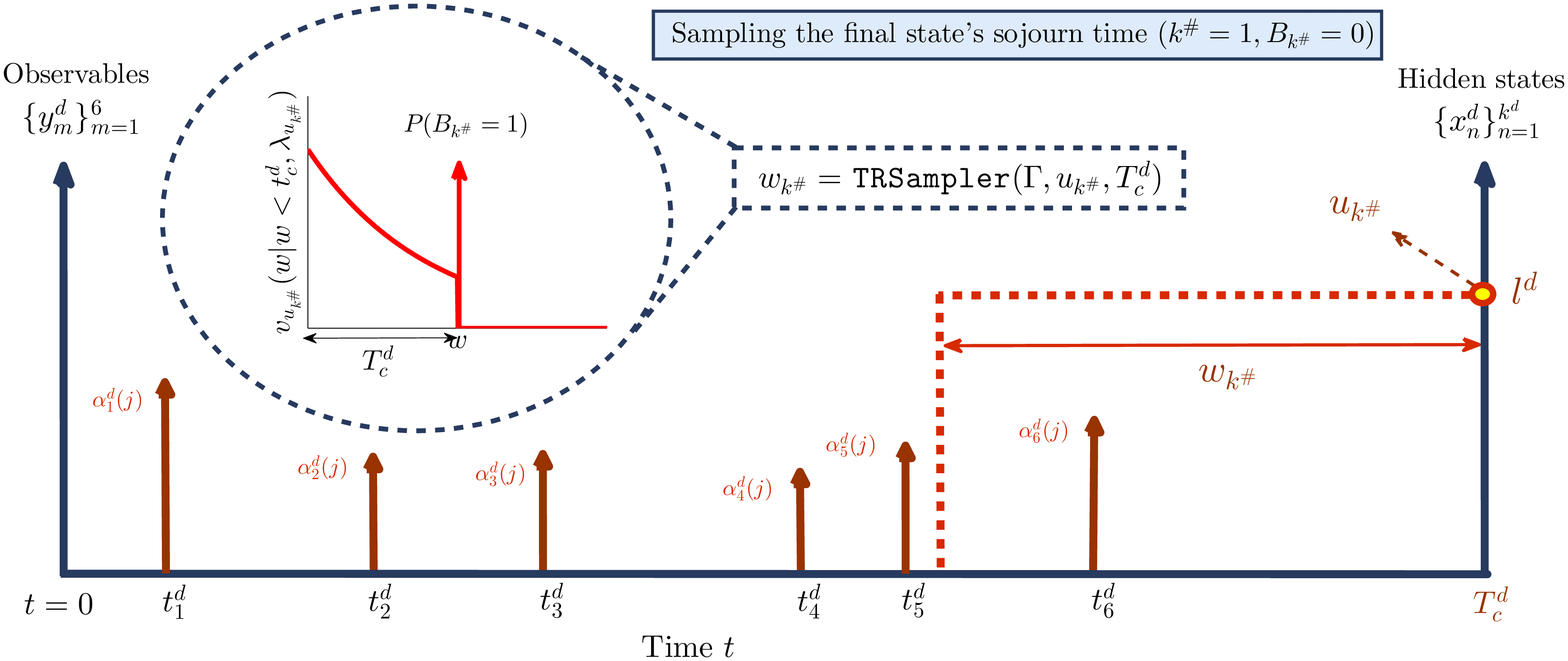}
				\captionsetup{font= small}
        \caption{Depiction of the backward sampling pass for the last state of an episode $d$.}
				\label{seMC0}
				\centering
        \includegraphics[width=6in]{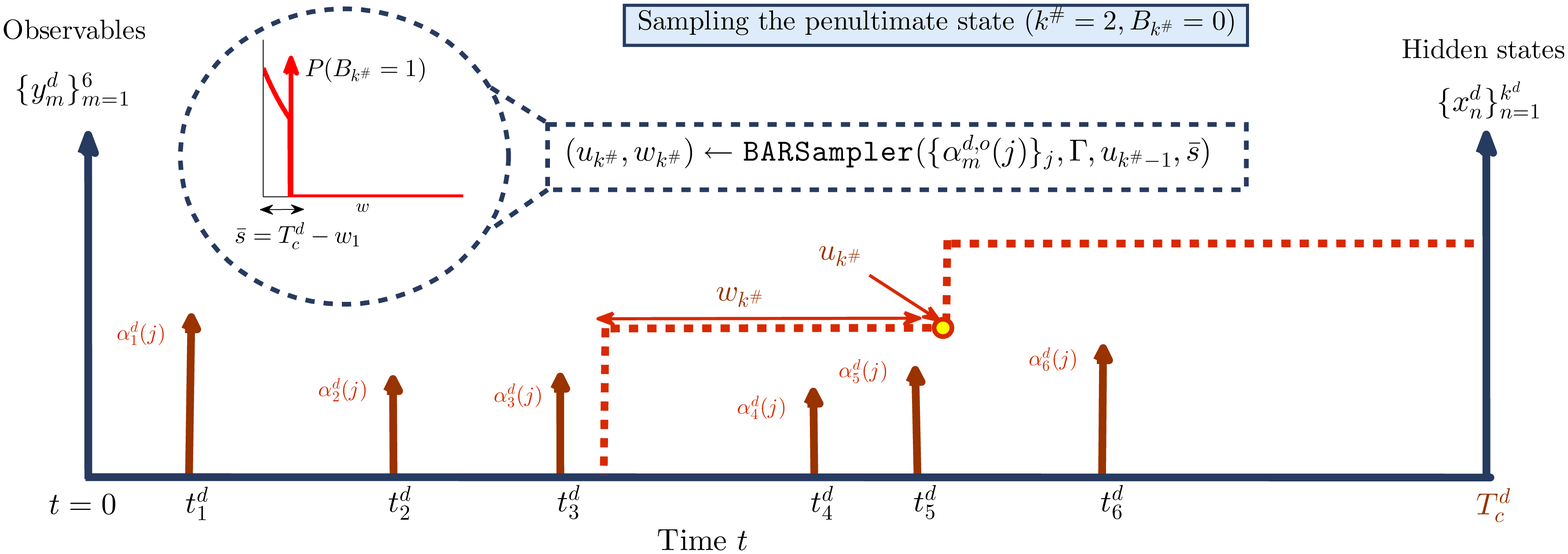}
				\captionsetup{font= small}
        \caption{Depiction of the backward sampling pass for the penultimate state after having sampled the last state as depicted in the Figure above.}
\label{seMC}
\end{figure*}

The forward-filtering backward-sampling procedure constitutes of a forward pass in which we compute the forward messages for all the data points in $\mathcal{D}$ using the dynamic programming algorithms presented in Section \ref{sec3}, and a backward pass in which these forward messages are used to sample latent state trajectories. The backward sampling procedure for every episode goes as follows. We start from the censoring time at which we know what state has actually materialized, i.e. the absorbing state. Since we do not know the number of states in the state trajectory, we initialize a placeholder index $k^{\#} = 1$ as an index for the absorbing state, and increment it whenever a new state is sampled. We start the sampling procedure as follows. Given the the censoring variables and the observable time series, we sample the sojourn time of the last state (the absorbing state): this is sampled from a truncated sojourn time distribution, with a truncation threshold at $T^d_c$, and a point mass at $T^d_c$ with an assigned measure that is equal to the posterior probability of the absorbing state being the initial state as depicted in Figure \ref{seMC0}. This is implemented by first sampling a Bernoulli random variable $B_{k^{\#}}$ with a success probability equal to the posterior probability of the absorbing state being the initial state, and then sampling the truncated sojourn time if $B_{k^{\#}} = 0$ using the simple rejection sample executed by the routine \texttt{TRSampler} which is provided in Algorithm \ref{alg4}. Having sampled the last state's sojourn time, we sample the penultimate state and its sojourn time jointly using the routine \texttt{BARSampler} (Algorithm \ref{alg5}) as depicted in Figure \ref{seMC}. The routine \texttt{BARSampler} uses a sampling algorithm, that we call the {\it bivariate adaptive rejection sampler}, which jointly samples the current state and its sojourn time given the next state as follows. First, a state is sampled from a Multinomial distribution with probability masses equal to the forward messages. Next, given the sampled state, we sample a sojourn time from the truncated sojourn time distribution. Finally, given the sampled state and the sampled sojourn time, we sample a dummy state from a Multinomial whose masses are equal to the transition functions, and we accept the sample only if the sampled dummy state is equal to the next state. It can be easily proven that \texttt{BARSampler} generates samples that are equal in distribution to the true state trajectory.\\     

The backward-sampling procedure operates sequentially by invoking the \texttt{BARSampler} to generate new state and sojourn times samples conditional on the previously sampled (future) states. The process terminates whenever $B_{k^{\#}} = 1$, i.e. a state is sampled as an ``initial state". The routine \texttt{BackwardSampling} (Algorithm \ref{alg6}) implements the overall backward-sampling procedure for every episode in $\mathcal{D}$. \\
  
Note that, unlike the slowly mixing MCMC methods, the backward-sampling algorithm can generate the latent state trajectory in an efficient manner, i.e. the run-time of the backward-sampling algorithm is stochastically dominated by a geometrically-distributed random variable with a success probability that, other than in a pathological HASMM parameter settings, would not be close to zero. Moreover, since \texttt{BackwardSampling} generates independent samples, no wasteful burn-in sampling iterations are involved in the FFBS-MCEM operation. We provide a pseudocode for the overall operation of the \texttt{FFBS-MCEM} algorithm in Algorithm \ref{alg7}. We omit the standard EM operations, such as the implementation of the M-step, for the sake of brevity. \\
\\
In Algorithm \ref{alg7}, we avoid the need for running the routine \texttt{BackwardSampling} in every iteration of the EM algorithm by re-using the sampled trajectories based on the initial parameter guess $\hat{\Gamma}^{o}$ through the usage of importance weights in the E-step. That is, in the $z^{th}$ iteration of the EM algorithm, we implement the E-step as follows (\cite{booth1999maximizing}) \[\hat{Q}_{G}(\Gamma; \hat{\Gamma}^{z-1}) = \sum_{d, g} \log(d\mathbb{P}(\{x^{d, g}_n, s^{d, g}_n\}_{n=1}^{k^{d, g}}, \{y^{d}_m, t^{d}_m\}_{m=1}^{M^{d}}\,|\,\Gamma))\,\cdot\, \underbrace{\frac{d\mathbb{P}(\{x^{d,g}_n, s^{d,g}_n\}_{n=1}^{k^{d,g}}\,|\,\hat{\Gamma}^{z-1})}{d\mathbb{P}(\{x^{d,g}_n, s^{d,g}_n\}_{n=1}^{k^{d,g}}\,|\,\hat{\Gamma}^{o})}}_{\small \mbox{Importance weights}}.\]
This implementation for the E-step offers a tremendous advantage in the computational cost of \texttt{FFBS-MCEM}. By using importance weights, we need to compute the forward messages and sample the latent state trajectories only once, and then reuse the sampled trajectories in all the subsequent EM iterations. In the following Theorem, we prove that \texttt{FFBS-MCEM} is as accurate as an EM algorithm with an exact implementation for the E-step when the number of Monte Carlo samples $G$ grows asymptotically large. 

\begin{theorem}[Convergence properties of FFBS-MCEM]
\label{thm5} 
The sequence of parameter set estimates $\{\hat{\Gamma}^z\}$ computed by the FFBS-MCEM algorithm converges in probability, i.e. $\hat{\Gamma}^z \overset{\tiny p}{\rightarrow}\, \bar{\Gamma}$. Furthermore, if $\Gamma^{*}$ is a local maximizer of $\Lambda(\mathcal{D}| \Gamma)$, then there exists a neighborhood of $\Gamma^{*}$ such that for any initial guess $\hat{\Gamma}^{o}$ in that neighborhood and for any $\epsilon > 0$ we have that
\[\lim_{G \uparrow \infty}\mathbb{P}(||\hat{\Gamma}^z - \Gamma^*|| < \epsilon) \rightarrow 1.\]
\begin{proof}
See Appendix \ref{appE}. 
\end{proof}
\end{theorem} 
In the next Section, we highlight the merits of our model and the associated algorithms through experiments conducted on a real-world dataset of informatively censored clinical time series data.

\section{Experiments: Intensive Care Unit Prognostication}
\label{sec4}
We investigate the utility of the HASMM in the setting of ICU prognostication; we use the HASMM as a model for the physiology of critically ill patients in regular hospital wards who are monitored for various vital signs and lab tests. Through the HASMM, we construct a risk score (based on the analysis in Section \ref{sec3.33}) that assesses the risk of clinical deterioration for the monitored patients, which allows for timely ICU admission whenever clinical decompensation is detected. Risk scoring in hospital wards and ICU admission management is a pressing problem with a huge social and clinical impact: qualitative medical studies have suggested that up to 50$\%$ of cardiac arrests on general wards could be prevented by earlier transfer to the ICU (\cite{hershey1982outcome}). Since over 200,000 in-hospital cardiac arrests occur in the U.S. each year with a mortality rate of 75$\%$ (\cite{merchant2011incidence}), improved patient monitoring and vigilant care in wards enabled by the HASMM would translate to a large number of lives saved yearly.

\subsection{Data}
\subsubsection{The Patients' Cohort}
Experiments were conducted on a heterogeneous cohort of 6,094 episodes for patients who were hospitalized in Ronald Reagan UCLA medical center during the period between March 3$^{rd}$, 2013 to March 29$^{th}$, 2016. The patients' population is heterogeneous: we considered admissions to all the floors and units in the medical center, those include the acute care pediatrics unit, cardiac observation unit, cardiothoracic unit, hematology and stem cell transplant unit and the liver transplant service. Patients admitted to those floors (or wards) are post-operative or pre-operative critically ill patients who are vulnerable to adverse clinical outcomes that may require an impending ICU transfer. The cohort comprised patients with a wide variety of ICD-9 codes and medical conditions, including leukemia, hypertension, septicemia, sepsis, abdomen and pelvis, pneumonia, and renal failure. Table \ref{Table2ICU} shows the distribution of the most common ICD-9 codes in the patient cohort together with the corresponding medical conditions. The notable heterogeneity of the cohort suggests that the results presented in this Section are generalizable to different cohorts extracted from different hospitals.
\smallskip
\begin{sidewaystable}
    \centering
    \caption{Characteristics of the patient cohort under study}
\begin{tabular}{|c||c||>{\centering\arraybackslash}m{4.5in}|}  
\toprule[2.5pt]
\multicolumn{2}{c}{{\bf Physiological data}} & {\bf ICD-9 codes' distribution}  \\
\cmidrule[1.5pt]{1-3}
{\bf Vital signs} & {\bf Lab tests} \\
\cmidrule[1.5pt]{1-2}%\cline{1-2}  
Diastolic blood pressure & Chloride & \multirow{10}{*}{\begin{minipage}{.5\textwidth}
      \includegraphics[width=4in]{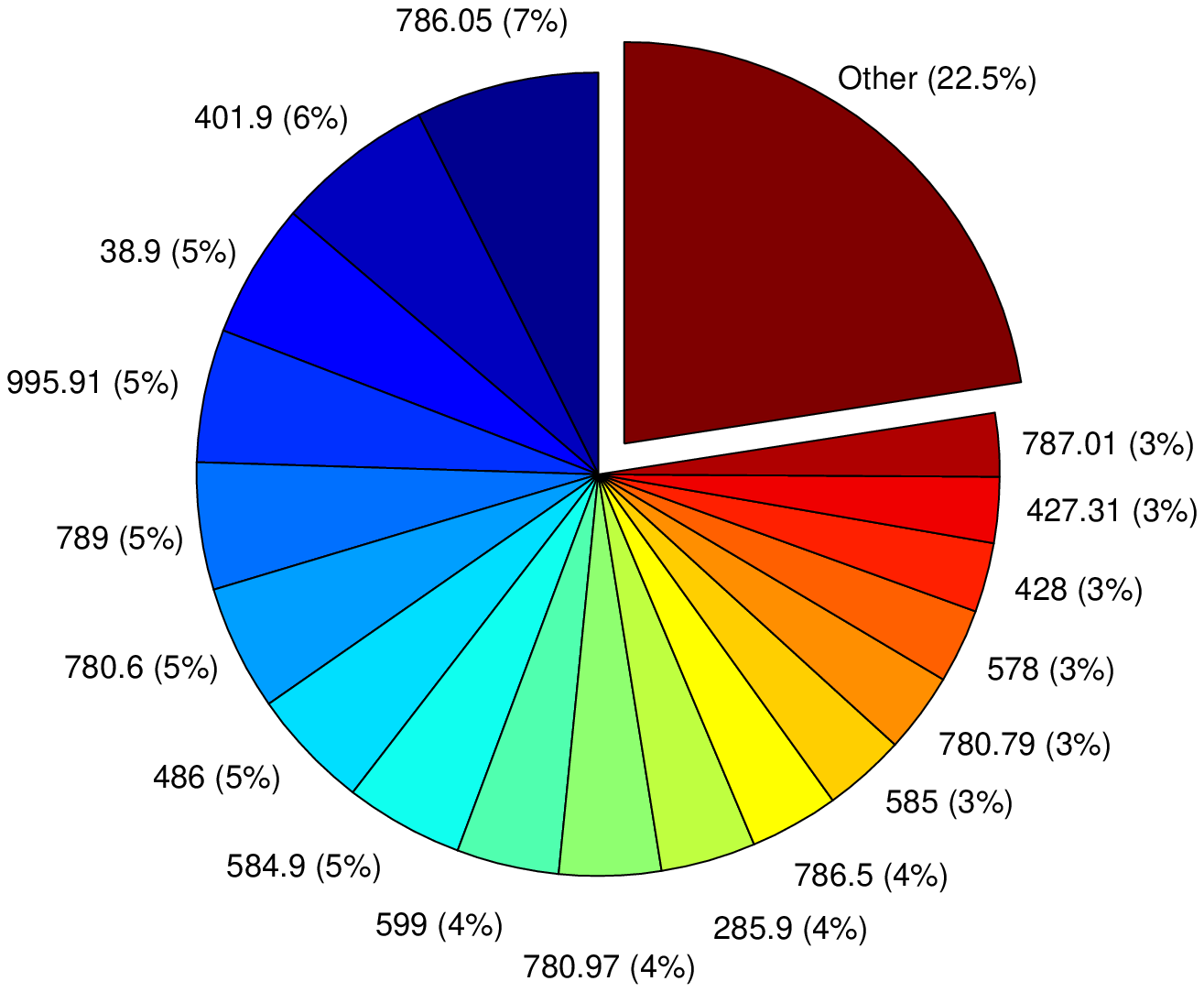}
    \end{minipage}}     \\
Eye opening & Glucose &       \\
Glasgow coma scale score & Urea Nitrogen &    \\
Heart rate & White blood cell count &    \\
Respiratory rate & Creatinine &     \\
Temperature & Hemoglobin &     \\
$O_2$ Device Assistance & Platelet Count &     \\
$O_2$ Saturation & Potassium &     \\
Best motor response & Saturation Sodium &     \\
Best verbal response & Total $CO_2$ &     \\
Systolic blood pressure &  &  \\ %\cline{1-2} 
\cmidrule[1.5pt]{1-2}
\multicolumn{2}{c}{{\bf ICD-9 codes}} & \\
\cmidrule[1.5pt]{1-2}
{\bf (786.05)} & Shortness of Breath & \\ \cline{1-2} 
{\bf (401.9)} & Hypertension & \\ \cline{1-2}
{\bf (38.9)} & Septicemia & \\ \cline{1-2}
{\bf (995.91)} & Sepsis & \\ \cline{1-2}
{\bf (789)} & Abdomen and pelvis & \\ \cline{1-2}
{\bf (780.6)} & Fever & \\ \cline{1-2}
{\bf (486)} & Pneumonia & \\ \cmidrule[0.5pt]{1-2} \cmidrule[1.5pt]{3-3}  
{\bf (584.9)} & Renal failure & {\bf Baseline Patient Characteristics (with 95$\%$ CI)} \\ \cmidrule[0.5pt]{1-2} \cmidrule[1.5pt]{3-3} 
{\bf (599)} & Urethra and urinary attack & $\bullet$ {\bf Gender distribution (Male percentage)}\\ \cline{1-2}
{\bf (780.97)} & Altered mental status & (Training: 50.31$\%$ $\Mypm$ 1.4$\%$ - Testing: 51.16$\%$ $\Mypm$ 2.92$\%$)\\ \cline{1-3}
{\bf (285.9)} & Anemia & $\bullet$ {\bf Transfers from other hospitals} \\ \cline{1-2}
{\bf (786.5)} & Chest pain & (Training: 11.88$\%$ $\Mypm$ 0.94$\%$ - Testing: 11.08$\%$ $\Mypm$ 1.95$\%$)\\ \cline{1-3}
{\bf (585)} & Chronic renal failure & $\bullet$ {\bf Average age} \\ \cline{1-2}
{\bf (780.79)} & Malaise and fatigue & (Training: 58.9 $\Mypm$ 0.55 years - Testing: 59.37 $\Mypm$ 1.11 years)\\ \cline{1-3}
{\bf (578)} & Gastrointestinal hemorrhage & $\bullet$ {\bf Patients with chemotherapy} \\ \cline{1-2} 
{\bf (428)} & Heart failure & (Training: 0.688$\%$ $\Mypm$ 0.272$\%$ - Testing: 1.558$\%$ $\Mypm$ 0.9$\%$)\\ \cline{1-3}
{\bf (427.31)} & Atrial fibrillation & $\bullet$ {\bf Patients with stem cell transplants} \\ \cline{1-2}
{\bf (787.01)} & Nausea & (Training: 0.121$\%$ $\Mypm$ 0.8$\%$ - Testing: 0.008$\%$ $\Mypm$ 0.004$\%$) \\
\bottomrule[2.5pt]
\end{tabular}
\label{Table2ICU}
\end{sidewaystable}
Every patient in the cohort is associated with a set of 21 (temporal) physiological streams comprising a set of vital signs and lab tests that are listed in Table \ref{Table2ICU}. The physiological measurements are gathered over time during the patient's stay in the ward, and they manifest -in a subtle fashion- the patient's clinical state. The physiological measurements are collected over irregularly spaced time intervals (usually ranging from 1 to 4 hours); for each physiological time series, we have access to the times at which each value was gathered.\\
\\
In all the experiments hereafter, we split the patient cohort into a training set and a testing set. The training set comprises 4,939 patients admitted to the medical center in the period between March 3$^{rd}$, 2013 to November 1$^{st}$, 2015; the testing set comprises 1,155 patients admitted in the period between November 1$^{st}$, 2015 to March 29$^{th}$, 2016. This split of the data allows us to assess the performance under the realistic scenario when a certain algorithm learns from the data available up to a certain date, and then is used to assess the risk for patients admitted in future dates. In Table \ref{Table2ICU}, we show statistics for the patients' baseline static features (e.g. gender, age, etc) in both the training and testing sets; as we can see, the characteristics of the patients admitted in the period (March 2013 - November 2015) has not significantly changed from those admitted in the period (November 2015 - March 2016). We have verified this fact using a two-sample $t$-test through which we compared the expected values of the baseline co-variates in both the training and testing sets. This means that the hospital's management policy with respect to the patients' acceptance and triaging has not significantly changed across the two time periods, and hence whatever is learned from the training data can be sensibly applied to the testing data.
\begin{figure*}[t]
        \centering
        \includegraphics[width=6in]{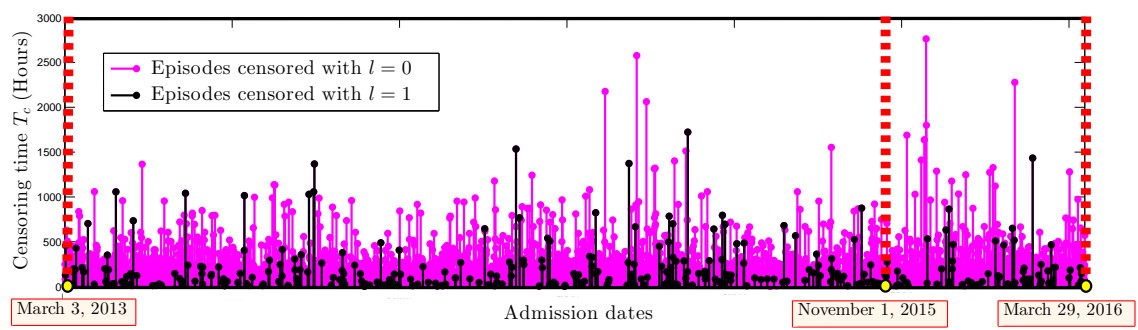}
				\captionsetup{font= small}
        \caption{Visualization for the episodes' censoring information.}
			  \label{Eps1}
\end{figure*}
\subsubsection{Informative Censoring}
All the patient episodes in the cohort were informatively censored. That is, for every patient in the cohort, we know the following information:\\
{\bf $\bullet$ The censoring time ($T_c$):} the length of stay of each patient in the ward is recorded in the dataset, and hence we have access to the HASMM's censoring time variable $T_c$. The average hospitalization time (or censoring time) in the cohort is 157 hours and 34 minutes (6.5 days). The patient episodes' censoring times ranged from 4 hours to 2672 hours. \\
\\
{\bf $\bullet$ The absorbing clinical state ($l$):} with the help of experts from the division of pulmonary and critical care medicine at Ronald Reagan UCLA medical center, we set the value of the variable $l$ (absorbing state) for every patient's episode based on the clinicians' interventions as reported in the dataset. That is, as advised by our medical collaborators, we assigned the label $l=1$ to every patient who was admitted to the ICU and underwent an intervention in the ICU (e.g. ventilator, drug, etc), or was reported to exhibit a cardiac or respiratory arrest (before or after the ICU transfer). According to the medical experts, those patients have experienced ``clinical deterioration" as their absorbing state, and would have benefited from an earlier admission to the ICU. We have excluded all patients who underwent a preplanned ICU admission from the dataset since those patients did not actually experience clinical deterioration, but were transferred routinely to the ICU after a surgery. We assigned the label $l=0$ to all patients who were discharged home after the clinician's in charge realized they were clinically stable. Since the readmission rate at the UCLA medical center is quite low, our medical collaborators believe that the labels $l=1$ and $l=0$ represent an accurate representation for the patients' true absorbing clinical states upon censoring. \\ 
\\
Patient episodes with the absorbing state $l=0$ had an average censoring time of 155 hours on average, whereas those with $l=1$ had an average censoring time of 204 hours. The percentage of episodes with an absorbing state $l=1$ was 4.98$\%$ $\Mypm$ 0.64$\%$ in the training period (March 2013 - November 2015), and was 5.19$\%$ $\Mypm$ 1.44$\%$ in the testing period (November 2015 - March 2016). A two-sample $t$-test reveals that the censoring information (distributions of $T_c$ and $l$) has not significantly changed from the training to testing periods, which suggests that the HASMM learned from the training data can be sensibly applied to the testing data. Figure \ref{Eps1} visualizes the informative censoring information over the time period between March 2013 and March 2016. Every patient episode, starting at a certain admission date, is represented by its censoring time (hospitalization time); light colored episodes are ones that were absorbed in the clinical stability state ($l=0$), whereas dark colored ones were absorbed in the clinical deterioration state ($l=1$).  
\subsection{Baseline Algorithms}
We compare the HASMM, as a model for the patients' episodes based on which an early warning system is constructed, with other baseline early warning methods. The comparisons involve both state-of-the-art clinical risk scores that are currently used in various healthcare facilities around the world, in addition to benchmark machine learning algorithms. The details of the baselines are provided in the following subsections. 
\subsubsection{State-of-the-art Clinical Risk Scores}
We have conducted comparisons with the most prominent clinical risk scores currently deployed in major healthcare facilities. We list the clinical risk scores involved in our comparisons as follows.
\begin{enumerate}[(i)]
\item {\bf Modified Early Warning System (MEWS)}: a risk scoring scheme used currently by many healthcare facilities and rapid response teams to quickly assess the severity of illness of a hospitalized patient (\cite{morgan1997early}). The score ranges from 0 to 3 and is based on the following cardinal vital signs: systolic blood pressure, respiratory rate, $SaO_2$, temperature, and heart rate. 
\item {\bf Sequential Organ Failure Assessment (SOFA)}: a risk score (ranging from 1 to 4) that is used to determine the extent of a hospitalized patient's respiratory, cardiovascular, hepatic, coagulation, renal and neurological organ function in the ICU (\cite{vincent1996sofa}). 
\item {\bf Acute Physiology and Chronic Health Evaluation (APACHE II)}: a risk scoring system (an integer score from 0 to 71) for predicting mortality of patients in the ICU (\cite{knaus1991apache}). The score is based on 12 physiological measurements, including creatinine, white blood cell count, and glasgow coma scale.
\item {\bf Rothman Index}: a regression-based data-driven risk score that utilizes physiological data to predict mortality, 30-days readmission, and ICU admissions for patients in regular wards (\cite{rothman2013development}). The Rothman index is the state-of-the-art risk score for regular ward patients and is currently used in more than 70 hospitals in the US, including the Houston Methodist hospital in Texas and the Yale-New Haven hospital in Connecticut (\cite{WSJ}). At the time of conducting these experiments, the Rothman index was also deployed in the Ronald Reagan UCLA medical center.
\end{enumerate}
We implemented the MEWS, SOFA, APACHE II and Rothman scores according to the specifications in (\cite{vincent1996sofa, knaus1991apache, rothman2013development}). Note that while the SOFA and APACHE II scores are usually deployed for patients in the ICU, both scores have been recently shown to provide a prognostic utility for predicting clinical deterioration for patients in regular wards (\cite{yu2014comparison}), and hence we consider both scores in our comparisons.  
\subsubsection{Machine Learning Algorithms}
In order to demonstrate the modeling gain of HASMMs, we make comparisons with other competing machine learning algorithms that adopt different modeling approaches for the clinical time series data. The details of these competing models are provided in the following.\\
\smallskip
\smallskip
\\
{\bf \underline{Discriminative models}:}\\
\smallskip
We consider the following set of discriminative predictors that directly predict clinical deterioration without explicitly modeling the clinical time series data: 
\begin{itemize}
\item {\bf Logistic regression.} 
\item {\bf Least absolute shrinkage and selection operator (LASSO).}
\item {\bf Random forest.}
\item {\bf Recurrent Neural Networks (RNN).}
\end{itemize}
In order to ensure that the censoring information is utilized by all the discriminative predictors, we train every predictor by constructing a training dataset that comprises the physiological data gathered within a temporal window before the censoring event (ICU admission or patient discharge), and using the censoring information (i.e. the variable $l$) as the labels. The size of this window is a hyper-parameter that is tuned separately for every predictor. For the testing data, the predictors are applied sequentially to a sliding window of every patient's episode, and the predictor's output is considered as the patient's real-time risk score\footnote{This differs from the static simulation setting in (\cite{ghassemi2015multivariate}) were predictions are issued in a one-shot fashion using only the data obtained within 24 hours after a patient's admission.}. We used the built-in MATLAB functions for training the logistic regression, LASSO and random forest predictors.\\
\smallskip
\\
Although RNNs are not clinically interpretable, they have been frequently applied to the problem of clinical time series prediction, and the recent work in (\cite{che2016recurrent}) have considered RNNs to predict mortality in the ICU using the MIMIC dataset (\cite{saeed2002mimic}). We have trained an RNN with 5 hidden layers, and 10 neurons with each layer, using the {\it Broyden-Fletcher-Goldfarb-Shanno} (BFGS) algorithm\footnote{We have also tried the Levenberg-Marquardt algorithm, but the network learned by BFGS offered a significantly better performance.}, where gradients are computed using the {\it Backpropagation Through Time} algorithm (\cite{werbos1990backpropagation}). All the training time series were temporally aligned via the endpoint censoring information, and training was accomplished via 1000 iterations of the BFGS algorithm. A top layer with a squashing sigmoid function was used to map the RNN hidden states to a risk score between 0 and 1 at each point in time.\\
\\  
\smallskip
We used the correlated feature selection algorithm to select the physiological stream for every predictor (\cite{yu2003feature}). To ensure a fair comparison, we did not include the static co-variates in any predictor, including the HASMM, since they are not used by the clinical risk scores.\\
\\
\smallskip
{\bf \underline{Generative models}:}\\
\smallskip
In addition to the discriminative models, we also considered the following probabilistic models for the patients' episodes: 
\begin{itemize}
\item {\bf Hidden Markov Models (HMM) with Gaussian emissions.} 
\item {\bf Multi-task Gaussian process.}
\end{itemize}
We used the Baum-Welch algorithm for learning the HMM (\cite{murphy2001bayes}); the informative censoring information was incorporated by including two absorbing states for clinical stability ($l=0$) and deterioration ($l=1$), and informing the forward-backward algorithm with the labeled states at the end of every episode. We tried many initializations for the HMM parameters and picked the initialization that led to the maximum likelihood for the training dataset. The complete data log likelihood after 100 EM iterations was -1.25$\times$ 10$^7$. In real-time, a patient's risk score at every point of time is computed by first applying forward filtering to obtain the posterior probability of the patient's states, and then averaging over the distribution of the absorbing states. Using the Bayesian Information Criterion, we selected an HMM model with 4 latent states.\\
\\
For the multi-task Gaussian process, we used the free-form parametrization in (\cite{bonilla2007multi}), and used the gradient method to learn the parameters of two Gaussian process models: one for patients with $l=0$, and one for patients with $l=1$. The risk score for a patient's risk score is computed as the test statistic of a sequential hypothesis test that is based on the two learned Gaussian process models.  
\subsection{Results}
\subsubsection{Performance metrics}
In order to assess the performance of every algorithm, we compute each algorithm's risk score $R(t)$ at every point of time in every patient's episode. We only use the patient episodes in the testing set for performance evaluation. The risk score that is based on an HASMM is evaluated as discussed in Section \ref{sec3.33}. We emulate the ICU admission decisions by setting a threshold on the risk score $R(t)$ above which a patient is identified as ``clinically deteriorating". The accuracy of such decisions are assessed via the following performance metrics: true positive rate (TPR), positive predictive value (PPV) and timeliness. These performance metrics are formally defined as follows:
\[\mbox{TPR} = \frac{\# \, \mbox{patients with $l=1$ and $R(t)$ exceeding threshold for some $t < Tc$}}{\# \, \mbox{patients with $l=1$}},\]
\[\mbox{PPV} = \frac{\# \, \mbox{patients with $l=1$ and $R(t)$ exceeding threshold for some $t < Tc$}}{\# \, \mbox{patients with $R(t)$ exceeding threshold for some $t < Tc$}},\]
and
\[\mbox{Timeliness} = \mathbb{E}\left[\left. \mbox{Time at which $R(t)$ exceeds threshold} - T_c \,\right|\, \mbox{$R(t)$ exceeds threshold}, l=1\right].\]    
The three performance metrics described above evaluate the different risk scoring algorithms in terms of their detection power, false alarm rate, and timeliness in detecting clinical deterioration. We sweep the threshold value of every risk scoring algorithm and report the AUC of the TPR vs. PPV ROC curve. All results reported hereafter are statistically significant ($p$-value $<$ 0.001).
\subsubsection{Learning the HASMM}
We applied the FFBS-MCEM algorithm to the training episodes in order to estimate the parameter set $\Gamma$. Based on the Bayesian information criterion, we have selected a model with 4 clinical states, i.e. $\mathcal{X} = \{1,2,3,4\}$. State 1 is the clinical stability state for which a patient can be discharged, whereas state 4 is the clinical deterioration state at which an ICU admission is necessary. We ran 100 MCEM iterations and used $\hat{\Gamma}^{\mbox{\tiny 100}}$ as the estimate for $\Gamma$.\\
\\
We discretized the time domain into steps of 1 hour while computing the elements of the look-up table holding the values of the tensor ${\bf \tilde{P}}$. With a granular 1-hour discretization of the time horizon, the Gaussian covariance matrix was found to be ill-conditioned for many patient episodes. To ensure the numerical stability of the computations involving the Gaussian process likelihood functions, we used the Moore-Penrose pseudo-inverse for the covariance matrix instead of direct matrix inversion. The function $\texttt{TransitionLookUp}$ was invoked once before running the MCEM iterations, and its run time was 2 minutes and 15 seconds on a dual-core 3 GHz machine. The function $\texttt{ForwardFilter}$ was invoked 150,852 times (all data points in all patients' episodes in the testing set), and its overall run time was 3 hours and 50 minutes (on a dual-core 3 GHz machine). The run time for every risk score update for a single patient is less than 1 second, which implies that the algorithm can efficiently prompt quick risk assessments if implemented on a machine with a reasonable computational power. \\
\\
From the learned HASMM, we were able to extract the following ``medical concept" out of the training data. The patients' clinical state space $\mathcal{X} = \{1,2,3,4\}$ comprises the following 4 states:
\begin{itemize}
\item {\bf State 1: clinical stability.}
\item {\bf State 2: type-1 critical state.}
\item {\bf State 3: type-2 critical state.}
\item {\bf State 4: clinical deterioration.}
\end{itemize}
\begin{sidewaysfigure}
        \centering
        \includegraphics[width=8.75in]{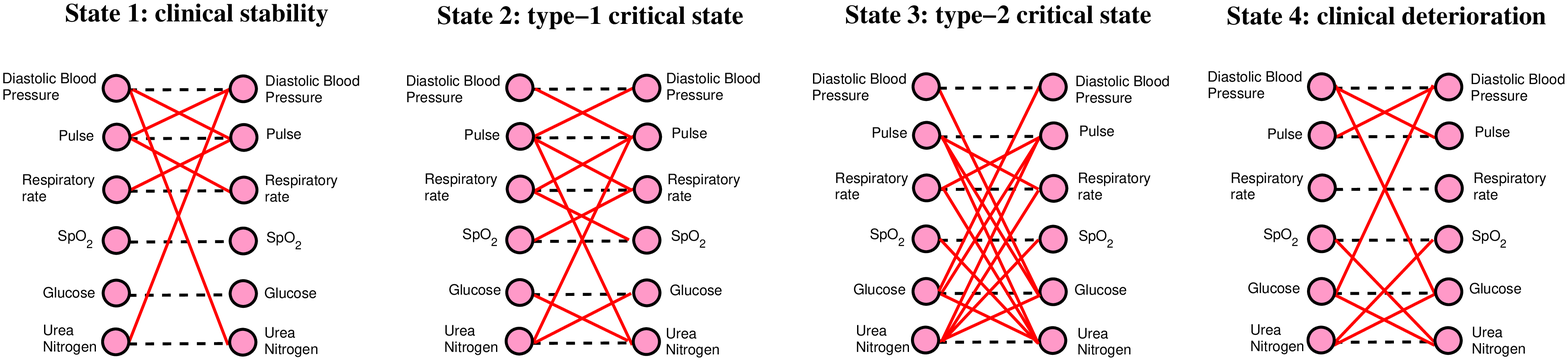}
				\captionsetup{font= small}
        \caption{Correlations between the patients' physiological streams in the different clinical states.}
			  \label{Eps2}
				\vspace{0.5in}
				\begin{minipage}[b]{0.325\textwidth}
				\includegraphics[width=2.75in]{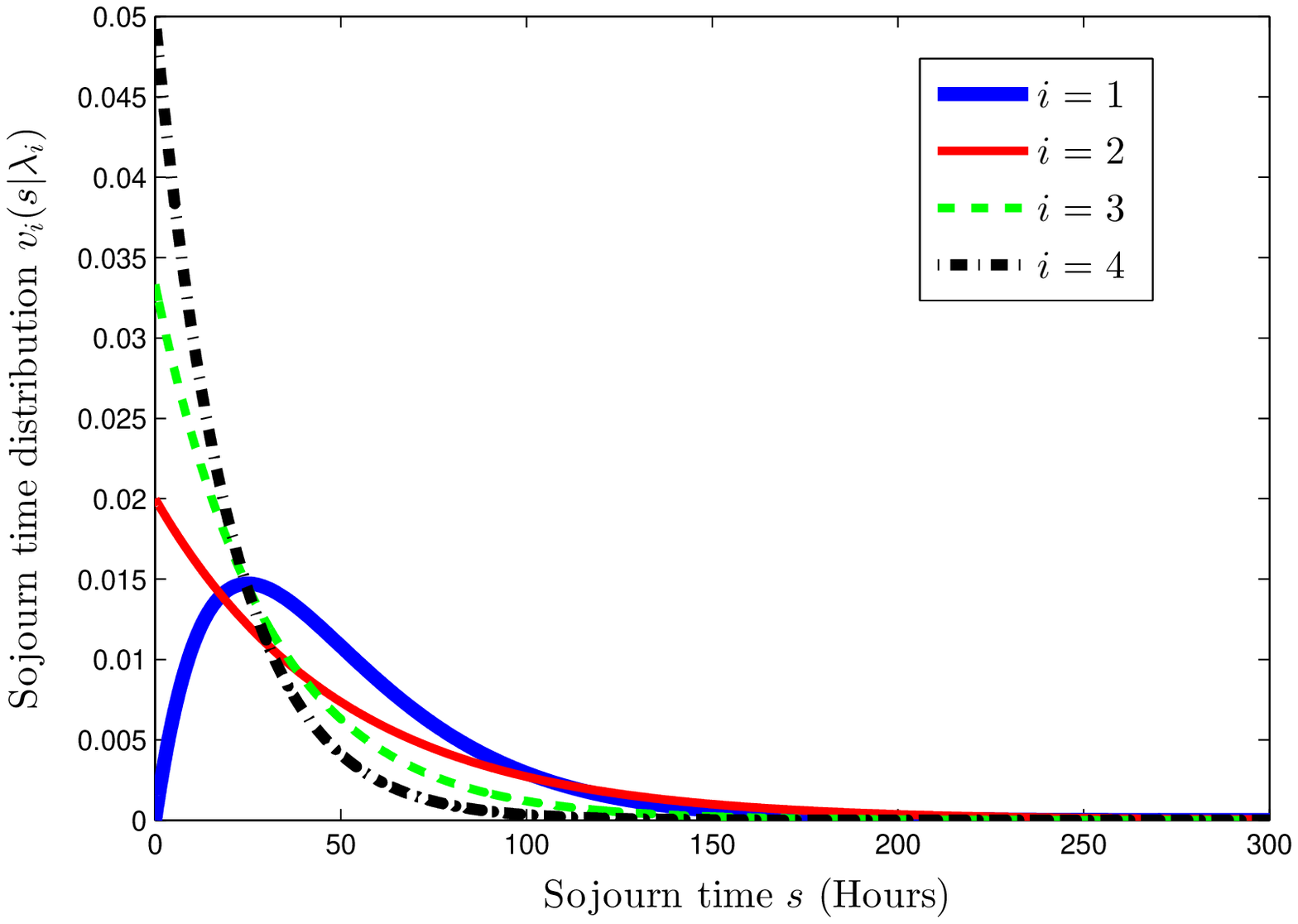}
				\captionsetup{font= small}
        \caption{Sojourn time distributions.}
				\label{FigQ11}			
				\end{minipage}
				%\hfill
				\begin{minipage}[b]{0.325\textwidth}
				\includegraphics[width=2.75in]{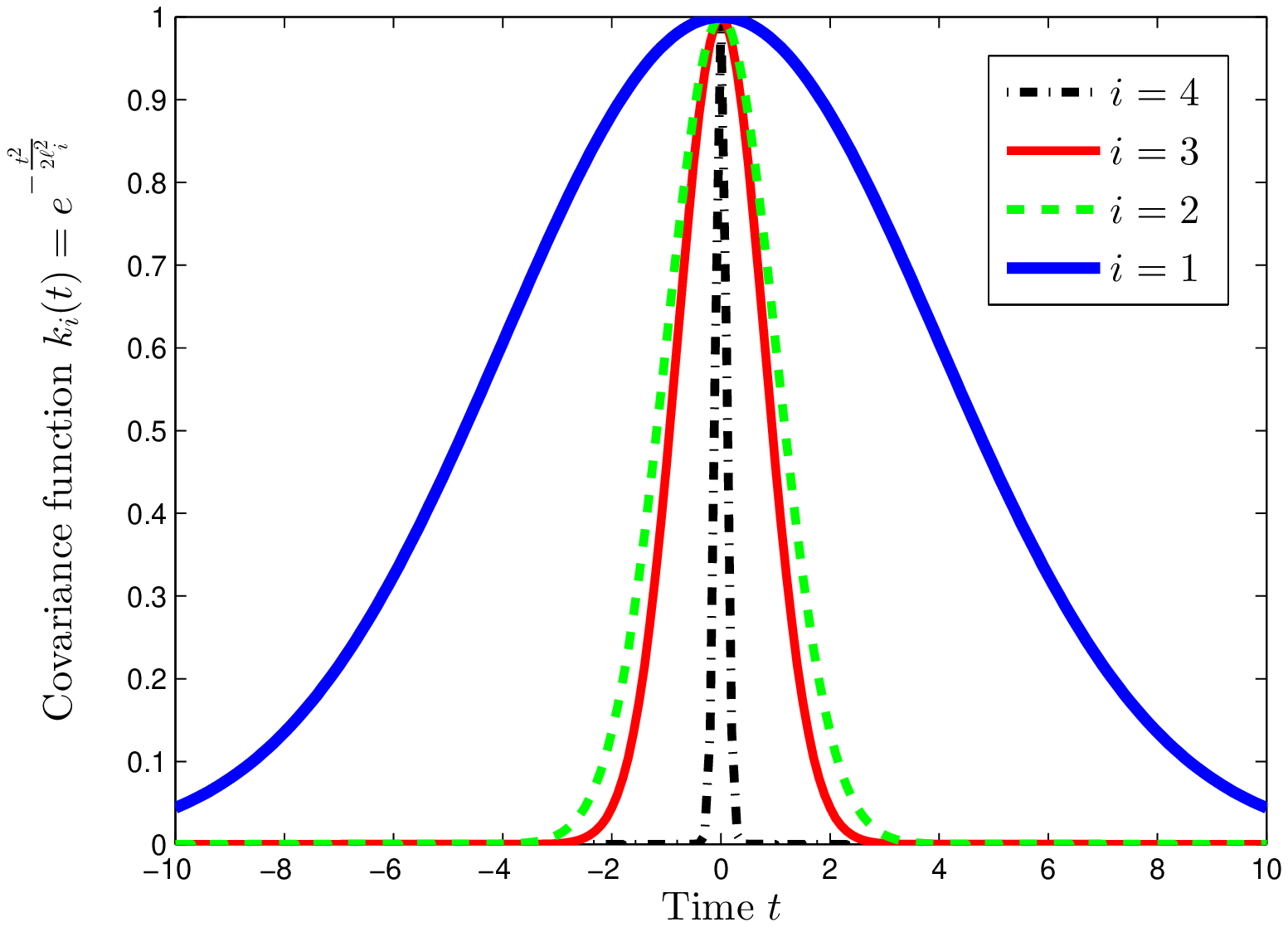}
				\captionsetup{font= small}
        \caption{Covariance functions.}
				\label{FigQ12}				
			  \end{minipage}
		    \begin{minipage}[b]{0.325\textwidth}
				\includegraphics[width=2.75in]{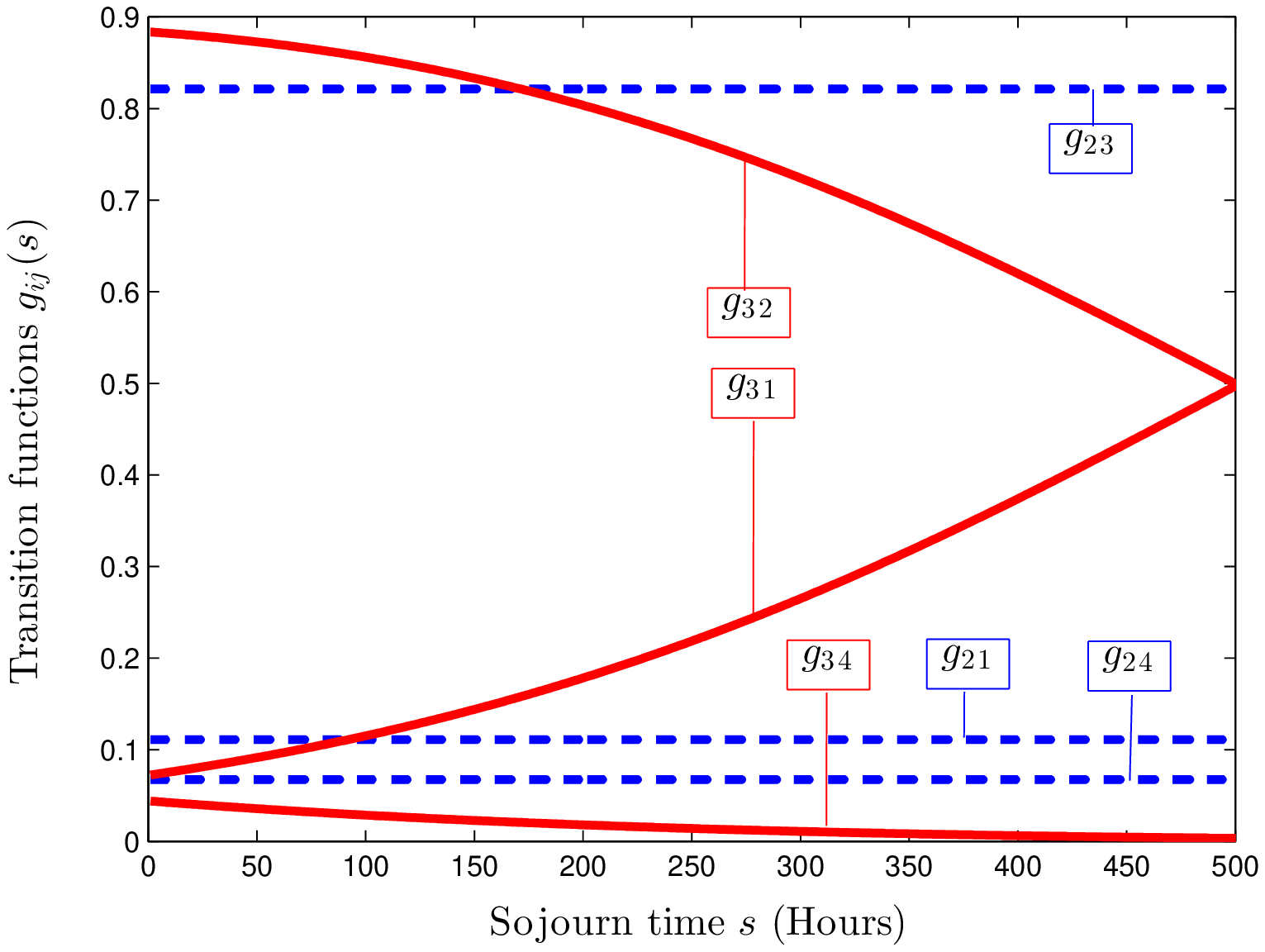}
				\captionsetup{font= small}
        \caption{Transition functions.}
				\label{FigQ13}				
			  \end{minipage}
\end{sidewaysfigure}
As implied by the model, states 1 and 4 are absorbing states: once the patient is believed to be in state 1, the clinicians should release her from care, whereas exhibiting clinical state 4 should be treated with an admission to the ICU. States 2 and 3 are critical states that require the patient to stay under vigilant care in the ward. The two states are different ways to manifest ``criticality". We characterize the properties of the four clinical states in the rest of this subsection.\\
\\
Figures \ref{Eps2}-\ref{FigQ13} depict the different characteristics of the four clinical states. In Figure \ref{Eps2}, we plot a bipartite correlation graph that shows the correlations among the relevant physiological streams in the different clinical states. These graphs were constructed by computing the {\it Pearson correlation coefficient} $\sigma_{Y^lY^v} = \frac{\mbox{cov}(Y^l,Y^v)}{\sigma_{Y^l}\,\cdot\,\sigma_{Y^v}}$ using the entries of the multi-task Gaussian process covariance matrix ${\bf \Sigma}$. An edge is connected between every two features for whom the Pearson correlation coefficient exceeds 0.1, i.e. $\sigma_{Y^lY^v} > 0.1$. As we can see, different physiological variables become less or more correlated in the different clinical states. For instances, only the clinically stable patients experience significant correlations between their urea Nitrogen and the diastolic blood pressure; the Pearson coefficient between those variables becomes insignificant in the other states. Clinicians can use this piece of information, extracted solely from the data, to construct simple tests for clinical stability by computing the correlations between blood pressure and urea Nitrogen for a hospitalized patient before deciding to discharge her. Generally speaking, we observe that the critical, transient states display more correlations among the physiological streams than the clinical stability and deterioration states. In particular, the type-2 critical state has most of the physiological streams being strongly correlated. We speculate that the reason behind these strong correlations is that some kinds of interventions (e.g. drugs, mechanical pumps, ventilators, etc) applied to hospitalized patients affect all the physiological streams simultaneously; and hence we believe that type-1 and type-2 critical state patients are hospitalized patients with and without clinical interventions. We will examine this claim when we retrieve information about interventions and the time they were applied from the Ronald Reagan medical center; such information was not available at the time of conducting these experiments.\\
\\
Figure \ref{FigQ11} shows the sojourn time distributions for the four states. Recall that the ``sojourn time" of an absorbing state (state 1 or 4) is defined as the time between entering the state and the censoring time; such a time interval corresponds to the clinicians' policy with respect to patient discharge and ICU admission. That is, the sojourn time of an absorbing state is not a natural physiological quantity, but it rather reflects the speed with which patients are released from care or receive leveraged level of care. As we can see, the sojourn time distributions significantly deviate from an exponential distribution of an ordinary, memoryless Markov model, which supports our assumption of semi-Markovianity.\\
\\
Figure \ref{FigQ12} displays the covariance function $k_i(t,t^{'})$ for the 4 clinical states; the state-specific covariance function quantifies the physiological streams' temporal correlations in a particular clinical conditions. Knowing such correlation patterns are useful for deciding the frequency with which nurses and clinicians should collect physiological measurements over time for different patients in different clinical conditions (\cite{ibrahim2016balancing}). We observed that, as one would expect, the temporal correlations increase when the patient becomes more stable; the temporal correlation is greatest in state 1 and smallest in state 4. This means that one would expect deteriorating patients to experience more physiological fluctuations over time. We also note that physiological stream for which the constant mean function differed significant among the clinical state was the urea Nitrogen. The level of urea Nitrogen increases significantly when the patient is in a more risky state; the average blood urea nitrogen is 11.7 milligrams per deciliter (mg/dL) in state 1, 23.8 mg/dL in state 2, 41.1 mg/dL in state 3 and 64.9 mg/dL in state 4.\\    
\\
Figure \ref{FigQ13} depicts the transition functions $g_{ij}$ out of the transient states 2 and 3 as a function of the sojourn time in those states. We note that the transition probabilities are almost a constant function of sojourn time for patients in state 2 ($\beta_{2j} \approx 0$), whereas the duration-dependence is more significant ($\beta_{3j}>0$); as the sojourn time in state 3 increases, the transition probabilities become more biased towards state 1. This reinforces our hypothesis that state 3 corresponds to patients for whom interventions were applied. That is, as time passes for a patient in state 3 after receiving an intervention, her chances for recovery (transiting to state 1) increases.\\
\\
\begin{figure*}[h]
        \centering
        \includegraphics[width=6in]{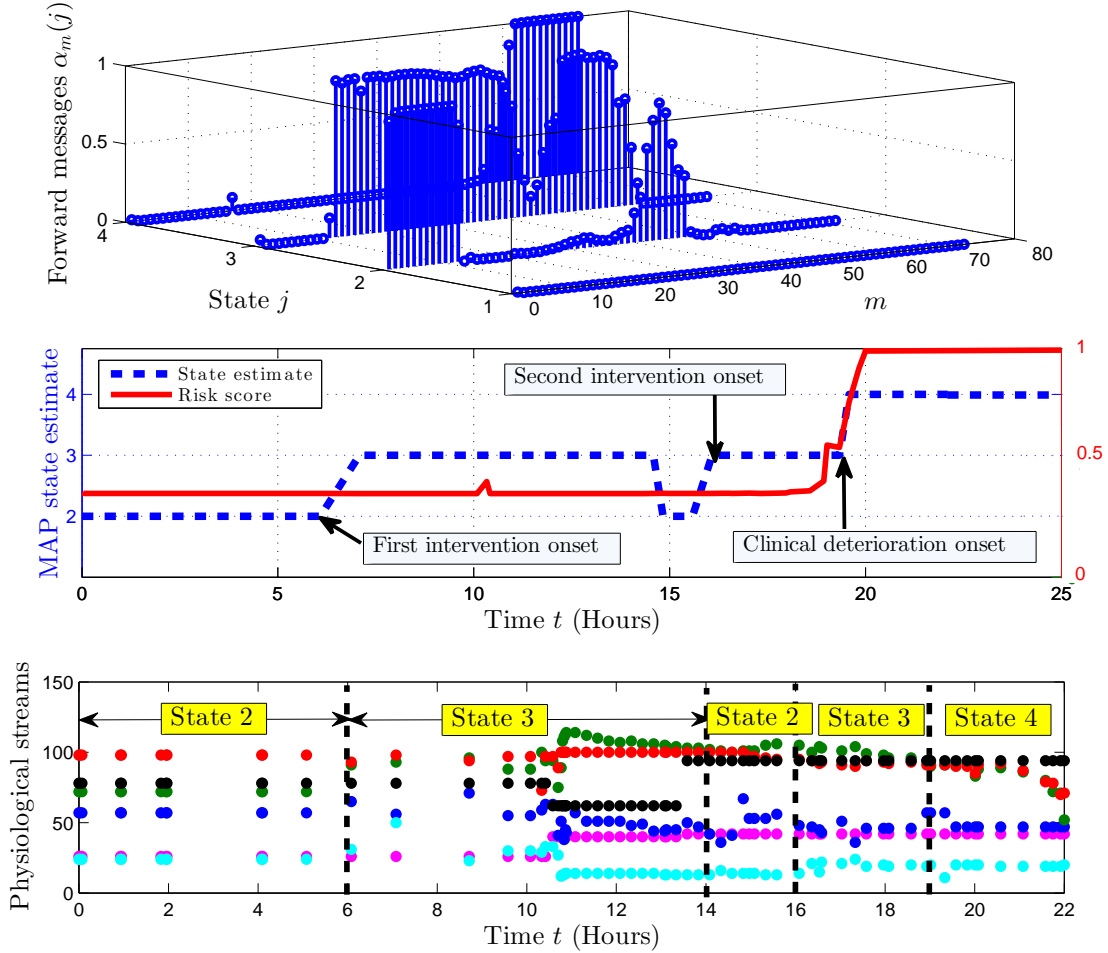}
				\captionsetup{font= small}
        \caption{Depiction for the episode of a clinically deteriorating patient.}
			  \label{Eps5}
\end{figure*}
Now we illustrate the real-time operation of the inference algorithm as it computes risk score over time by focusing on an episode of a particular patient who was hospitalized for 1 day and then admitted to the ICU. As shown in Figure \ref{Eps5} (top), the inference algorithm computes the forward messages whenever new physiological measurements become available. Using the forward messages, the algorithm can display the maximum a posteriori (MAP) state estimates to the clinicians over time. As we can see in Figure \ref{Eps5} (middle), the patient under consideration was in clinical state 2 (type-1 critical state) at the time of admission to the ward. After 6 hours, the patient switched to state 3 (type-2 critical state), probably due to a clinical intervention. After around 9 hours, the patient switched back to the type-1 critical state for a brief 2-hour period, before switching to the type-2 critical state (probably due to a second intervention). Our algorithm was able to detect clinical deterioration (state 4) conclusively (through both the MAP state estimate and the risk score) more than 6 hours before the clinicians actually sent the patient to the ICU. Had the clinicians used the algorithm for monitoring that patient, they would have been able to send the patient to the ICU 6 hours early, allowing for a potentially much more efficient therapeutic intervention in intensive care. In Figure \ref{Eps5} (bottom), we plot the patient's physiological stream and tag the different time intervals with the corresponding clinical state estimates. The clinicians can rely on these clinically interpretable tags to describe the patient's states at each point of time rather than using a high-dimensional, and potentially inexpressive set of physiological measurements.    
\subsubsection{Performance comparisons}
\begin{table}
    \centering
    \caption{Performance comparisons for various algorithms.}
\begin{tabular}{|c||c|c|c|c|c|c|c|c|c|c|c|}  
\toprule[2.5pt]
{\bf Algorithm} & \begin{turn}{90}HASMM\end{turn} & \begin{turn}{90}Random Forest\end{turn} & \begin{turn}{90}Logistic regression\end{turn} & \begin{turn}{90}LASSO\end{turn} & \begin{turn}{90}RNN\end{turn} & \begin{turn}{90}HMM\end{turn} & \begin{turn}{90}MTGP\end{turn} & \cellcolor{lightgray} \begin{turn}{90}Rothman\end{turn} & \cellcolor{lightgray} \begin{turn}{90}MEWS\end{turn} & \cellcolor{lightgray} \begin{turn}{90}SOFA\end{turn} & \cellcolor{lightgray} \begin{turn}{90}APACHE\end{turn} \\
\hline
\hline
{\bf AUC} & 0.49 & 0.36 & 0.27 & 0.26 & 0.29 & 0.32 & 0.3 & 0.25 & 0.18 & 0.13 & 0.14 \\
\bottomrule[2.5pt]
\end{tabular}
\label{Table3ICU}
\end{table}
In order to handle class imbalance, we focused on the TPR vs. PPV performance rather than the TPR vs. FPR analysis commonly adopted in the medical literature, which overlooks the class imbalance problem\footnote{Since we focus on the AUC for the TPR vs. PPV, the AUC values are nominally less than that for the TPR vs. FPR curves. The AUC values in the TPR vs. PPV are usually less than 0.5, whereas in the TPR vs. FPR analysis they can reach 0.8 (\cite{rothman2013development}).}. We report the AUC values for all the algorithms under consideration in Table \ref{Table3ICU}. As we can see, all the machine learning algorithms significantly outperform the state-of-the-art clinical risk scores (Rothman, MEWS, APACHE and SOFA). The reason behind the significant performance gain of the HASMM as compared to the clinical risk scores is that it incorporates the patients' history when updating the forward messages (as shown in Figure \ref{Eps5}), and reasons about the future trajectory when computing the risk score (as discussed in Section \ref{sec3.33}). Clinical risk scores are instantaneous in that they map the current physiological measurements to a risk score without considering the previously measured physiological variables, and hence they are vulnerable to high false alarm rates (low PPV). Moreover, the clinical risk scores do not reason about the future trajectory given the current physiological measurements, and hence they display a sluggish risk signal that fail to quickly cope with subtle clinical deterioration.\\       
\\
We also note that HASMMs outperforms conventional HMMs; this is a consequence of incorporating temporal correlations and semi-Markovian state transitions, which more accurately describe the patient's physiology. This manifests in the sojourn time distributions in Figure \ref{FigQ11}, which largely deviate from the exponential distribution adopted by an HMM, and also manifests in the temporal correlation patterns in Figure \ref{FigQ12}, which largely deviate from the Dirac-delta function. Moreover, not only that the HASMM outperforms discriminative classifiers that operate on a sliding window of the clinical time series, but unlike these classifiers, it provides a clinically interpretable model as well (see Figure \ref{Eps5}). Such an interpretable model cannot be provided by discriminative approaches such as logistic regression or RNNs. In addition to the accuracy gains demonstrated in Table \ref{Table3ICU}, we also note that ICU alarms issued by the HASMM precedes actual ICU admission decisions issued by the clinicians with 8-9 hours on average for a TPR of 50$\%$ and PPV of 35$\%$.     
\section{Conclusions}
\label{sec5}
We developed a versatile model, which we call the Hidden Absorbing Semi-Markov Model (HASMM), for clinical time series data which accurately represents physiological data in modern EHRs. The HASMM can deal with irregularly sampled, temporally correlated, and
informatively censored physiological data with non-stationary clinical state transitions. We also proposed an efficient Monte Carlo EM learning algorithms that is based on particle filtering, and developed an inference algorithm that can effectively carry out real-time inferences. We have shown, using a real-world dataset for patients admitted to the Ronald Reagan UCLA Medical Center, that HASMMs provide a significant gain in critical care prognosis when utilized for constructing an early warning and risk scoring system.

\setcounter{secnumdepth}{-1}
\section{Acknowledgments}
We would like to thank Dr. Scott Hu (Division of Pulmonary and Critical Care Medicine, Department of Medicine, David Geffen School of Medicine, UCLA) for providing us with the clinical data and the appropriate medical background and insights used in Section \ref{sec4}. We also thank Mr. Jinsung Yoon for his valuable help with the simulations in Section \ref{sec4}. This research was funded by grants from the Office of Naval Research (ONR) and NSF ECCS 1462245.

\setcounter{secnumdepth}{1}
\appendix

\section{Proof of Theorem 1}
\label{appA}
We start by rewriting (\ref{eqqInf6}) as follows:
\begin{align}
&\begin{bmatrix}
    \tilde{p}_{11}(\tau, \munderbar{s}, \bar{s}) & \dots  & \tilde{p}_{1N}(\tau, \munderbar{s}, \bar{s}) \\
    \vdots & \ddots & \vdots \\
    \tilde{p}_{N1}(\tau, \munderbar{s}, \bar{s}) & \dots  & \tilde{p}_{NN}(\tau, \munderbar{s}, \bar{s})
\end{bmatrix}
= \begin{bmatrix}
    1-\bar{Q}_{1}(\tau, \munderbar{s}, \bar{s}) & \dots  & 0 \\
    \vdots & \ddots & \vdots \\
    0& \dots  & 1-\bar{Q}_{N}(\tau, \munderbar{s}, \bar{s})
\end{bmatrix}
+ \nonumber \\
&\int_{u=0}^{\tau} \left(\frac{\partial}{\partial u} \begin{bmatrix}
    \bar{Q}_{11}(u, \munderbar{s}, \bar{s}) & \dots  & \bar{Q}_{1N}(u, \munderbar{s}, \bar{s}) \\
    \vdots & \ddots & \vdots \\
    \bar{Q}_{N1}(u, \munderbar{s}, \bar{s}) & \dots  & \bar{Q}_{NN}(u, \munderbar{s}, \bar{s})
\end{bmatrix}\right)\,\times\,
\begin{bmatrix}
    \tilde{p}_{11}(\tau-u, 0, 0) & \dots  & \tilde{p}_{1N}(\tau-u, 0, 0) \\
    \vdots & \ddots & \vdots \\
    \tilde{p}_{N1}(\tau-u, 0, 0) & \dots  & \tilde{p}_{NN}(\tau-u, 0, 0)
\end{bmatrix}
\, du. 
\label{appeq1}
\end{align}
Starting with the left hand side, we can use a first-step analysis to write every term $\tilde{p}_{ij}(\tau, \munderbar{s}, \bar{s})$ as follows 
\begin{align}
\tilde{p}_{ij}(\tau, \munderbar{s}, \bar{s}) &= \mathbb{P}(X(t+\tau)=j|X(t)=i, \munderbar{s} \leq S(t)\leq \bar{s})\nonumber \\
&= \delta_{ij}\,(\mathbb{P}(S_i<\tau|X(t)=i,\munderbar{s} \leq S(t)\leq \bar{s})) +\nonumber \\
&\int_{u=0}^{\tau}\mathbb{P}(X(t+u)=k|X(t)=i, \munderbar{s} \leq S(t)\leq \bar{s})\,\cdot\,\mathbb{P}(X(t+\tau)=j|X(t+u)=k)\, du \nonumber \\
&= \delta_{ij}\,(1-\bar{Q}_{i}(\tau,\munderbar{s},\bar{s})) +\nonumber \\
&\int_{u=0}^{\tau}\mathbb{P}(X(t+u)=k|X(t)=i, \munderbar{s} \leq S(t)\leq \bar{s})\,\cdot\,\mathbb{P}(X(t+\tau-u)=j|X(t)=k)\, du \nonumber \\
&= \delta_{ij}\,(1-\bar{Q}_{i}(\tau,\munderbar{s},\bar{s})) + \int_{u=0}^{\tau}\frac{\partial}{\partial u}\,\sum_{k \neq i}\bar{Q}_{ik}(u,\munderbar{s},\bar{s})\,\cdot\,\tilde{p}_{kj}(\tau-u,0,0)\, du,
\label{appendixeq1} 
\end{align}
$\forall i,j \in \mathcal{X},$ where $S(t)$ is the time elapsed in state $X(t)$, and $S_i$ is the sojourn time of state $i$. The integral equation in (\ref{appendixeq1}) can be written in a matrix form as in the right hand side of (\ref{appeq1}), and hence the Theorem follows.  
\section{Proof of Theorem 2}
\label{appB}
Recall that the operation
\[{\bf \tilde{P}}(\tau, \munderbar{s}, \bar{s}) = \mathcal{B}\{{\bf \bar{Q}}(\tau, \munderbar{s}, \bar{s})\}({\bf \tilde{P}}(\tau, \munderbar{s}, \bar{s}))\]  
can be written as
\[{\bf \tilde{P}}(\tau, \munderbar{s}, \bar{s}) = {\bf I}_{N \times N} - \mbox{\textnormal{diag}}\left(\bar{Q}_{1}(\tau, \munderbar{s}, \bar{s}), \dots, \bar{Q}_{N}(\tau, \munderbar{s}, \bar{s})\right) + \left(\frac{\partial{\bf \bar{Q}}(., \munderbar{s}, \bar{s})}{\partial u} \,\star\, {\bf \tilde{P}}(., 0, 0)\right)(\tau).\]
Now consider $n$ applications of the operator $\mathcal{B}(.)$, we have that
\begin{align}
\left(\frac{\partial{\bf \bar{Q}}(., \munderbar{s}, \bar{s})}{\partial u_1} \,\star\,.\,.\,.\,\star\,\frac{\partial{\bf \bar{Q}}(., \munderbar{s}, \bar{s})}{\partial u_n}\,\star\, {\bf \tilde{P}}(., 0, 0)\right)(\tau) &\leq N^n\,\cdot\,\int^{\tau}_{0}\int^{\tau-u_{n-1}}_{0}.\,.\,.\int^{\tau-u_{1}}_{0} du_1\,du_2.\,.\,,du_n \nonumber \\
&= N^n\,\cdot\,\frac{\tau^n}{n!}.
\end{align}
Thus, for every ${\bf \tilde{P}}(\tau, \munderbar{s}, \bar{s}) \in \mathcal{P}$ and every ${\bf \bar{Q}}(\tau, \munderbar{s}, \bar{s}) < 1$, there exists $n$ such that $\mathcal{B}^n\{.\}(.)$ is a contraction mapping. Therefore, the operation ${\bf \tilde{P}}(\tau, \munderbar{s}, \bar{s}) = \mathcal{B}\{{\bf \bar{Q}}(\tau, \munderbar{s}, \bar{s})\}({\bf \tilde{P}}(\tau, \munderbar{s}, \bar{s}))$ has a unique fixed point that can be reached via $n \in \mathbb{N}$ successive approximations.  
\section{Proof of Theorem 3}
\label{appE}
The Theorem can be proven using the proof of Theorem 5 in (\cite{neath2013convergence}).

\vskip 0.2in
%\nocite{*}
\bibliography{jmlr_ref2}

\end{document}